\documentclass{article}

\usepackage{microtype}
\usepackage{graphicx}
\usepackage{subfigure}
\usepackage{booktabs} 

\usepackage{hyperref}

\usepackage{amsmath}
\usepackage{amsthm}
\usepackage{amssymb}
\usepackage{mathtools}
\usepackage{color}
\usepackage[switch]{lineno}
\usepackage[ruled,vlined]{algorithm2e}
\usepackage{dsfont}
\usepackage{footnote}
\makesavenoteenv{tabular}

\usepackage[accepted]{icml2021}

\icmltitlerunning{Accelerating Gossip SGD with Periodic Global Averaging}

\newcommand{\ba}{\left[ \begin{array}}
\newcommand{\ea}{\\ \end{array} \right]}

\def\g{{\boldsymbol{g}}}

\def\s{{\boldsymbol{s}}}
\def\x{{\boldsymbol{x}}}
\def\y{{\boldsymbol{y}}}

\newcommand{\vx}{{\mathbf{x}}}
\newcommand{\vy}{{\mathbf{y}}}

\newcommand{\cE}{{\mathcal{E}}}
\newcommand{\cF}{{\mathcal{F}}}
\newcommand{\cG}{{\mathcal{G}}}

\newcommand{\cN}{{\mathcal{N}}}

\newcommand{\cU}{{\mathcal{U}}}
\newcommand{\cV}{{\mathcal{V}}}

\newcommand{\bxi}{{\boldsymbol{\xi}}}

\newcommand{\RR}{\mathbb{R}}

\newcommand{\vvvert}{{\vert\kern-0.25ex\vert\kern-0.25ex\vert}}

\newtheorem{theorem}{Theorem}
\newtheorem{definition}{Definition}
\newtheorem{assumption}{Assumption}
\newtheorem{corollary}{Corollary}
\newtheorem{remark}{Remark}
\newtheorem{lemma}{Lemma}

\allowdisplaybreaks

\begin{document}

\twocolumn[
\icmltitle{Accelerating Gossip SGD with Periodic Global Averaging}



\icmlsetsymbol{equal}{*}

\begin{icmlauthorlist}
\icmlauthor{Yiming Chen}{equal,ali}
\icmlauthor{Kun Yuan}{equal,ali}
\icmlauthor{Yingya Zhang}{ali}
\icmlauthor{Pan Pan}{ali}
\icmlauthor{Yinghui Xu}{ali}
\icmlauthor{Wotao Yin}{ali}
\end{icmlauthorlist}

\icmlaffiliation{ali}{Alibaba Group, Hangzhou, China}

\icmlcorrespondingauthor{Kun Yuan}{kun.yuan@alibaba-inc.com}

\icmlkeywords{Machine Learning, ICML}

\vskip 0.3in
]

\SetKwInput{KwInput}{Input}                
\SetKwInput{KwOutput}{Output}              
\SetKwInput{KwRequire}{Require}              



\printAffiliationsAndNotice{\icmlEqualContribution} 

\begin{abstract}

Communication overhead hinders the scalability of large-scale distributed training. Gossip SGD, where  each node averages only with its neighbors, is more communication-efficient than the prevalent parallel SGD. However, its convergence rate is reversely proportional to quantity $1-\beta$ which measures the network connectivity. On large and sparse networks where $1-\beta \to 0$, Gossip SGD requires more iterations to converge, which offsets  against its communication benefit. This paper introduces Gossip-PGA, which adds \underline{P}eriodic  \underline{G}lobal \underline{A}veraging into \underline{Gossip} SGD. Its transient stage, i.e., the iterations required to reach asymptotic linear speedup stage, improves from $\Omega(\beta^4 n^3/(1-\beta)^4)$ to $\Omega(\beta^4 n^3 H^4)$ for non-convex problems. The influence of network  topology in Gossip-PGA can be controlled by the averaging period $H$. Its transient-stage complexity is also superior to Local SGD which has order $\Omega(n^3 H^4)$. Empirical results of large-scale training on image classification (ResNet50) and language modeling (BERT) validate our theoretical findings.

\end{abstract}

\section{Introduction}
\label{submission}

The scale of deep learning nowadays calls for efficient {large-scale} distributed training across multiple computing nodes {in the data-center clusters}. In distributed optimization, a network of $n$ nodes cooperate to solve the problem
\begin{align}\label{eq:general-prob}
&\ \min_{x \in \RR^d}\ \frac{1}{n}\sum_{i=1}^n [f_i(x) : = \mathbb{E}_{\boldsymbol{\xi}_i \sim D_i}F_i(x; \boldsymbol{\xi}_i)]
\end{align}
where each component $f_i$ is local and private to node $i$ and the random variable $\bxi_i$ denotes the local data that follows distribution $D_i$. We assume each node $i$ can locally evaluate stochastic gradients $\nabla F_i(x; \boldsymbol{\xi}_i)$ where $\bxi_i \sim D_i$, but must communicate to access information from other nodes.

{Parallel SGD methods are leading algorithms to solve \eqref{eq:general-prob}}, in which every node processes local training samples independently, and synchronize {gradients} every iteration either using a central {\em Parameter Server (PS)} \cite{li2014scaling} or the {\em All-Reduce} communication primitive \cite{patarasuk2009bandwidth}. The global synchronization in Parallel SGD either incurs significant bandwidth cost or high latency, which hampers the training scalability.

Many alternative methods have been proposed to reduce communication overhead in distributed training. Gossip SGD, also known as decentralized SGD \cite{nedic2009distributed, chen2012diffusion, lian2017can, lian2018asynchronous, assran2019stochastic}, recently received lots of attention. This line of work lets each node communicate with (some of) their direct neighbors. In a sparse topology such as one-peer exponential graph \cite{assran2019stochastic}, each node only communicates with {\em one} neighbor each time. This gossip-style communication is much faster than {\em PS} and {\em All-Reduce} but the computed average can be highly inaccurate. Local SGD \cite{stich2019local,yu2019parallel,lin2018don} is another line of work that increases the computation-to-communication ratio. Local SGD lets each node to run local gradient descent for multiple rounds and only average their parameters globally once in a while. By communicating less frequently, Local SGD reduces the communication overhead. 

\begin{table}[t]
  \begin{center}
  \begin{small}
  \begin{sc}
  \begin{tabular}{rccc}
    \toprule
    Method  & Epoch   &  Acc.\%  &  Time(hrs.)\\
    \midrule
    Parallel SGD & 120 & 76.26   & 2.22  \\ 
    Gossip SGD (ring)   & 120  & 74.86        & 1.56  \\
    Gossip SGD (expo)   & 120  & 75.34        & 1.55 \\
    Gossip SGD (ring) & 240   & 75.62        & 3.02 \\
    Gossip SGD (expo) & 240   & 76.18        & 3.03 \\
    \bottomrule
  \end{tabular}
  \end{sc}
\end{small}
\end{center}
\vskip -0.15in
\caption{Top-1 validation accuracy for ImageNet with 256 GPUs connected with the ring or one-peer exponential network. Gossip SGD takes more time to reach the same accuracy as Parallel SGD.}
  \label{Table:Motivation}
  \vspace{-6mm}
\end{table}

\begin{table*}[t!]
\vskip -0.05in
\begin{center}
\begin{small}
\begin{sc}
\begin{tabular}{lcccc}
\toprule
     & \multicolumn{2}{c}{Gossip SGD} & \multicolumn{2}{c}{Gossip-PGA} \\
     & iid          & non-iid         & iid            & non-iid (proposed)   \hspace{1.5mm}    \\ \midrule
small or dense network (when $\frac{1}{1-\beta} < H$) & $\Omega(\frac{n^3 \beta^4 }{\boldsymbol{(1-\beta)^2}})$     & $\Omega(\frac{n^3 \beta^4 }{\boldsymbol{(1-\beta)^4}})$     & $\Omega(n^3 \beta^4  \boldsymbol{C_\beta^2})$      &$\Omega(\frac{n^3 \beta^4  \boldsymbol{C_\beta^2}}{\boldsymbol{(1-\beta)^2}})$   \vspace{1.5mm}   \\
large or sparse network (when $\frac{1}{1-\beta} \ge H$) & $\Omega(\frac{n^3 \beta^4 }{\boldsymbol{(1-\beta)^2}})$    & $\Omega(\frac{n^3 \beta^4 }{\boldsymbol{(1-\beta)^4}})$        & $\Omega(n^3 \beta^4  \boldsymbol{C_\beta^2})$       &$\Omega(n^3 \beta^4 \boldsymbol{C_\beta^2 H^2})$     \\ \bottomrule
\end{tabular}
\end{sc}
\end{small}
\end{center}
\vskip -0.18in
\caption{The lengths of the transient stages of Gossip SGD and  Gossip-PGA. Since $C_\beta = \sum_{k=0}^{H-1} \beta^k = (1-\beta^H)/(1-\beta) < \min\{1/(1-\beta), H\}$, Gossip-PGA always has shorter transient stage, more evident on large and sparse networks where $\beta \to 1$.}
\label{table-transient-stage}
\end{table*}

The reduced communication in Gossip and Local SGDs comes at a cost: slower convergence rate. While both algorithms are proved to have convergence linear speedup asymptotically, they are sensitive to network topology and synchronization period, respectively. For Gossip SGD, the convergence rate is inversely proportional to $1-\beta$ ($\beta$ is defined in Remark \ref{remark-beta-definition}). {Since $\beta \to 1$ on the large and sparse network topology which is most valuable for deep training,} Gossip SGD will converge very slow and {require more iterations than Parallel SGD to achieve a desired solution. This may nullify its communication efficiency and result in even more training time (see Table \ref{Table:Motivation}). Local SGD with a large averaging period meets the same issue.}

This paper proposes Gossip-PGA, which adds \underline{p}eriodic 
All-Reduce \underline{g}lobal \underline{a}veraging into Gossip to accelerate its convergence especially on large and sparse networks. Gossip-PGA also extends Local SGD with fast gossip-style communication after local updates. When the same averaging period $H$ is used, the additional gossip communication in Gossip-PGA endows it with faster convergence than Local SGD.

\noindent \textbf{Challenges}. Gossip-PGA can be regarded as a special form of the topology-changing Gossip SGD \citep{koloskova2020unified} and SlowMo \citep{wang2019slowmo} (in which the base optimizer is set as Gossip SGD, and the momentum coefficient $\beta = 0$). However, its theory and practical performance were not carefully investigated in literature. Unanswered important questions include how much acceleration can PGA bring to Gossip and Local SGDs, in what scenario can PGA benefits most, how to adjust the averaging period  effectively, and how Gossip-PGA performs in large-scale deep learning systems. Providing quantitative answers to these questions requires new understanding on the interplay between gossip communication and global averaging period. Simply following existing analysis in \citep{koloskova2020unified} will result in incomplete conclusions, see Remark \ref{remark-comparison-with-jaggi}. Also, the analysis in SlowMo \citep{wang2019slowmo} does not consider  heterogeneous data distributions and cannot cover our results.

\begin{table}[t]

\begin{center}
\begin{small}
\begin{sc}
\begin{tabular}{lcc}
\toprule
& Local SGD & Gossip-PGA \\ 
\midrule
iid scenario     & $\Omega(n^3 \boldsymbol{H^2})$ & $\Omega(n^3 \boldsymbol{\beta^4 C_\beta^2})$     \\
non-iid scenario & $\Omega(n^3 \boldsymbol{H^4})$ & $\Omega(n^3 \boldsymbol{\beta^4 C_\beta^2 H^2})$ \\
\bottomrule
\end{tabular}
\end{sc}
\end{small}
\end{center}
\vskip -0.1in
\caption{The lengths of the transient stages of Local SGD and Gossip-PGA. Gossip-PGA always has shorter transient stages than Local SGD since $\beta < 1$ and $C_\beta < H$. Such superiority becomes more significant on well-connected networks where $\beta \to 0$.}
\vspace{-5mm}
\label{table-transient-stage-local}
\end{table}

\vskip -0.2in
\subsection{Main Results} \label{sec-intro-main-result}
This paper proves that Gossip-PGA converges at 
\begin{align}\label{rate-introduction}
{O}\Big( \underbrace{\frac{\sigma}{\sqrt{nT}}}_{\mathrm{SGD\ rate}} \hspace{-0.5mm}+ \hspace{-0.5mm}
\underbrace{\frac{C_\beta^{\frac{1}{3}}\beta^{\frac{2}{3}}(\sigma^{\frac{2}{3}} + D_\beta^{\frac{1}{3}} {b}^{\frac{2}{3}})}{T^{\frac{2}{3}}} + \frac{\beta D_\beta}{T}\Big)}_{\mathrm{Extra\ overhead}}
\end{align}
for both smooth {\em convex} and {\em non-convex} functions $f_i$ (the metrics used for both scenarios can be referred to Theorems \ref{thm-convex} and \ref{thm:nc}), where $n$ is the network size, $T$ is the total number of iterations, $\sigma^2$ denotes gradient noise, $b^2$ gauges data heterogeneity, $\beta \in (0, 1)$ measures how well the network is connected, $H$ is the global averaging period, and we define $C_\beta = \sum_{k=0}^{H-1}\beta^{k}$ and $D_\beta = \min\{H, 1/(1-\beta)\}$.

\noindent \textbf{Linear speedup}. When $T$ is sufficiently large, the first term $1/\sqrt{nT}$ dominates \eqref{rate-introduction}. This also applies to Parallel, Local, and Gossip SGDs. Gossip-PGA and these algorithms all require $T = \Omega(1/(n\epsilon^2))$ iterations to reach a desired accuracy $\epsilon$, which is inversely proportional to $n$. We say an algorithm is in its linear-speedup stage at $T$th iteration if, for this $T$, the term involving $nT$ is dominating the rate.

\noindent \textbf{Transient stage}. Transient stage is referred to those iterations before an algorithm reaches its linear-speedup stage, that is iterations $1,\ldots,T$ where $T$ is relatively small so non-$nT$ terms (i.e., the extra overhead  terms in \eqref{rate-introduction}) still dominate the rate. We take Gossip-PGA in the non-iid scenario ($b^{2/3}\ge \sigma$) as example. To reach linear speedup, $T$ has to satisfy  $T^{\frac{2}{3}}/(C_\beta^{\frac{1}{3}}\beta^{\frac{2}{3}} D_\beta^{1/3}) \ge n^{\frac{1}{2}} T^{\frac{1}{2}}$, i.e., $T \ge n^3\beta^4C_\beta^2 D_\beta^2$. So, the transient stage has $\Omega(n^3\beta^4C_\beta^2 D_\beta^2)$ iterations. \textbf{Transient stage is an important metric to measure the scalability of distributed algorithms.}

\noindent \textbf{Shorter transient stage than Gossip SGD}. The transient stage comparison between Gossip SGD and Gossip-PGA is shown in Table \ref{table-transient-stage}. Since $C_\beta = (1-\beta^H)/(1-\beta) < \min\{H, 1/(1-\beta)\}$, we conclude Gossip-PGA {\em always} has a shorter transient stage than Gossip SGD for any $\beta$ and $H$. Moreover, the superiority of Gossip-PGA becomes evident when the network is large and sparse, i.e., $1-\beta \to 0$. In this case, the transient stage of Gossip SGD can 
grow dramatically (see the second line in Table \ref{table-transient-stage}) while Gossip-PGA is controlled by the global period $H$ because $C_\beta < H$. As a result, Gossip-PGA improves the transient stage of Gossip-SGD from $O(n^3/(1-\beta)^4)$ (or $O(n^3/(1-\beta)^2$ in the iid scenario) to $O(n^3)$ when $\beta \to 1$.

\noindent \textbf{Shorter transient stage than Local SGD}. The transient stage comparison between Local SGD and Gossip-PGA is shown in Table \ref{table-transient-stage-local}. Using $C_\beta < H$, we find  Gossip-PGA is {\em always} endowed with a shorter transient stage than Local SGD. Moreover, when the network is well-connected such that $\beta \to 0$, it holds that {\color{blue}$C_\beta \to 1$}. Gossip-PGA will have a significantly shorter transient stage than Local SGD. 
\vskip -0.1in
\subsection{Contributions}
\begin{itemize}

\item We establish the convergence rate of Gossip-PGA for both smooth convex and non-convex problems. Our results clarify how gossip communication and periodic global averaging collaborate to improve the transient stage of Gossip and Local SGDs. We also established shorter wall-clock training {\em times} of Gossip-PGA.

\item We propose Gossip-AGA, which has \underline{a}daptive \underline{g}lobal \underline{a}veraging periods. Gossip-AGA automatically adjusts $H$ and has convergence guarantees.

\item We conduct various experiments (convex logistic regression and large-scale deep learning tasks) to validate all established theoretical results. In particular, the proposed Gossip-PGA/AGA achieves a similar convergence speed to parallel SGD in {\em iterations}, but provides 1.3 $\sim$ 1.9$\times$ {\em runtime} speed-up. The introduced global averaging steps in Gossip-PGA/AGA remedy the accuracy degradation in Gossip SGD and Local SGD.

\end{itemize}

\section{Related Work} 
Decentralized optimization algorithms can be tracked back to \cite{tsitsiklis1986distributed}. After that, decentralized optimization has been intensively studied in signal processing and control community. Decentralized gradient descent (DGD) \cite{nedic2009distributed}, diffusion \cite{chen2012diffusion} and dual averaging \cite{duchi2011dual} are among the first decentralized algorithms that target on general optimization problems. However, these algorithms suffer from a bias caused by data heterogeneity \cite{yuan2016convergence}. Various primal-dual algorithms are proposed to overcome this issue, and they are based on alternating direction method of multipliers (ADMM) \cite{shi2014linear}, explicit bias-correction \cite{shi2015extra, yuan2017exact1,li2017decentralized}, gradient tracking \cite{xu2015augmented, di2016next, nedic2017achieving, qu2018harnessing}, coordinate-descent methods \cite{he2018cola}, and dual acceleration \cite{scaman2017optimal, scaman2018optimal, uribe2020dual}.

In the context of machine learning, decentralized SGD, also known as Gossip SGD, have gained a lot of attention recently. \cite{lian2017can} first proves Gossip SGD can reach the same linear speedup as vanilla parallel SGD.  After that, \cite{assran2019stochastic} comes out to extend Gossip SGD to directed topology. 
A recent work \cite{koloskova2020unified} proposes a unified framework to analyze algorithms with changing topology and local updates. While it covers Gossip-PGA as a special form, the theoretical and practical benefits of periodic global averaging were not studied therein. The data heterogeneity issue suffered in Gossip SGD is discussed and addressed in \cite{tang2018d, yuan2020influence,lu2019gnsd,xin2020improved}. Gossip SGD is also extended to asynchronous scenarios in \cite{lian2018asynchronous, luo2020prague}. 

Local SGD can be traced back to \cite{zinkevich2010parallelized} which proposed a one-shot averaging. More frequent averaging strategy is proposed in \cite{zhang2016parallel}, and the convergence property of Local SGD is established in \cite{yu2019parallel,stich2019local,bayoumi2020tighter}. Local SGD is also widely-used in federated learning \cite{mcmahan2017communication, li2019convergence}.

Another closely related work \cite{wang2019slowmo} proposes a slow momentum (SlowMo) framework, where each node, similar to the Gossip-PGA algorithm proposed in this paper, periodically synchronizes across the network and performs a momentum update. The analysis in SlowMo cannot cover the convergence results in this paper due to its data-homogeneous setting. In addition, we will clarify some new questions such as how much acceleration can PGA bring to Gossip and Local SGDs, and how to adjust the averaging period effectively.

Various techniques can be integrated to Gossip SGD to improve its communication efficiency. This paper does not consider quantization \cite{alistarh2017qsgd, bernstein2018signsgd}, gradient compression \cite{tang2019doublesqueeze, koloskova2019decentralized, koloskova2019decentralized2} and lazy communication \cite{chen2018lag, liu2019communication}, but these orthogonal techniques can be added to our 
methods.

\section{Gossip SGD with Periodic Global Average}
  
Assume all computing nodes are connected over a graph $\cG=\{\cV, \cE\}$ where $\cV=\{1,2,\cdots, n\}$ denote the node index and $\cE$ denote the communication links between all nodes. Similar to existing decentralized algorithms \cite{nedic2009distributed,chen2012diffusion,lian2017can,assran2019stochastic}, information exchange in the gossip step is only allowed to occur between connected neighbors. To characterize the decentralized communication, we let $W\in \RR^{n \times n}$ be a doubly stochastic matrix, i.e., $W \ge 0$, $W\mathds{1}_n = \mathds{1}_n$ and $\mathds{1}_n^T W = \mathds{1}_n^T$. The $(i,j)$-th element $w_{ij}$ is the weight to scale information flowing from node $j$ to node $i$. If nodes $i$ and $j$ are not neighbors then $w_{ij}=0$, and if they are neighbors or
identical then the weight $w_{ij}> 0$. Furthermore, we define $\cN_i$
as the set of neighbors of node $i$ which also includes node $i$ itself. 

 \begin{algorithm}[t]
  \DontPrintSemicolon
  \KwRequire{Initialize learning rate $\gamma>0$, weight matrix $W$, global averaging period $H$, and let each $\x^{(0)}_{i}$ to be equivalent to each other.}

  \For{$k=0, 1,2,...,T-1$, every node $i$}{
    Sample $\bxi^{(k\hspace{-0.3mm}+\hspace{-0.3mm}1)}_{i}$, update $\g_i^{(k)} \hspace{-0.8mm}=\hspace{-0.8mm} {\nabla}F_i(\x^{(k)}_{i}\hspace{-0.4mm};\bxi^{(k\hspace{-0.3mm}+\hspace{-0.3mm}1)}_{i})$ \;
    
    $\x^{(k+\frac{1}{2})}_i = \x^{(k)}_{i} - \gamma \g_i^{(k)}\hspace{1.1cm} \triangleright \mbox{\footnotesize{Local SGD update}}$ \;
    
    \If{$\mathrm{mod}(k+1, H)=0$}{
      $\x^{(k+1)}_i  = \frac{1}{n}\sum_{j=1}^n \x^{(k+\frac{1}{2})}_j \quad \ \  \triangleright  \mbox{\footnotesize{global average}}$\;}
    \Else{
      $\x^{(k+1)}_i  = \sum_{j\in \cN_i} w_{ij} \x^{(k+\frac{1}{2})}_j \ \ \triangleright \mbox{\footnotesize{one gossip step}}$\;}
  }
  \caption{Gossip-PGA}
  \label{Algorithm: Gossip-PGA}
 \end{algorithm}

The Gossip-PGA algorithm is listed in Algorithm \ref{Algorithm: Gossip-PGA}. In the gossip step, every node $i$ collects information from all its {\em connected} neighbors. For global average step, nodes synchronize their model parameters using the efficient All-Reduce primitives. When $H\to \infty$, Gossip-PGA will reduce to standard Gossip SGD; when $W = \frac{1}{n}\mathds{1}\mathds{1}_n$, Gossip-PGA will reduce to vanilla parallel SGD; when $W = I$, Gossip-PGA will reduce to Local SGD.

\noindent \textbf{All-Reduce v.s. multiple Gossips.} In a computing cluster with $n$ nodes, global averaging is typically conducted in an efficient Ring All-Reduce manner, rather than via multiple gossip steps as in \cite{berahas2018balancing}. The communication time comparison between a single gossip and Ring All-Reduce step is listed in Appendix \ref{app-all-reduce}. In the one-peer exponential network, the exact global average can be achieved via $\ln(n)$ gossip communications, which generally takes more wall-clock time than a single Ring All-Reduce operation. Therefore, we recommend exploiting All-Reduce to conduct global averaging in Gossip-PGA.

\noindent \textbf{Data-center v.s. wireless network.} This paper considers deep training within high-performance data-center clusters, in which all GPUs are connected with high-bandwidth channels and the network topology can be fully controlled. Under such setting, the periodic global averaging conducted with Ring All-Reduce has tolerable communication cost, see Appendix \ref{app-all-reduce}. For scenarios where global averaging is extremely expensive to conduct such as in wireless sensor network, the global averaging can be  approximated via multiple gossip steps, or may not be recommended.

\subsection{Assumptions and analysis highlights}
\label{sec-ass-and-highlights}
We now establish convergence rates for Gossip-PGA on  smooth convex and non-convex problems. For all our theoretical results we make the following standard assumptions.
  
\begin{assumption}[\sc $L$-smoothness]\label{ass:smoothness}
  Each local cost function $f_i(x)$ is differentiable, and there exists a constant $L$ such that for each $\x, \y\in\RR^d$:
  \begin{align}
  \|\nabla f_i(\x) - \nabla f_i(\y)\| \le L \|\x - \y\|. \label{smooth-1}
  \end{align}
\end{assumption}
  
\begin{assumption}[\sc Gradient noise]\label{ass:gradient-noise} Recall $\g_i^{(k)}$ is the stochastic gradient noise defined in line 2 of Algorithm 1. It is assumed that for any $k$ and $i$ that 
\begin{align}
\mathbb{E}[\g_i^{(k)} - \nabla f_i(\x_i^{(k)})|\cF^{(k-1)}] &= 0, \label{gd-1}\\
\mathbb{E}[\|\g_i^{(k)} - \nabla f_i(\x_i^{(k)})\|^2|\cF^{(k-1)}] &\le \sigma^2 \label{gd-2}
\end{align}
for some constant $\sigma^2 > 0$. Moreover, we assume $\bxi_i^{(k)}$ is independent of each other for any $k$ and $i$. Filtration is defined as $\cF^{(k)} \hspace{-0.8mm}=\hspace{-0.8mm} \big\{\hspace{-0.8mm} \{\hspace{-0.4mm}\x_i^{(k)}\hspace{-0.4mm}\}_{i=1}^n,  \{\hspace{-0.4mm}\bxi_i^{(k)}\hspace{-0.4mm}\}_{i=1}^n, \cdots, \{\hspace{-0.4mm}\x_i^{(0)}\hspace{-0.4mm}\}_{i=1}^n,  \{\hspace{-0.4mm}\bxi_i^{(0)}\hspace{-0.4mm}\}_{i=1}^n\hspace{-0.8mm} \big\}$
\end{assumption}

\begin{assumption}[\sc Weighting matrix]\label{ass:weight-matrix}
The network is strongly connected and the weight matrix $W$ satisfies 
$W\mathds{1}_n = \mathds{1}_n, \ \mathds{1}_n^T W = \mathds{1}_n^T, \ \mathrm{null}(I-W) = \mathrm{span}(\mathds{1}_n).$ 
We also assume $\|W - \frac{1}{n}\mathds{1}\mathds{1}^T\|_2 \le \beta$ for some $\beta \in (0,1)$.
\end{assumption}
\begin{remark}\label{remark-beta-definition}
Quantity $\beta \in (0, 1)$ indicates how well the topology is connected. Smaller $\beta$ indicates better-connected network while larger $\beta$ implies worse-connected topology.
\end{remark} 

\textbf{Analysis highlights.} To derive the influence of periodic global averaging, we have to exploit all useful algorithm structures to achieve its superiority. These structures are:
\begin{itemize}
    \item $\x_i^{(k)} = \bar{\x}^{(k)}$ when $\mod(k,H)=0$. This structure relieves the influence of network topology;
    \item Gossip communications within each period also contribute to consensus among nodes. This structure is crucial to establish superiority to Local SGD;
    \item When network is large and sparse, i.e., $H < \frac{1}{1-\beta}$, the global averaging is more critical to drive consensus. This structure is crucial to establish superiority to Gossip SGD when $H < \frac{1}{1-\beta}$.
    \item When network is small or dense, i.e., $H > \frac{1}{1-\beta}$, gossip communication is more critical to drive consensus. This structure is crucial to establish superiority to Gossip SGD when $H > \frac{1}{1-\beta}$. 
\end{itemize}
Ignoring any of the above structures in the analysis will result in incomplete conclusions on comparison among Gossip-PGA, Gossip SGD and Local SGD.

\begin{table*}[t!]
\begin{center}
\begin{sc}
\begin{tabular}{lcc}
\toprule
& Gossip SGD \cite{koloskova2020unified}    
& Gossip-PGA        
\\ \midrule
Rates (General form)
& $O\Big(\frac{\sigma}{\sqrt{n T}} \hspace{-0.8mm}+\hspace{-0.8mm} \frac{\beta^{\frac{2}{3}}\sigma^{\frac{2}{3}}}{ T^{\frac{2}{3}}\boldsymbol{(1-\beta)^{\frac{1}{3}}}} \hspace{-0.8mm}+\hspace{-0.8mm} \frac{\beta^{\frac{2}{3}} {b}^{\frac{2}{3}}}{ T^{\frac{2}{3}}\boldsymbol{(1-\beta)^{\frac{2}{3}}}} \hspace{-0.8mm}+\hspace{-0.8mm} \frac{\beta}{\boldsymbol{(1-\beta)} T} \Big)$ 
& $O\Big( \frac{\sigma}{\sqrt{nT}} \hspace{-0.8mm}+\hspace{-0.8mm} \frac{\boldsymbol{C_\beta^{\frac{1}{3}}}\beta^{\frac{2}{3}}\sigma^{\frac{2}{3}}}{T^{\frac{2}{3}}} \hspace{-0.8mm}+ \hspace{-0.8mm}\frac{\boldsymbol{C_\beta^{\frac{1}{3}} D_\beta^{\frac{1}{3}}} \beta^{\frac{2}{3}} {b}^{\frac{2}{3}}}{T^{\frac{2}{3}}} \hspace{-0.8mm}+\hspace{-0.8mm} \frac{\beta \boldsymbol{D_\beta}}{T}\Big)$ 
\vspace{1mm}\\

Rates (when $\frac{1}{1-\beta} < H$)   

& $O\Big(\frac{\sigma}{\sqrt{n T}} \hspace{-0.8mm}+\hspace{-0.8mm} \frac{\beta^{\frac{2}{3}} \sigma^{\frac{2}{3}}}{ T^{\frac{2}{3}}\boldsymbol{(1-\beta)^{\frac{1}{3}}}} \hspace{-0.8mm}+\hspace{-0.8mm} \frac{\beta^{\frac{2}{3}} {b}^{\frac{2}{3}}}{ T^{\frac{2}{3}}\boldsymbol{(1-\beta)^{\frac{2}{3}}}} \hspace{-0.8mm}+\hspace{-0.8mm} \frac{\beta}{(1-\beta) T} \Big)$ 
& $O\Big( \frac{\sigma}{\sqrt{nT}} \hspace{-0.8mm}+\hspace{-0.8mm} \frac{\boldsymbol{C_\beta^{\frac{1}{3}}}\beta^{\frac{2}{3}}\sigma^{\frac{2}{3}}}{T^{\frac{2}{3}}} \hspace{-0.8mm}+ \hspace{-0.8mm}\frac{\boldsymbol{C_\beta^{\frac{1}{3}}} \beta^{\frac{2}{3}}{b}^{\frac{2}{3}}}{\boldsymbol{(1-\beta)^{\frac{1}{3}}}T^{\frac{2}{3}}} \hspace{-0.8mm}+\hspace{-0.8mm} \frac{\beta}{(1-\beta)T}\Big)$ \vspace{1mm}\\

Rates (when $\frac{1}{1-\beta} \ge H$)    

& $O\Big(\frac{\sigma}{\sqrt{n T}} \hspace{-0.8mm}+\hspace{-0.8mm} \frac{\beta^{\frac{2}{3}} \sigma^{\frac{2}{3}}}{ T^{\frac{2}{3}}\boldsymbol{(1-\beta)^{\frac{1}{3}}}} \hspace{-0.8mm}+\hspace{-0.8mm} \frac{\beta^{\frac{2}{3}} {b}^{\frac{2}{3}}}{ T^{\frac{2}{3}}\boldsymbol{(1-\beta)^{\frac{2}{3}}}} \hspace{-0.8mm}+\hspace{-0.8mm} \frac{\beta}{(1-\beta) T} \Big)$ 
& $O\Big( \frac{\sigma}{\sqrt{nT}} \hspace{-0.8mm}+\hspace{-0.8mm} \frac{\boldsymbol{C_\beta^{\frac{1}{3}}} \beta^{\frac{2}{3}} \sigma^{\frac{2}{3}}}{T^{\frac{2}{3}}} \hspace{-0.8mm}+ \hspace{-0.8mm}\frac{\boldsymbol{C_\beta^{\frac{1}{3}} H^{\frac{1}{3}}} \beta^{\frac{2}{3}} {b}^{\frac{2}{3}}}{T^{\frac{2}{3}}} \hspace{-0.8mm}+\hspace{-0.8mm} \frac{\beta \boldsymbol{H}}{T}\Big)$ \vspace{1mm}

\\ \bottomrule
\end{tabular}
\end{sc}
\end{center}
\vskip -0.1in
\caption{Convergence rate comparison between Gossip SGD and Gossip-PGA for smooth convex/non-convex problems. We use notation $b^2$ to indicate the data heterogeneity for both convex and non-convex scenarios. }
\label{table-comparison-with-Gossip-SGD}
\end{table*}

\subsection{Convergence analysis: convex scenario}
\begin{assumption}[\sc convexity]\label{ass-strongly-convex}
Each $f_i(x)$ is convex. 
\end{assumption}

\begin{definition}[\sc Data heterogeneity]
When each $f_i(x)$ is convex, we let $b^2=\frac{1}{n}\sum_{i=1}^n\|\nabla f_i(x^\star)\|^2$ denote the data heterogeneity. 
\end{definition}
When each local data follows the same distribution, it holds that $f_i(x) = f(x) \ \forall i$ and hence $\nabla f_i(x^\star) = \nabla f(x^\star) =0$ which also implies $b^2=0$. With Assumption \ref{ass-strongly-convex}, we let $x^\star$ be one of the global solutions to problem \eqref{eq:general-prob}. 
\begin{theorem}\label{thm-convex}
Under Assumptions  \ref{ass:smoothness}--\ref{ass-strongly-convex}, if $\gamma$ is chosen as
\begin{align}\label{xcnxcnxcnxcxcnxcxncsd8}
\hspace{-2mm}\gamma \hspace{-0.5mm}=\hspace{-0.5mm} \min\hspace{-0.5mm}\Big\{\hspace{-0.5mm}\frac{1}{12\beta L D_\beta}, \left(\hspace{-0.5mm}\frac{r_0}{r_1(T+1)}\right)^{\frac{1}{2}}\hspace{-2mm}, \left(\frac{r_0}{r_2(T+1)}\right)^{\frac{1}{3}}\hspace{-1mm}\Big\} 
\end{align}
with constants $r_0 = 2 \mathbb{E}\|\bar{\x}^{(0)} - x^\star\|^2, r_1 = 2\sigma^2/n$, and $r_2 = 6L\beta^2 C_\beta \sigma^2 + 18L\beta^2 C_\beta D_\beta$, 
it holds for any $T$ that 
\begin{align}
& \mathbb{E} f(\hat{\x}^{(T)}) - f(x^\star) \nonumber \\
=&\ {O}\Big(\frac{\sigma}{\sqrt{nT}} \hspace{-0.5mm}+ \hspace{-0.5mm}
\frac{C_\beta^{\frac{1}{3}}\beta^{\frac{2}{3}}(\sigma^{\frac{2}{3}} + D_\beta^{\frac{1}{3}} {b}^{\frac{2}{3}})}{T^{\frac{2}{3}}} + \frac{\beta D_\beta}{T}\Big) \label{sc-convergence}
\end{align}
where $\bar{\x}^{(k)}=\frac{1}{n}\sum_{i=1}^n\x_i^{(k)}$, $\hat{\x}^{(T)} = \frac{1}{T+1}\sum_{k=0}^T  \bar{\x}^{(k)}$, 
$C_\beta = \sum_{k=0}^{H-1}\beta^k$ and $D_\beta = \min\{H, 1/(1-\beta)\}$. (Proof is in Appendix \ref{app-sc-descent}.)
\end{theorem}

\begin{remark}
When $\beta \to 0$, i.e., the network tends to be fully connected, Gossip-PGA will converge at rate $O(\sigma/\sqrt{nT})$, which recovers the rate of parallel SGD.
\end{remark}
\begin{remark}
When $\beta \to 1$, i.e., the information exchange via gossip communication is inefficient, it holds that $C_\beta \to H$ and $D_\beta = \min\{H, 1/(1-\beta)\} = H$. Substituting them to \eqref{sc-convergence} will recover the rate of Local SGD, see Table \ref{table-comparison-with-Gossip-local}.
\end{remark}
\begin{remark}
When $H \to \infty$, i.e., the networked agents tend not to conduct global synchronization, it holds that $C_\beta  \to 1/(1-\beta)$ and $D_\beta  = \frac{1}{1-\beta}$.
Substituting these values to \eqref{sc-convergence} will recover the rate of Gossip SGD, see Table \ref{table-comparison-with-Gossip-SGD}.
\end{remark}

\subsection{Convergence analysis: non-convex scenario}
\noindent We first introduce an assumption about data heterogeneity specifically for non-convex problems:
\begin{assumption}[\sc Data heterogeneity]\label{ass:data-h}
There exists constant $\hat{b} > 0$ such that $\frac{1}{n}\sum_{i=1}^n \|\nabla f_i(\x) - \nabla f(\x)\|^2 \le \hat{b}^2$ for any $x\in \RR^d$. 
If local data follows the same distribution, it holds that $\hat{b}=0$.
\end{assumption}

\begin{theorem}\label{thm:nc}
Under Assumptions \ref{ass:smoothness}--\ref{ass:weight-matrix} and \ref{ass:data-h}, if $\gamma$ satisfies the condition \eqref{xcnxcnxcnxcxcnxcxncsd8} (replace $b^2$ with $\hat{b}^2$ and use $r_0 = 4 \mathbb{E}f(\bar{\x}^{(0)})$), it holds for any $T > 0$ that
\begin{align}\label{2bnsdb}
&\ \frac{1}{T+1}\sum_{k=0}^T \mathbb{E}\|\nabla f(\bar{\x}^{(k)})\|^2 \nonumber \\
=&\ {O}\Big(\frac{\sigma}{\sqrt{nT}} \hspace{-0.5mm}+ \hspace{-0.5mm}
\frac{C_\beta^{\frac{1}{3}}\beta^{\frac{2}{3}}(\sigma^{\frac{2}{3}} + D_\beta^{\frac{1}{3}} {b}^{\frac{2}{3}})}{T^{\frac{2}{3}}} + \frac{\beta D_\beta}{T}\Big)
\end{align}
where $\bar{\x}^{(k)}=\frac{1}{n}\sum_{i=1}^n\x_i^{(k)}$.  (Proof is in Appendix \ref{app-convg-nc}.)
\end{theorem}

\subsection{Comparison with Gossip SGD}

To better illustrate how periodic global averaging helps relieve the affects of network topology in Gossip SGD, we list convergence rates of Gossip SGD and Gossip-PGA for smooth convex or non-convex problems in Table \ref{table-comparison-with-Gossip-SGD}. The first line is the general rate expression for both algorithms. In the second line we let $D_\beta = \min\{H, 1/(1-\beta)\} = 1/(1-\beta)$ for Gossip-PGA, and in the third line we let $D_\beta = H$. According to this table, we derive the transient stages of Gossip SGD and Gossip-PGA for each scenarios (i.e., large/small network, iid/non-iid data distributions) in Table \ref{table-transient-stage} (see the derivation detail in Appendix \ref{app-tran-time}). As we have explained in Main Results subsection in the introduction, it is observed from Tables \ref{table-transient-stage} and \ref{table-comparison-with-Gossip-SGD} that: (i) Gossip-PGA always converges faster (or has shorter transient stages) than Gossip SGD for any $\beta$ and $H$ value. (ii) Such superiority gets evident for large and sparse networks where $\beta \to 1$.

\begin{table}[t]
\begin{center}
\begin{small}
\begin{sc}
\begin{tabular}{lcc}
\toprule
& Gossip SGD & Gossip-PGA  \\
\midrule
Transient iter. & $O(n^7)$                     & $O(n^5)$                           \\

Single comm.    & $O(\theta d + \alpha)$       & $O(\theta d + \sqrt{n}\alpha)$     \\  
\hline
Transient time      & $O(n^7\theta d + n^7\alpha)$ & $O(n^5 \theta d + n^{5.5} \alpha)$ \\ \bottomrule
\end{tabular}
\end{sc}
\end{small}
\end{center}
\vskip -0.1in
\caption{Transient time comparison between non-iid Gossip SGD and Gossip-PGA over the specific grid ($1-\beta = O(1/n)$) topology. We choose $H=\sqrt{n}$ as the period in Gossip-PGA.}
\label{table-specfic-example}
\vspace{-5mm}
\end{table}

\begin{remark} \label{remark-comparison-with-jaggi}
The convergence analysis in topology-changing Gossip SGD \citep{koloskova2020unified} covers Gossip-PGA.
By letting $p=1$ and $\tau=H$ in Theorem~2 of  \citep{koloskova2020unified}, it is derived that Gossip-PGA has a transient stage on the order of $\Omega(n^3 H^4)$ for {non-convex non-iid scenario.} Such transient stage cannot quantify the superiority to Gossip and Local SGDs. In fact, it may even show PGA can do harm to Gossip SGD when $H > \frac{1}{1-\beta}$, which is counter-intuitive. This is because \citep{koloskova2020unified} is for the general time-varying topology. It does not utilize the structures listed in Sec.~\ref{sec-ass-and-highlights}.
\end{remark}

{\noindent \textbf{Transient stage in runtime}. Table \ref{table-transient-stage} compares transient stages between Gossip-PGA and Gossip SGD in {\em iterations}. But what people really care about in practice is {\em runtime}. Since  both Gossip SGD and Gossip-PGA have the same computational overhead per iteration, we will focus on communication time spent in the transient stage.

Given the bandwidth in a computing cluster with size $n$, we let $\alpha$ denote the point-to-point
latency in the network, and $\theta$ denote the communication time cost to transmit a scalar variable. Since variable $x$ in problem \eqref{eq:general-prob} has dimension $d$, it will take $\theta d$ time to transmit $x$ between two nodes. Under this setting, the All-Reduce global averaging step will take $2\theta d + n\alpha = O(\theta d + n\alpha)$ time (see section 2.5 in \cite{ben2019demystifying}). The gossip-style communication time varies with different network topologies. For the commonly-used ring or grid topology, it takes $|\cN_i| \theta d + \alpha = O(\theta d + \alpha)$ for one gossip communication, where $|\cN_i|$ is the neighborhood size of node $i$, and $|\cN_i|=3$ for the ring and $5$ for the grid. As to Gossip-PGA, if we amortize the periodic All-Reduce cost into each communication, it will have $|\cN_i| \theta d + \alpha + (2\theta d + n\alpha)/H = O(\theta d + \sqrt{n}\alpha)$ when we set $H = \sqrt{n}$. With the formula 
$
 \mbox{Total time} = \mbox{transient stage (in iteration)} \times \mbox{comm. per iter.}    $
We calculate and compare the transient time between non-iid Gossip-PGA and Gossip-SGD (over the grid topology) in Table \ref{table-specfic-example}. Other comparisons for iid scenario or the ring topology can be found in Appendix \ref{app-tran-time}. It is observed in all tables that Gossip-PGA has shorter transient time.}

\begin{table}[t!]
\begin{center}
\begin{sc}
\begin{tabular}{rc}
\toprule
& Rates \\
\midrule
{L-SGD} & $O\Big( \frac{\sigma}{\sqrt{nT}} \hspace{-0.8mm}+\hspace{-0.8mm} \frac{\boldsymbol{H^{\frac{1}{3}}}\sigma^{\frac{2}{3}}}{T^{\frac{2}{3}}} \hspace{-0.8mm}+ \hspace{-0.8mm}\frac{\boldsymbol{H^{\frac{2}{3}}}{b}^{\frac{2}{3}}}{T^{\frac{2}{3}}} \hspace{-0.8mm}+\hspace{-0.8mm} \frac{H}{T}\Big)$ \\
{G-PGA} & $O\Big( \frac{\sigma}{\sqrt{nT}} \hspace{-0.8mm}+\hspace{-0.8mm} \frac{\boldsymbol{C_\beta^{\frac{1}{3}}\beta^{\frac{2}{3}}}\sigma^{\frac{2}{3}}}{T^{\frac{2}{3}}} \hspace{-0.8mm}+ \hspace{-0.8mm}\frac{\boldsymbol{C_\beta^{\frac{1}{3}} H^{\frac{1}{3}} \beta^{\frac{2}{3}} {b}^{\frac{2}{3}}}}{T^{\frac{2}{3}}} \hspace{-0.8mm}+\hspace{-0.8mm} \frac{\beta H}{T}\Big)$ \\ 
\bottomrule
\end{tabular}
\end{sc}
\end{center}
\caption{Convergence rate comparison between Local SGD (L-SGD) and Gossip-PGA (G-PGA) over smooth convex/non-convex problems. The rate for Local SGD is from \cite{koloskova2020unified,yu2019parallel,li2019communication}. }
\label{table-comparison-with-Gossip-local}
\vspace{-5mm}
\end{table}

\subsection{Comparison with Local SGD}
The convergence rates of Gossip-PGA and Local SGD are listed in Table \ref{table-comparison-with-Gossip-local}, from which we derive the transient stages of them in Table \ref{table-transient-stage-local} (details are in Appendix \ref{app-tran-time}). As we have explained in the introduction, it is observed from Tables \ref{table-transient-stage-local} and \ref{table-comparison-with-Gossip-local} that (i) Gossip-PGA always converges faster (or has shorter transient stages) than Local SGD for any $\beta$ and $H$ value, and (ii) Such superiority gets more evident for well-connected network where $\beta \to 0$.

As to the wall-clock transient time of Local SGD, if we amortize the periodic All-Reduce cost into each local update, it will take $(2\theta d + n \alpha)/H = O(\theta d/H + n \alpha/H)$ communication time per iteration. Using the transient iteration derived in Table \ref{table-transient-stage-local}, the total transient time for Local SGD (non-iid scenario) will be $O(n^3H^3(\theta d + n\alpha))$. Comparing it with the total transient time $O(n^3 H C_\beta^2  \beta^4(H \theta d + n\alpha))$ for Gossip-PGA, we find Gossip-PGA always has shorter transient runtime for a large $H > \beta^4 C_\beta^2$.

\begin{remark}
While we discuss in detail that the transient time of Gossip-PGA is shorter than Gossip and Local SGDs, it is worth noting that the communication time during the linear speedup stage (i.e., after the transient stage) also contributes to the total training time. In this stage, Gossip-PGA is less efficient due to its periodic global averaging. However, we illustrate that Gossip-PGA is always endowed with shorter total training time than Gossip and Local SGDs with extensive deep learning experiments in Sec.~\ref{sec:experiments}.
\end{remark}

\section{Gossip SGD with Adaptive Global Average}
Gossip-PGA suffers from the burden of tuning $H$ by hand. A small $H$ will incur more communication overhead while a large value can slow down the convergence. We further propose Gossip-AGA, an adaptive extension of Gossip-PGA.

\vspace{1mm}
\noindent \textbf{Intuition}. A small consensus variance $\sum_{i=1}^n \mathbb{E}\|\x_i - \bar{\x}\|^2$  would  accelerate Gossip-PGA. To see that, if $\sum_{i=1}^n \mathbb{E}\|\x_i - \bar{\x}\|^2 = 0$ for each iteration, then Gossip-PGA reduces to parallel SGD and can reach its fastest convergence. Recall from Lemma \ref{lm-weighted-consensus-nc} in the appendix that the averaged consensus $\frac{1}{T+1}\sum_{k=0}^T \mathbb{E}\|\vx^{(k)} - \bar{\vx}^{(k)}\|^2$ is bounded by $\frac{d_1 \gamma^2}{T+1} \sum_{k =0}^T \mathbb{E}\| \nabla f(\bar{\x}^{(k)})\|^2 \hspace{-0.8mm}+\hspace{-0.8mm} d_2 \gamma^2$
where $d_1$ and $d_2$ are constants. It is observed that the initial consensus variance (when $T$ is small) can be significant due to large $\gamma$ and $\mathbb{E}\| \nabla f(\bar{\x}^{(k)})\|^2$. In the later stage when $T$ is sufficiently large, both the diminishing step-size $\gamma$ and gradient $\mathbb{E}\| \nabla f(\bar{\x}^{(k)})\|^2$ go to $0$ and hence leading to a small consensus variance naturally. With these observations, it is intuitive to take global synchronizations more frequently in initial stages to reduce the overall consensus variance. 

\vspace{1mm}
\noindent \textbf{Convergence}. We denote $H^{(\ell)}$ as the duration of the $\ell$-th period. The following corollary establishes convergence for Gossip-PGA with any  time-varying but finite global averaging period sequence $\{H^{(\ell)}\}$:

\begin{corollary}\label{thm:nc-adaptive}
Suppose Assumptions \ref{ass:smoothness}--\ref{ass:weight-matrix} and \ref{ass:data-h} hold and the time-varying period $H^{(\ell)}$ is upper bounded by $H_{\rm max} = \max_{\ell \ge 0}\{H^{(\ell)}\}$. If $\gamma$ satisfies the condition in Theorem \ref{thm-convex} with $H = H_{\rm max}$, then Gossip-AGA converges at rate \eqref{2bnsdb} in which $H$ is replaced by $H_{\rm max}$. (Proof is in Appendix \ref{app-corollary-1}.)
\end{corollary}

\vspace{1mm}
\noindent \textbf{Adaptive Strategy}. This subsection will propose an adaptive strategy that is inspired by \cite{wang2019adaptive}. If we recover the influence of the initial value $F_0 = \mathbb{E}f(\bar{\x}^{(0)})$ on convergence rate \eqref{2bnsdb}, Gossip-PGA for non-convex problems will converge at  
\begin{align} 
{O}\Big(\frac{\sigma F_0^{\frac{1}{2}}}{\sqrt{nT}} \hspace{-0.5mm}+ \hspace{-0.5mm}
\frac{H^{\frac{1}{3}}\beta^{\frac{2}{3}}\sigma^{\frac{2}{3}}F_0^{\frac{2}{3}}}{T^{\frac{2}{3}}} +
 \frac{H^{\frac{2}{3}} \beta^{\frac{2}{3}}\hat{b}^{\frac{2}{3}}F_0^{\frac{2}{3}}}{T^{\frac{2}{3}}} \hspace{-0.5mm}+ \hspace{-0.5mm} \frac{\beta D_\beta F_0}{T}\Big). \nonumber
\end{align}
For a fixed $T$, a period $H = {\sigma^{\frac{3}{2}} T^{\frac{1}{4}}}/(\beta \hat{b} F_0^{\frac{1}{4}} n^{\frac{3}{4}})$
will guarantee the linear speedup.  Therefore, the initial period $H^{(0)}$ can be chosen as $H^{(0)} = d_1/[\mathbb{E}f(\bar{\x}^{(0)})]^{\frac{1}{4}}$ for some constant $d_1$. Similarly, for the $\ell$-th period, workers can be viewed as restarting training at a new initial point $\bar{\x}^{(T_{\ell-1})}$ where $T_{\ell-1} = H^{(0)} + \cdots + H^{(\ell-1)}$. As a result, the $\ell$-th period $H^{(\ell)}$ can be chosen as $H^{(\ell)} = d_1/[\mathbb{E}f(\bar{\x}^{(T_{\ell-1})})]^{\frac{1}{4}}$. With such 
choice of $H^{(0)}$ and $H^{(\ell)}$, it is not difficult to have  
\begin{align}\label{xnsdnsdnsdnds}
H^{(\ell)} = \Big(\frac{\mathbb{E}f(\bar{\x}^{(0)})}{\mathbb{E}f(\bar{\x}^{(T_{\ell - 1})})}\Big)^{\frac{1}{4}} H^{(0)}.
\end{align}
Since $\mathbb{E}f(\bar{\x}^{(k)})$ will decrease as $k$ increases, \eqref{xnsdnsdnsdnds} will generate an  increasing sequence of period $H^{(\ell)}$. We list Gossip-AGA as Algorithm 2 in Appendix \ref{app-implem-AGA} and elaborate on implementation details there.

\section{Experimental Results}\label{sec:experiments}
\begin{figure*}[t]
\begin{center}
\centerline{\includegraphics[width=0.95\textwidth]{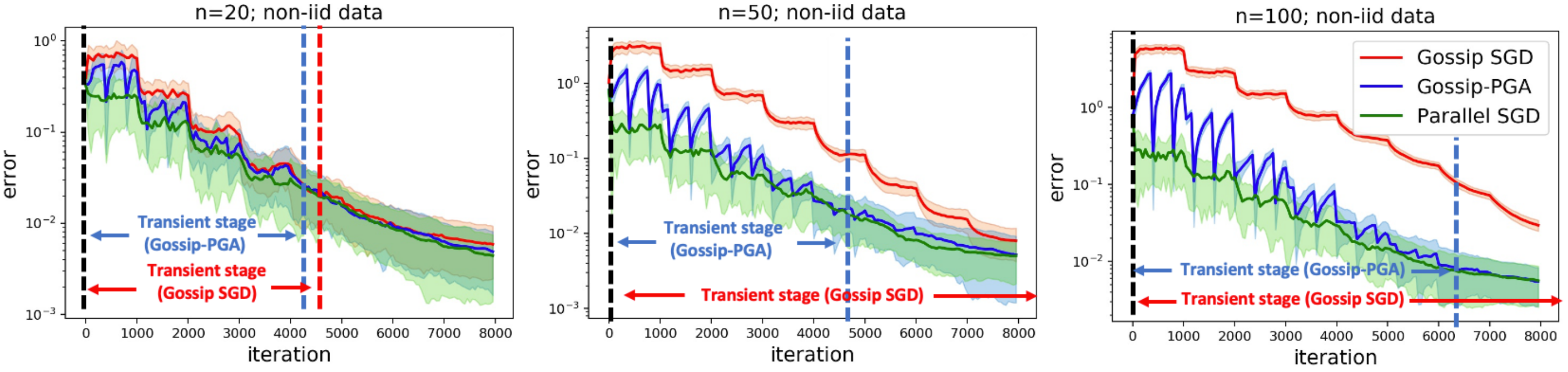}}
\vspace{-4mm}
\caption{Convergence comparison between Gossip-PGA, Gossip and parallel SGDs on the logistic regression problem over ring topology. The transient stage is determined by counting iterations before an algorithm exactly matches the convergence curve of Parallel SGD. {Note that the transient stage for Gossip SGD in the middle and right sub-figures is beyond the plotting canvas.}}
\label{Fig:convex-comparison-gossip}
\end{center}
\end{figure*}

\begin{figure*}[t]
\vskip -0.1in
\begin{center}
\centerline{\includegraphics[width=0.42\textwidth]{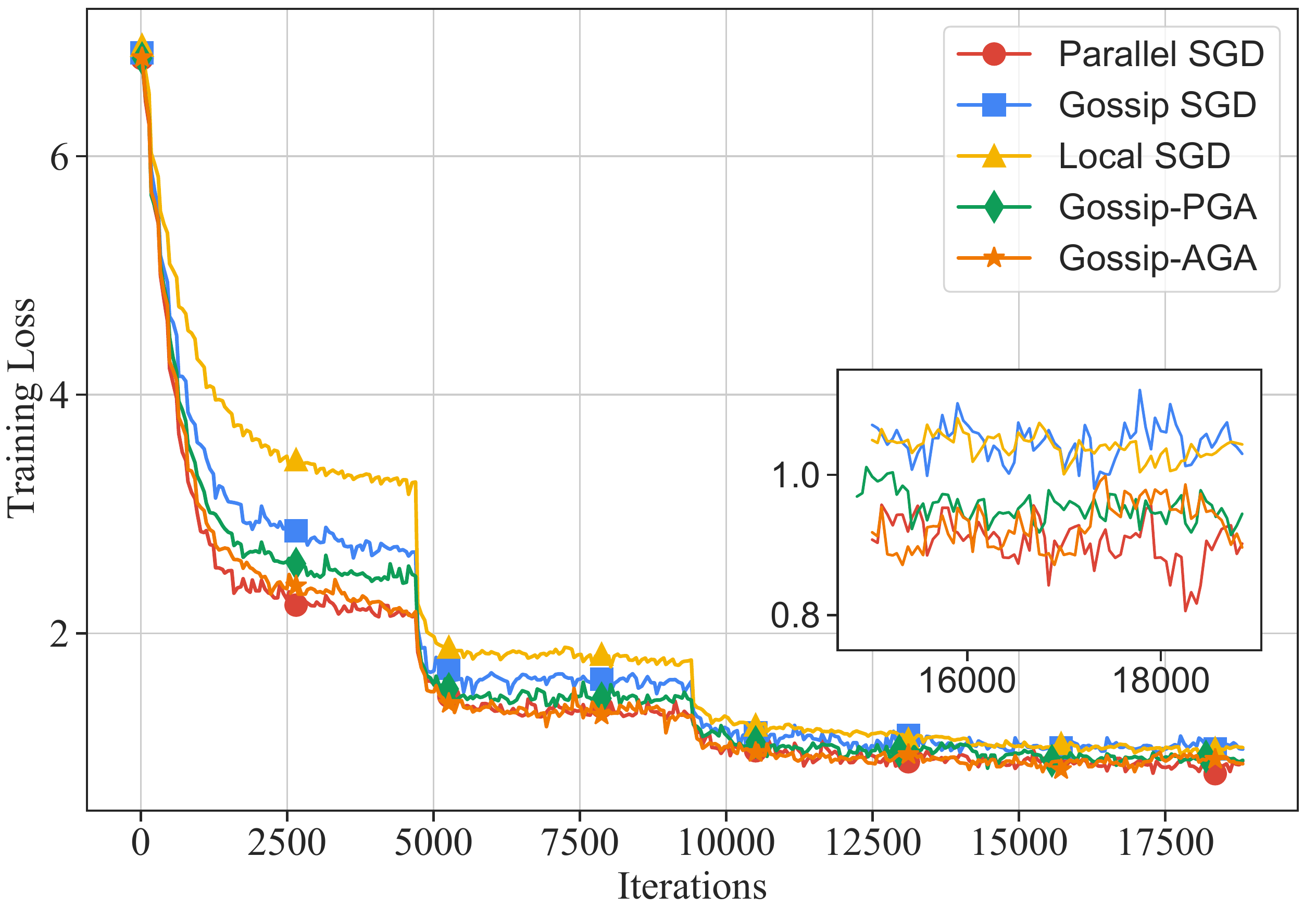} 
\includegraphics[width=0.42\textwidth]{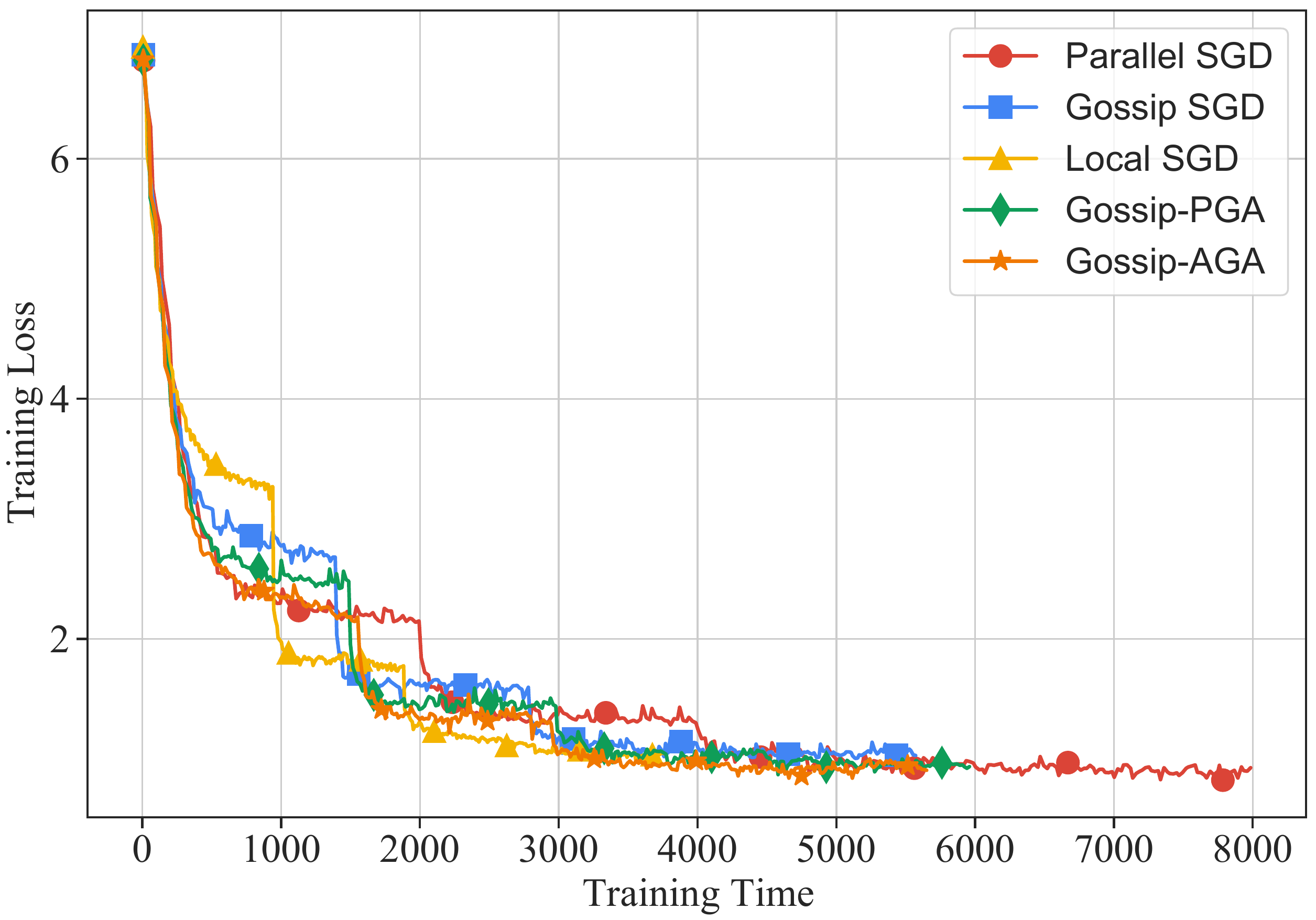}}
\vskip -0.15in
\caption{Convergence results on the ImageNet in terms of iteration and runtime. More results are in Appendix \ref{app-more-exp-on-imagenet}.}
\vskip -0.15in
\label{Fig:imagenet}
\end{center}
\end{figure*}

In this section, we first examine how the transient stage differs for Gossip-PGA, Gossip and Local SGDs on networks with different topology and size on convex logistic regression. Next, we systematically evaluate the aforementioned methods on two typical large-scale deep learning tasks: image classification (over 256 GPUs) and language modeling (over 64 GPUs). See Appendix \ref{app-add-experiments} for implementation details.

\subsection{Logistic Regression}
We consider a distributed logistic regression problem with $f_i(x) = \frac{1}{M}\sum_{m=1}^M \ln[1 + \exp(-y_{i,m} h_{i,m})^T x ]$, where $\{h_{i,m}, y_{i,m}\}_{m=1}^M$ are local data samples at agent $i$ with $h_{i,m} \in \mathbb{R}^d$ being the feature vector and $y_{i,m}\in\{+1,-1\}$ being the corresponding label. Each $h_{i,m}$ is generated from the normal distribution $\cN (0; 10 I_d)$. To generate $y_{i,m}$, we first generate an auxiliary random vector $x^\star_{i}\in \RR^d$ with each entry following $\cN(0, 1)$. Next, we generate $y_{i,m}$ from a uniform distribution $\cU(0, 1)$. If $y_{i,m} \le 1/[1 + \exp(- h_{i,m}^T x_i^\star)]$ then $y_{i,m}$ is set as $+1$; otherwise $y_{i,m}$ is set as $-1$. We let $x_i^\star = x^\star\ \forall i$ to generate data for iid scenario and $x_i^\star \neq x_j^\star\ \forall i,j$ for non-iid scenario. Each $x_i^\star$ is normalized.

Figure \ref{Fig:convex-comparison-gossip} compares how Gossip-PGA performs against parallel and Gossip SGD over the ring topology and non-iid data distribution. The network sizes are set as $n=20, 50, 100$ which results in $\beta = 0.967, 0.995, 0.998$. We set $d=10$ and $M=8000$. $H$ is set as $16$ in Gossip-PGA. The step-size $\gamma$ is initialized as $0.2$ and gets decreased by half for every $1000$ iterations. We repeat all simulations 50 times and illustrate the mean of all trials with solid curve and standard deviation with shaded area. It is observed that both Gossip SGD and Gossip-PGA will asymptotically converge at the same rate as parallel SGD (i.e., the linear speedup stage), albeit with different transient stages. Gossip-PGA always has shorter transient stages than Gossip SGD, and such superiority gets more evident when network size increases (recall that $1-\beta=O(1/n^2)$). For experiments on different topologies such as grid and exponential graph, on iid data distribution, and comparison with Local SGD, see Appendix \ref{app-add-experiments}. All experiments are consistent with the theoretical transient stage comparisons in Tables \ref{table-transient-stage} and \ref{table-transient-stage-local}.

\subsection{Image Classification}

The ImageNet-1k \cite{deng2009imagenet} \footnote{The usage of ImageNet dataset in this paper is for non-commercial research purposes only.} dataset consists of 1,281,167 training images and 50,000 validation images in 1000 classes. We train ResNet-50 \cite{he2016deep} model ($\sim$25.5M parameters) following the training protocol of \cite{goyal2017accurate}. We train total 120 epochs. The learning rate is warmed up in the first 5 epochs and is decayed by a factor of 10 at 30, 60 and 90 epochs.
We set the period to 6 for both Local SGD and Gossip-PGA. In Gossip-AGA, the period is set to 4 initially and changed adaptively afterwards, roughly 9\% iterations conduct global averaging.

\begin{table}
  \begin{center}
\begin{small}
\begin{sc}
  \begin{tabular}{rccc}
    \toprule
    Method     &  Acc.\%  &  Hrs & Epochs/Hrs to 76\%.\\
    \midrule
    Parallel SGD  & {76.26}   & 2.22  & 94 / 1.74  \\ 
    Local SGD     & 74.20   & 1.05  & N.A.  \\
    Local SGD $\times3$    & 75.41   & 3 & N.A.\\
    Gossip SGD     & 75.34        & 1.55 &  N.A. \\
    Gossip SGD $\times2$    & 76.18        & 3 &  198/2.55\\
    OSGP & 75.04 & 1.32 & N.A. \\
    OSGP $\times2$ & 76.07 & 2.59 & 212/2.28 \\
    Gossip-PGA     & {76.28}       & 1.66 & 109/1.50 \\
    Gossip-AGA     & {76.25}        & 1.57 & $\boldsymbol{91/1.20}$ \\
    \bottomrule
  \end{tabular}
  \end{sc}
\end{small}
\end{center}
  \vskip -0.15in
    \caption{Comparison of Top-1 validation accuracy (Column 2) and wall-clock training time (Column 3) on different methods after finishing all epochs. We also report the epochs and training time required to reach 76\% accuracy (Column 4). ``N.A.'' implies that the target accuracy is not reached when all epochs are completed.}
  \label{Table:imagetnet_result}
\end{table}

Table \ref{Table:imagetnet_result} shows the top-1 validation accuracy and wall-clock training time of aforementioned methods. It is observed both Gossip-PGA and Gossip-AGA can reach comparable accuracy with parallel SGD after all $120$ epochs but with roughly 1.3x $\sim$ 1.4x training time speed-up. On the other hand, while local and Gossip SGD completes all $120$ epochs faster than Gossip-PGA/AGA and parallel SGD, they suffer from a 2.06\% and 0.92\% accuracy degradation separately. Moreover, both algorithms cannot reach the 76\% top-1 accuracy within $120$ epochs. We also compare with OSGP \cite{assran2019stochastic}, which adding overlapping on the Gossip SGD. We find OSGP $\times2$, while faster than Gossip SGD$\times2$, still needs more time than Gossip-PGA to achieve 76\% accuracy. To further illustrate how much time it will take local and Gossip SGD to reach the target  accuracy, we run another Local SGD and Gossip SGD experiments with extended epochs (i.e., Gossip SGD $\times2$ trains total 240 epochs and the learning rate is decayed at 60, 120, and 180 epoch. Local SGD $\times3$ trains total 360 epochs and the learning rate is decayed at 90, 180, and 270 epochs). It is observed that Gossip-SGD $\times2$ can reach the target with notably more time expense than Gossip-PGA/AGA and parallel SGD, and Local SGD $\times3$ still cannot reach the 76\% accuracy. All these observations validate that periodic global averaging can accelerate Gossip SGD significantly.

Figure \ref{Fig:imagenet} shows the iteration-wise and runtime-wise convergence in terms of training loss. In the left figure, it is observed Gossip-PGA/AGA converges faster (in iteration) and more accurate than local and Gossip SGD, which is consistent with our theory. In the right figure, it is observed that Gossip-PGA/AGA is the fastest method (in time) that can reach the same training loss as parallel SGD.

\noindent \textbf{Compare with SlowMo.} Gossip-PGA is an instance of SlowMo, in which the base optimizer is set as Gossip SGD, slow momentum $\beta=0$, and slow learning rate $\alpha=1$. 
We made experiments to compare Gossip-PGA with SlowMo. It is observed the additional slow momentum update helps SlowMo with large $H$ but degrades it when $H$ is small. This observation is consistent with Fig.~3(a) in \cite{wang2019slowmo}. This observation implies that the slow momentum update may not always be beneficial in SlowMo.

\vspace{-2mm}
\begin{table}[h]
\centering
\begin{tabular}{lcc}
\toprule
\footnotesize \textbf{Period} &\footnotesize \textbf{Gossip-PGA} &\footnotesize \textbf{SlowMo} \\ \midrule
\footnotesize $H=6$                & \footnotesize $\boldsymbol{76.28}$            & \footnotesize $75.23$         \\
\footnotesize $H=48$               & \footnotesize $75.66$             & \footnotesize $\boldsymbol{75.81}$         \\ \bottomrule
\end{tabular}
\vskip -0.05in
\caption{Comparison of Top-1 validation accuracy with SlowMo with different periods.}
\vskip -0.1in
\label{slowmo-comparison}
\end{table}

\noindent \textbf{Ring Topology.} While the convergence property of Gossip-PGA is established over the {\em static} network topology, we utilize the dynamic one-peer exponential topology in the above deep experiments because it usually achieves better accuracy. To illustrate the derived theoretical results, we make an additional experiment, over the static ring topology, to compare Gossip-PGA with Gossip SGD in Table \ref{table-ring}. It is observed that 
Gossip-PGA can achieve better accuracy than Gossip SGD after running the same epochs, which coincides with our analysis that Gossip-PGA has faster convergence.

\begin{figure*}[t]

\begin{center}
\centerline{\includegraphics[width=0.42\textwidth]{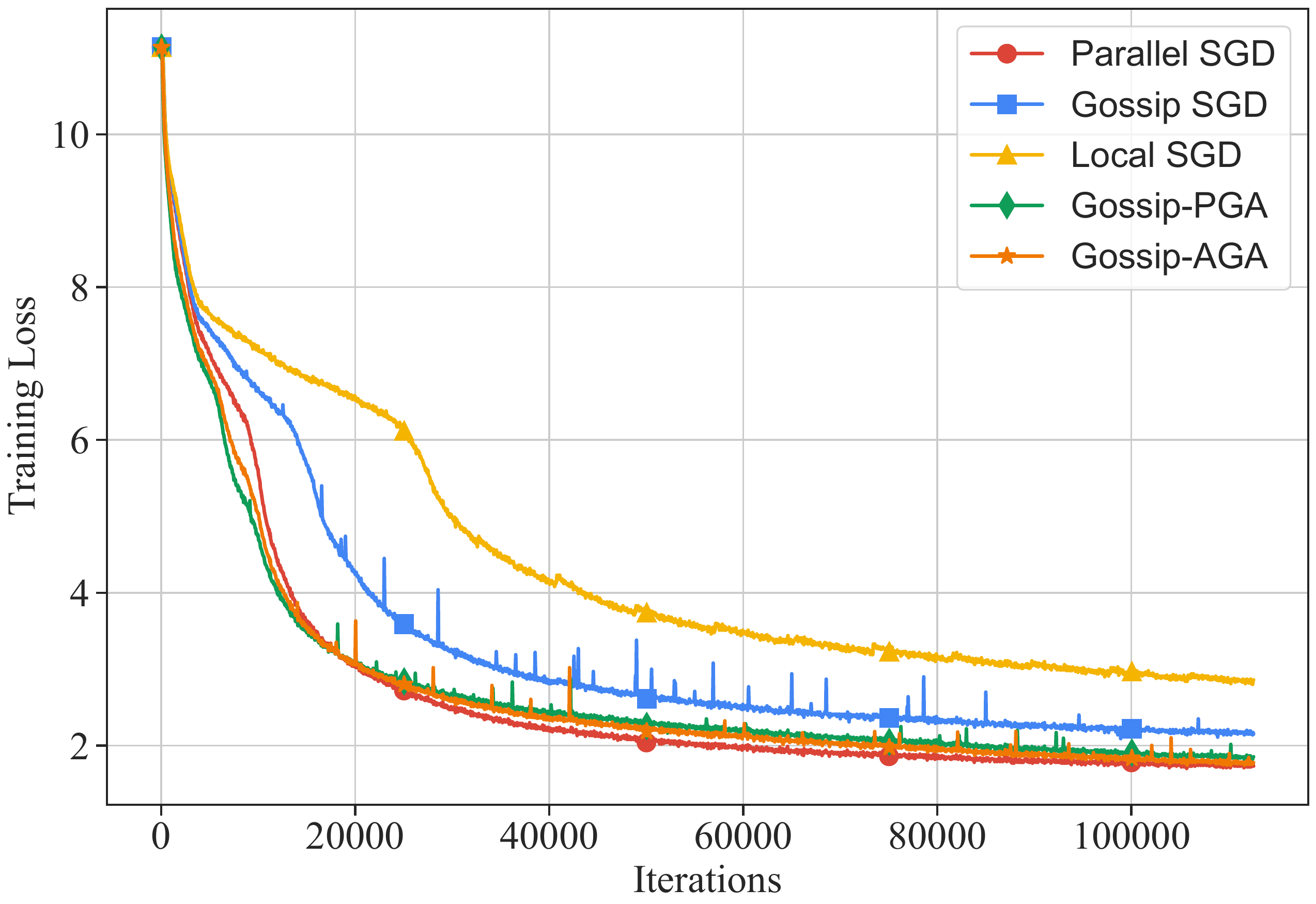} 
\quad
\quad 
\includegraphics[width=0.42\textwidth]{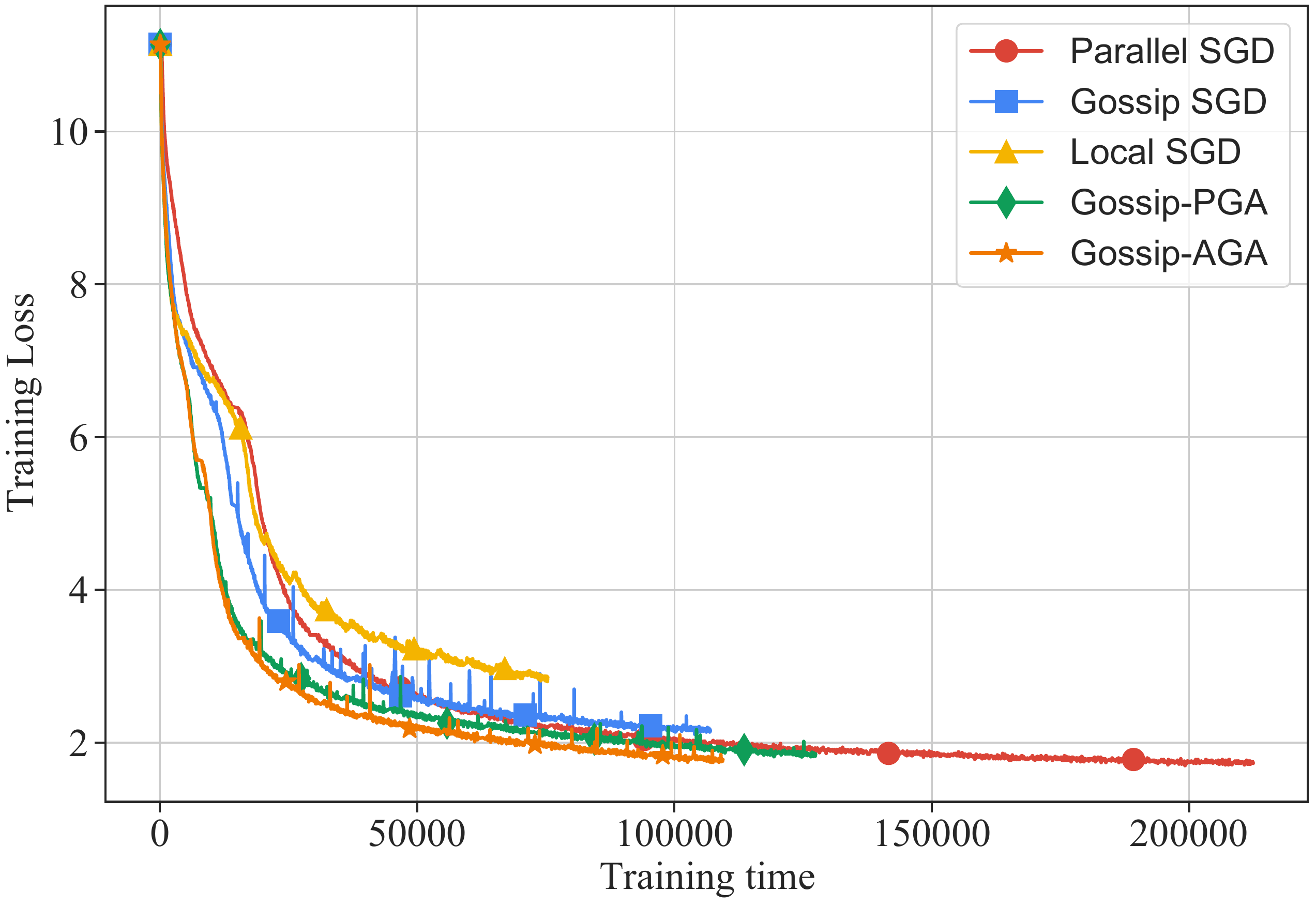}}
\end{center}
\vskip -0.25in
\caption{Convergence results of BERT on the language modeling task in terms of iteration and runtime.}
\label{Fig:BERT}
\vskip -0.15in
\end{figure*}

\begin{table}[h]
\centering
\begin{tabular}{rccc}
\toprule
\textbf{\footnotesize{Method}} & \footnotesize{\textbf{Epoch}} & \textbf{\footnotesize Acc$\%$} & \textbf{\footnotesize{Time(Hrs.)}} \\ \midrule
\footnotesize{Gossip SGD } &      \footnotesize  120          &   \footnotesize  74.86           &  \footnotesize 1.56                              \\ 
\footnotesize{Gossip PGA} &      \footnotesize   120         &    \footnotesize  75.94          & \footnotesize 1.68             \\\bottomrule
\end{tabular}
\vskip -0.05in
\caption{Comparison of Top-1 validation accuracy on Gossip-PGA and Gossip SGD with ring topology.}
\label{table-ring}
\end{table}

\noindent \textbf{Scalability.} We establish in Theorem \ref{thm:nc} that Gossip-PGA can achieve linear speedup in the non-convex setting. To validate it, we conduct a scaling experiment and list the result in Table \ref{table:scaling}. Figures represent the final accuracy and hours to finish training. It is observed that Gossip-PGA can achieve a roughly linear speedup in training time  without notably performance degradation.

\begin{table}[h]
\centering
\setlength{\tabcolsep}{1.75mm}{
\begin{tabular}{ccccc}
\toprule
\footnotesize \textbf{Method} &\footnotesize \textbf{4 nodes} & \footnotesize \textbf{8 nodes}  &\footnotesize \textbf{16 nodes} & \footnotesize \textbf{32 nodes} \\ 
\midrule
\footnotesize Parallel SGD  & \footnotesize ${76.3/11.6}$  & \footnotesize $76.4/6.3$   & \footnotesize $76.3/3.7$ & \footnotesize $76.2/2.2$     \\
\footnotesize Gossip SGD & \footnotesize $76.3/11.1$  & \footnotesize $76.4/5.7$ & \footnotesize $75.9/2.8$ & \footnotesize $75.0/1.5$        \\ 
\footnotesize Gossip PGA & \footnotesize $76.4/11.2$  & \footnotesize $76.7/5.9$ & \footnotesize $76.3/3.0$ & \footnotesize $76.2/1.6$ \\
\bottomrule
\end{tabular}}
\vskip -0.05in
\caption{Scaling effects on different methods with different numbers of nodes. Figures represent the final accuracy and hours to complete training.}
\label{table:scaling}
\end{table}

\subsection{Language Modeling}

\begin{table}
  \begin{center}
  \begin{small}
  \begin{sc}
  \begin{tabular}{rcc}
    \toprule
    Method     & Final Loss  & runtime (hrs) \\
    \midrule
    Parallel SGD      & 1.75  & 59.02 \\
    Local SGD      & 2.85  & 20.93 \\
    Local SGD $\times3$      & 1.88  & 60 \\
    Gossip SGD    & 2.17       & 29.7 \\
    Gossip SGD $\times2$    & 1.81       & 59.7 \\
    Gossip-PGA    & 1.82       & 35.4 \\
    Gossip-AGA      & 1.77 & 30.4 \\
    \bottomrule
  \end{tabular}
\end{sc}
\end{small}
\end{center}
  \vskip -0.1in
    \caption{Comparison of training loss and training time of BERT training on different algorithms after completing all training steps. }
  \label{Table:BERT_result}
\end{table}

BERT \cite{devlin2018bert} is a widely used pre-training language representation model for NLP tasks. We train a BERT-Large model ($\sim$330M parameters) on the Wikipedia and BookCorpus datasets. We set the period to 6 for both Local SGD and Gossip-PGA. 
In Gossip-AGA, the period is set to 4 initially and changed adaptively afterwards,  roughly 9.6\% iterations conduct global averaging.

Table \ref{Table:BERT_result} shows the final training loss and training runtime of the aforementioned methods. Gossip-AGA can reach comparable training loss with parallel SGD, but with roughly 1.94 x training time speed-up. Gossip SGD and Local SGD cannot reach training loss that below 1.8 even if they are trained over $60$ hours (see Local SGD $\times3$ and Gossip SGD $\times2$.) Figure \ref{Fig:BERT} shows the iteration-wise and runtime-wise convergence w.r.t training loss of the aforementioned methods. The left plot shows Gossip-PGA/AGA has almost the same convergence as Gossip SGD in iterations; the right plot shows that Gossip-AGA is the fastest method in training time that can reach the same accuracy as parallel SGD.

\section{Conclusion}
We introduce Gossip-PGA/AGA to mitigate the slow convergence rate of Gossip SGD in distributed training. Theoretically, we prove the convergence improvement in smooth convex and non-convex problem. Empirically, experimental results of large-scale training validate our theories.


\bibliography{references}
\bibliographystyle{icml2021}

\newpage
\onecolumn
\appendix
\section{Preliminary}
\vspace{2mm}
\noindent \textbf{Notation.} We first introduce necessary notations as follows.
\begin{itemize}
  \item $\vx^{(k)} = [(\x_1^{(k)})^T; (\x_2^{(k)})^T; \cdots; (\x_n^{(k)})^T]\in \mathbb{R}^{n\times d}$
  \item $\nabla F(\vx^{(k)};\bxi^{(k+1)}) = [\nabla F_1(\x_1^{(k)};\bxi_1^{(k+1)})^T; \cdots; \nabla F_n(\x_n^{(k)};\bxi_n^{(k+1)})^T]\in \mathbb{R}^{n\times d}$
  \item $\nabla f(\vx^{(k)}) = [\nabla f_1(\x_1^{(k)})^T; \nabla f_2(\x_2^{(k)})^T; \cdots; \nabla f_n(\x_n^{(k)})^T]\in \mathbb{R}^{n\times d}$ 
  \item $\bar{\vx}^{(k)} = [(\bar{\x}^{(k)})^T; (\bar{\x}^{(k)})^T; \cdots; (\bar{\x}^{(k)})^T]\in \mathbb{R}^{n\times d}$ where $\bar{\x}^{(k)} = \frac{1}{n}\sum_{i=1}^n \x_i^{(k)}$
  \item ${\vx}^{\star} = [({x}^{\star})^T; ({x}^{\star})^T; \cdots; ({x}^{\star})^T]\in \mathbb{R}^{n\times d}$ where ${x}^{\star}$ is the global solution to problem \eqref{eq:general-prob}.
  \item $W=[w_{ij}]\in \mathbb{R}^{n\times n}$.
  \item $\mathds{1}_n = \mathrm{col}\{1,1,\cdots, 1\} \in \RR^n$.
  \item Given two matrices $\vx, \vy \in \RR^{n\times d}$, we define inner product $\langle \vx, \vy \rangle = \mathrm{tr}(\vx^T \vy)$ and the Frobenius norm $\|\vx\|_F^2 = \langle \vx, \vx \rangle$. 
  \item Given $W\in \RR^{n\times n}$, we let $\|W\|_2 = \sigma_{\max}(W)$ where   $\sigma_{\max}(\cdot)$ denote the maximum sigular value.
\end{itemize}

\vspace{2mm}
\noindent \textbf{Gossip-PGA in matrix notation.} For ease of analysis, we rewrite the main recursion of Gossip-PGA in matrix notation: 
\begin{align}
\vx^{(k+1)} & = 
\begin{cases}
W(\vx^{(k)}  - \gamma \nabla F(\vx^{(k)}; \bxi^{(k+1)}) )\hspace{3mm}& \mbox{If $\mathrm{mod}(k+1, H) \neq 0$ } \\
\frac{1}{n}\mathds{1}_n \mathds{1}_n^T(\vx^{(k)}  - \gamma \nabla F(\vx^{(k)}; \bxi^{(k+1)}) ) \hspace{3mm}& \mbox{otherwise} 
\end{cases}\label{eq:DSGD_global_average}
\end{align}

\vspace{2mm}
\noindent \textbf{Gradient noise.} We repeat the definition of filtration in Assumption \ref{ass:gradient-noise} here for convenience.
\begin{align}
    \hspace{-2mm}\cF^{(k)} \hspace{-0.8mm}=\hspace{-0.8mm} \big\{\hspace{-0.8mm} \{\hspace{-0.4mm}\x_i^{(k)}\hspace{-0.4mm}\}_{i=1}^n,  \{\hspace{-0.4mm}\bxi_i^{(k)}\hspace{-0.4mm}\}_{i=1}^n, \cdots, \{\hspace{-0.4mm}\x_i^{(0)}\hspace{-0.4mm}\}_{i=1}^n,  \{\hspace{-0.4mm}\bxi_i^{(0)}\hspace{-0.4mm}\}_{i=1}^n\hspace{-0.8mm} \big\}
\end{align}

\vspace{1mm}
\begin{itemize}
    \item With Assumption \ref{ass:gradient-noise}, we can evaluate the magnitude of the averaged gradient noise: 
\begin{align}\label{wendn}
&\hspace{-2cm} \mathbb{E}[\|\frac{1}{n}\sum_{i=1}^n \nabla F_i(\x_i^{(k-1)};\bxi_i^{(k)}) - \frac{1}{n}\sum_{i=1}^n \nabla f_i(\x_i^{(k-1)}) \|^2|\cF^{(k-1)}] \nonumber \\
\overset{(a)}{=}& \frac{1}{n^2}\sum_{i=1}^n \mathbb{E}[\|\nabla F_i(\x_i^{(k-1)};\bxi_i^{(k)}) - \nabla f_i(\x_i^{(k-1)}) \|^2|\cF^{(k-1)}] \overset{\eqref{gd-2}}{\le} \frac{\sigma^2}{n}
\end{align}
where (a) holds because $\bxi_i^{(k)}$ is independent for any $i$ and $\mathbb{E}[\nabla F_i(\x_i^{(k-1)};\bxi_i^{(k)}) - \nabla f_i(\x_i^{(k-1)})|\cF^{(k-1)}]=0$.

\item We define gradient noise as $\s_i^{(k)} =  \nabla F_i(\x_i^{(k-1)};\bxi_i^{(k)}) - \nabla f_i(\x_i^{(k-1)})$. For any $0 \le j \le k < \ell$, it holds that
\begin{align}
\mathbb{E}[\big(\s_i^{(k)}\big)^T \s_i^{(\ell)}|\cF^{(j)}] \overset{(a)}{=} \mathbb{E}_{\cF^{(\ell-1)}/\cF^{(j)}}\big[\mathbb{E}[\big(\s_i^{(k)}\big)^T \s_i^{(\ell)}|\cF^{(\ell-1)}]\big] = \mathbb{E}_{\cF^{(\ell-1)}/\cF^{(j)}}\big[\big(\s_i^{(k)}\big)^T \mathbb{E}[ \s_i^{(\ell)}|\cF^{(\ell-1)}]\big] \overset{\eqref{gd-1}}{=} 0 \label{gd-1-appendix}
\end{align}
where  $\cF^{(\ell-1)}/\cF^{(j)} := \big\{\hspace{-0.8mm} \{\hspace{-0.4mm}\x_i^{(j+1)}\hspace{-0.4mm}\}_{i=1}^n,  \{\hspace{-0.4mm}\bxi_i^{(j+1)}\hspace{-0.4mm}\}_{i=1}^n, \cdots, \{\hspace{-0.4mm}\x_i^{(\ell-1)}\hspace{-0.4mm}\}_{i=1}^n,  \{\hspace{-0.4mm}\bxi_i^{(\ell-1)}\hspace{-0.4mm}\}_{i=1}^n\hspace{-0.8mm} \big\}$ and (a) holds due to the law of total expectation. 

\item For any $0\le k < \ell$, it holds that
\begin{align}\label{gd-2-appendix}
\mathbb{E}[\|\s_i^{(\ell)}\|^2|\cF^{(k)}] = \mathbb{E}_{\cF^{(\ell-1)}/\cF^{(k)}}\big[\mathbb{E}[\|\s_i^{(\ell)}\|^2|\cF^{\ell-1}]\big] \overset{\eqref{gd-2}}{\le} \mathbb{E}_{\cF^{(\ell-1)}/\cF^{(k)}}\big[ \sigma^2 \big] = \sigma^2
\end{align}

\end{itemize}

\vspace{2mm}
\noindent \textbf{Smoothness.} Since each $f_i(x)$ is assumed to be $L$-smooth in Assumption \ref{ass:smoothness}, it holds that $f(x) = \frac{1}{n}\sum_{i=1}^n f_i(x)$ is also $L$-smooth. As a result, the following inequality holds for any $\x, \y \in \mathbb{R}^d$:
\begin{align}
f_i(\x) - f_i(\y) - \frac{L}{2}\|\x - \y\|^2 &\le  \langle \nabla f_i(\y), \x- \y \rangle \label{sdu-2}
\end{align}

\noindent \textbf{Smoothness and convexity.} If each $f_i(x)$ is further assumed to be convex (see Assumption \ref{ass-strongly-convex}), it holds that $f(x) = \frac{1}{n}\sum_{i=1}^n f_i(x)$ is also  convex. For this scenario, it holds for any $\x, \y \in \mathbb{R}^d$ that:
\begin{align}
\|\nabla f(\x) - \nabla f(x^\star)\|^2 &\le 2L\big(f(\x) - f(x^\star)\big) \label{sdu-1} \\
f_i(\x) - f_i(\y) &\le  \langle \nabla f_i(\x), \x - \y \rangle  \label{sdu-3}
\end{align}

\noindent \textbf{Network weighting matrix.} Suppose a weighting matrix $W\in \RR^{n\times n}$ satisfies Assumption \ref{ass:weight-matrix}, it holds that 
\begin{align}\label{network-inequaliy}
\|W - \frac{1}{n}\mathds{1}\mathds{1}^T\|_2 \le \beta, \quad \|(W - \frac{1}{n}\mathds{1}\mathds{1}^T)^{k}\|_2 \le \beta^k,\quad \forall~k.
\end{align}

\noindent \textbf{Submultiplicativity of the Frobenius norm.} Given matrices $W\in \RR^{n\times n}$ and $\vy\in \RR^{n\times d}$, it holds that 
\begin{align}\label{submulti}
\|W\vy\|_F \le \|W\|_2 \|\vy\|_F.
\end{align}
To verify it, by letting $y_j$ be the $j$-th column of $\vy$, we have $\|W\vy\|_F^2 = \sum_{j=1}^d \|Wy_j\|_2^2 \le \sum_{j=1}^d \|W\|_2^2 \|y_j\|_2^2=\|W\|_2^2\|\vy\|_F^2$.

\section{Convergence analysis for convex scenario }\label{app-sc-descent}

\vspace{2mm}
\subsection{Proof Outline for Theorem \ref{thm-convex}}

The following lemma established in \citep[Lemma~8]{koloskova2020unified} shows how $\mathbb{E}\|\bar{\x}^{{(k)}} - x^\star\|^2$ evolves with iterations.
\begin{lemma}[{\sc Descent Lemma} \cite{koloskova2020unified}] \label{lm-descent}
  When Assumptions  \ref{ass:smoothness}--\ref{ass-strongly-convex} hold and step-size $\gamma < \frac{1}{4L}$, it holds for $k=1,2,\cdots$ that 
  \begin{align}\label{23bsd999}
  & \mathbb{E}\|\bar{\x}^{{(k)}} - x^\star\|^2 \le \mathbb{E}\|\bar{\x}^{(k-1)} - x^\star\|^2 - \gamma \big(\mathbb{E} f(\bar{\x}^{(k-1)}) - f(x^\star)\big) + \frac{3L\gamma}{2n}\mathbb{E}\|\vx^{(k-1)}- \bar{\vx}^{(k-1)}\|_F^2 + \frac{\gamma^2 \sigma^2}{n},
  \end{align}
  where $\bar{\vx}^{(k)} = [(\bar{\x}^{(k)})^T; \cdots; (\bar{\x}^{(k)})^T]\in \mathbb{R}^{n\times d}$.
\end{lemma}

\begin{remark}
It is worth noting that Lemma \ref{lm-descent} also holds for the standard Gossip SGD algorithm. The periodic global averaging step does not affect this descent lemma.
\end{remark}

Next we establish the consensus lemmas in which Gossip-PGA is fundamentally different from Gossip SGD. Note that Gossip-PGA takes global average every $H$ iterations. For any $k=0,1,\cdots$, we define 
\begin{align}
\tau(k) = \max\{\ell: \ell \le k \mbox{ and }  \mathrm{mod}(\ell, H) =0 \}
\end{align}
as the most recent iteration when global average is conducted. In Gossip-PGA, it holds that $\bar{\x}^{\tau(k)}= \x^{\tau(k)}_i$ for any $i \in [n]$. This is different from Gossip SGD in which $\bar{\x}^{(k)} = \x_i^{(k)}$ can only happen when $k=0$.

For Gossip-PGA, the real challenge is to investigate how the periodic global averaging helps reduce consensus error and hence accelerate the convergence rate. In fact, there are two forces in Gossip-PGA that drive local model parameters to reach consensus: the gossip communication and the periodic global averaging. Each of these two forces is possible to dominate the consensus controlling in different scenarios:

\textbf{Scenario I.} Global averaging is more critical to guarantee consensus on large or sparse network, or when global averaging is conducted frequently.

\textbf{Scenario II.} Gossip communication is more critical to guarantee consensus on small or dense network, or when global averaging is conducted infrequently.
  
   
  
Ignoring either of the above scenario will lead to incomplete or even incorrect conclusions, as shown in Remark \ref{remark-comparison-with-jaggi}. In the following, we will establish a specific consensus lemma for each scenario and then unify them into one that precisely characterize how the consensus distance evolves with iterations in Gossip-PGA.

\begin{lemma}[\sc Consensus Lemma: Global averaging dominating]\label{lm-consensus-I} Under Assumptions \ref{ass:smoothness}--\ref{ass-strongly-convex}, it holds for $k = \tau(k), \tau(k)+1, \cdots, \tau(k)+H-1$ that 
\begin{align}\label{2enn}
\mathbb{E}\|\vx^{(k+1)} - \bar{\vx}^{(k+1)}\|_F^2 \le&\ 6 {H} \gamma^2 \beta^2 L^2 \sum_{\ell=\tau(k)}^{k} \beta^{k-\ell} \mathbb{E}\|\vx^{(\ell)} \hspace{-0.8mm}-\hspace{-0.8mm} \bar{\vx}^{(\ell)}\|_F^2 \nonumber \\ 
&\ + 12 n {H} \gamma^2 \beta^2 L \sum_{\ell=\tau(k)}^{k}\beta^{k-\ell} \mathbb{E}(f(\bar{\x}^{(\ell)}) - f(x^\star)) \hspace{-0.8mm}+\hspace{-0.8mm} 2n \gamma^2 \beta^2 C_\beta (3 b^2 \hspace{-0.8mm}+\hspace{-0.8mm} \sigma^2)
\end{align}
where $b^2 = \frac{1}{n}\sum_{i=1}^n \|\nabla f_i(x^\star)\|^2$ implies  data heterogeneity and $C_\beta = \sum_{k=0}^{H-1}\beta^{k} =  (1-\beta^{H})/(1-\beta)$.
\end{lemma} 

\begin{lemma}[\sc Consensus Lemma: Gossip Dominating]\label{lm-consensus-2} Under Assumptions \ref{ass:smoothness}--\ref{ass-strongly-convex}, it holds for $k = \tau(k), \tau(k)+1, \cdots, \tau(k)+H-1$ that 
\begin{align}\label{2enn-2}
\mathbb{E}\|\vx^{(k+1)} - \bar{\vx}^{(k+1)}\|_F^2 \le&\ \frac{6  \gamma^2 \beta^2 L^2}{{1-\beta}} \sum_{\ell=\tau(k)}^{k} \beta^{k-\ell} \mathbb{E}\|\vx^{(\ell)} \hspace{-0.8mm}-\hspace{-0.8mm} \bar{\vx}^{(\ell)}\|_F^2 \nonumber \\ 
&\ + \frac{12 n \gamma^2 \beta^2 L}{1-\beta} \sum_{\ell=\tau(k)}^{k}\beta^{k-\ell} \mathbb{E}(f(\bar{\x}^{(\ell)}) - f(x^\star)) \hspace{-0.8mm}+\hspace{-0.8mm} 2n \gamma^2 \beta^2 C_\beta (\frac{3 b^2}{1-\beta} \hspace{-0.8mm}+\hspace{-0.8mm} \sigma^2)
\end{align}
\end{lemma} 

Observing Lemmas  \ref{lm-consensus-I} and \ref{lm-consensus-2}, it is found that bounds \eqref{2enn} and \eqref{2enn-2} are in the same shape except for some critical coefficients. With the following relation:
\begin{align}
\begin{cases}
y \le a_1 x + b \\
y \le a_2 x + b
\end{cases}
\quad
\Longrightarrow 
\quad
y\le \min\{a_1, a_2\} x + b,
\end{align}
we can unify Lemmas \ref{lm-consensus-I} and \ref{lm-consensus-2} into:

\begin{lemma}[\sc Unified Consensus Lemma]\label{lm-consensus-unified} Under Assumptions \ref{ass:smoothness}--\ref{ass-strongly-convex}, it holds for $k = \tau(k), \tau(k)+1, \cdots, \tau(k)+H-1$ that 
\begin{align}\label{2enn-unified}
\mathbb{E}\|\vx^{(k+1)} - \bar{\vx}^{(k+1)}\|_F^2 \le&\ 6 D_\beta \gamma^2 \beta^2 L^2 \sum_{\ell=\tau(k)}^{k} \beta^{k-\ell} \mathbb{E}\|\vx^{(\ell)} \hspace{-0.8mm}-\hspace{-0.8mm} \bar{\vx}^{(\ell)}\|_F^2 \nonumber \\ 
&\ + 12 n D_\beta \gamma^2 \beta^2 L \sum_{\ell=\tau(k)}^{k}\beta^{k-\ell} \mathbb{E}(f(\bar{\x}^{(\ell)}) - f(x^\star)) \hspace{-0.8mm}+\hspace{-0.8mm} 2n \gamma^2 \beta^2 C_\beta (3 D_\beta b^2 \hspace{-0.8mm}+\hspace{-0.8mm} \sigma^2)
\end{align}
where $b^2 = \frac{1}{n}\sum_{i=1}^n \|\nabla f_i(x^\star)\|^2$ implies  data heterogeneity, $C_\beta = \sum_{k=0}^{H-1}\beta^{k} =  (1-\beta^{H})/(1-\beta)$, and 
$D_\beta = \min\{\frac{1}{1-\beta}, H\}$.
\end{lemma}

\begin{remark}
This lemma reflects how the network topology and the global averaging period contribute to the consensus controlling. For scenario I where the network is large or sparse such that $1/(1-\beta) > H$, Lemma \ref{lm-consensus-unified} indicates that the consensus error is mainly controlled by the global averaging period (i.e., $D_\beta = H$). On the other hand, for scenario II where the network is small or dense such that $1/(1-\beta) < H$, Lemma \ref{lm-consensus-unified} indicates that the consensus error is mainly controlled by gossip communication (i.e., $D_\beta = 1/(1-\beta)$).
\end{remark}

Using Lemma \ref{lm-consensus-unified}, we derive the upper bound of the weighted running average of $\mathbb{E}\|\vx^k - \bar{\vx}^k\|_F^2$:

\begin{lemma}[\sc Running consensus lemma] \label{lm-weighted-consensus}
Suppose Assumptions \ref{ass:smoothness}--\ref{ass-strongly-convex} hold and step-size $\gamma < {1}/(4 L \beta D_\beta )$, it holds for $T>0$ that 
\begin{align}\label{23nbd}
\frac{1}{T+1}\sum_{k=0}^T \mathbb{E}\|\vx^{(k)} - \bar{\vx}^{(k)}\|_F^2 \le \frac{2 c_2 D_\beta }{T+1} \sum_{k =0}^T \big(\mathbb{E} f(\bar{\x}^{(k)}) - f(x^\star)\big)+ 2c_3
\end{align}
where $c_2$ and $c_3$ are constants defined as
\begin{align}
c_2 &= 12n \beta^2 D_\beta \gamma^2 L, \\  
c_3 &= 2n  \beta^2 \gamma^2 C_\beta (3 D_\beta b^2 + \sigma^2)
\end{align}

\end{lemma}
With Lemmas \ref{lm-descent} and \ref{lm-weighted-consensus}, we can  establish the final convergence Theorem \ref{thm-convex}, see the proof in Sec. \ref{proof-thm1}.

\subsection{Proof of Lemma \ref{lm-descent}.} This lemma was first established in \citep[Lemma~8]{koloskova2020unified}. We made slight improvement to tight constants appeared in step-size ranges and upper bound \eqref{23bsd999}. For readers' convenience, we repeat arguments here.

Recall the algorithm in \eqref{eq:DSGD_global_average}. By taking the average on both sides, we reach that 
\begin{align}
\bar{\x}^{{(k)}} - x^\star = \bar{\x}^{{(k-1)}} - x^\star -  \frac{\gamma}{n}\sum_{i=1}^n \nabla F_i(\x_i^{(k-1)};\bxi_i^{(k)}),\quad \forall k=1,2,\cdots
\end{align}
By taking expectation over the square of both sides of the above recursion conditioned on $\cF^{(k-1)}$, we have
\begin{align}\label{23hnsd}
&\hspace{-1cm} \mathbb{E}[\|\bar{\x}^{{(k)}} - x^\star \|^2|\cF^{(k-1)}] \nonumber \\
\overset{\eqref{gd-1}}{=}&\ \|\bar{\x}^{{(k-1)}} - x^\star -  \frac{\gamma}{n}\sum_{i=1}^n \nabla f_i(\x_i^{(k-1)})\|^2 + \gamma^2 \mathbb{E}[\|\frac{1}{n}\sum_{i=1}^n \nabla F_i(\x_i^{(k-1)};\bxi_i^{(k)}) - \frac{1}{n}\sum_{i=1}^n \nabla f_i(\x_i^{(k-1)}) \|^2|\cF^{(k-1)}] \nonumber \\
\overset{\eqref{wendn}}{\le}&\ \|\bar{\x}^{{(k-1)}} - x^\star -  \frac{\gamma}{n}\sum_{i=1}^n \nabla f_i(\x_i^{(k-1)})\|^2 + \frac{\gamma^2 \sigma^2}{n}
\end{align}
Note that the first term can be expanded as follows. 
\begin{align}\label{shdshsd}
&\ \|\bar{\x}^{{(k-1)}} - x^\star -  \frac{\gamma}{n}\sum_{i=1}^n \nabla f_i(\x_i^{(k-1)})\|^2 \nonumber \\
=&\ \|\bar{\x}^{{(k-1)}} - x^\star -  \frac{\gamma}{n}\sum_{i=1}^n [\nabla f_i(\x_i^{(k-1)}) - \nabla f_i(x^\star)]\|^2 \nonumber \\
=&\ \|\bar{\x}^{{(k-1)}} - x^\star\|^2 - \underbrace{\frac{2\gamma}{n}\sum_{i=1}^n \langle \bar{\x}^{{(k-1)}} - x^\star, \nabla f_i(\x_i^{(k-1)}) - \nabla f_i(x^\star) \rangle}_{(A)} + \underbrace{\gamma^2\| \frac{1}{n}\sum_{i=1}^n [\nabla f_i(\x_i^{(k-1)}) - \nabla f_i(x^\star)]\ \|^2}_{(B)}
\end{align}

We now bound the term (A):
\begin{align}\label{23nbsdn-2}
& \frac{2\gamma}{n}\sum_{i=1}^n \langle \bar{\x}^{{(k-1)}} - x^\star, \nabla f_i(\x_i^{(k-1)}) - \nabla f_i(x^\star) \rangle \nonumber \\
=&\  \frac{2\gamma}{n}\sum_{i=1}^n \langle \bar{\x}^{{(k-1)}} - x^\star, \nabla f_i(\x_i^{(k-1)})  \rangle \nonumber \\
=&\ \frac{2\gamma}{n}\sum_{i=1}^n \langle \bar{\x}^{{(k-1)}} - \x_i^{(k-1)}, \nabla f_i(\x_i^{(k-1)}) \rangle  + \frac{2\gamma}{n}\sum_{i=1}^n \langle \x_i^{(k-1)} - x^\star, \nabla f_i(\x_i^{(k-1)})  \rangle \nonumber \\
\overset{(a)}{\ge}&\ \frac{2\gamma}{n}\sum_{i=1}^n \Big( f_i(\bar{\x}^{{(k-1)}}) - f_i(\x_i^{(k-1)}) - \frac{L}{2}\|\bar{\x}^{{(k-1)}} - \x_i^{(k-1)}\|^2 \Big) + \frac{2\gamma}{n}\sum_{i=1}^n \Big(f_i(\x_i^{(k-1)}) - f_i(x^\star) \Big)  \nonumber \\
=&\ \frac{2\gamma}{n}\sum_{i=1}^n\Big( f_i(\bar{\x}^{{(k-1)}}) - f_i(x^\star) \Big) - \frac{\gamma L}{n}\|\bar{\vx}^{{(k-1)}} - \vx^{(k-1)}\|_F^2 \nonumber \\
=&\ 2\gamma \big( f(\bar{\x}^{{(k-1)}}) - f(x^\star)\big) - \frac{\gamma L}{n}\|\bar{\vx}^{{(k-1)}} - \vx^{(k-1)}\|_F^2
\end{align}
where (a) holds because of the inequality \eqref{sdu-2} and \eqref{sdu-3}. We next bound term (B) in \eqref{shdshsd}:
\begin{align}\label{23nbsdn-1}
& \gamma^2\| \frac{1}{n}\sum_{i=1}^n [\nabla f_i(\x_i^{(k-1)}) - \nabla f_i(x^\star)]\ \|^2 \nonumber \\
=&\ \gamma^2\| \frac{1}{n}\sum_{i=1}^n [\nabla f_i(\x_i^{(k-1)}) - \nabla f_i(\bar{\x}^{(k-1)}) + \nabla f_i(\bar{\x}^{(k-1)})  -  \nabla f_i(x^\star)]\ \|^2 \nonumber \\
\overset{\eqref{smooth-1}}{\le}&\ \frac{2\gamma^2 L^2}{n}\|\vx^{(k-1)} - \bar{\vx}^{(k-1)}\|_F^2 + 2\gamma^2 \|\nabla f(\bar{\x}^{(k-1)}) - \nabla f(x^\star)\|^2 \nonumber \\
\overset{\eqref{sdu-1}}{\le}&\ \frac{2\gamma^2 L^2}{n}\|\vx^{(k-1)} - \bar{\vx}^{(k-1)}\|_F^2 + 4L \gamma^2\big( f(\bar{\x}^{(k-1)}) - f({x}^{\star}) \big).
\end{align}
Substituting \eqref{23nbsdn-1} and \eqref{23nbsdn-2} into \eqref{shdshsd}, we have
\begin{align}\label{shdshsd-2}
&\ \|\bar{\x}^{{(k-1)}} - x^\star -  \frac{\gamma}{n}\sum_{i=1}^n \nabla f_i(\x_i^{(k-1)})\|^2 \nonumber \\
\le&\ \|\bar{\x}^{(k-1)} - x^\star\|^2 - 2\gamma(1 - 2L\gamma) \big( f(\bar{\x}^{{(k-1)}}) - f(x^\star)\big) + \Big( \frac{\gamma L}{n} + \frac{2\gamma^2 L^2}{n}\Big)\|\bar{\vx}^{{(k-1)}} - \vx^{(k-1)}\|_F^2  \nonumber \\
\le&\ \|\bar{\x}^{(k-1)} - x^\star\|^2 - \gamma \big( f(\bar{\x}^{{(k-1)}}) - f(x^\star)\big) + \frac{3\gamma L}{2n}\|\bar{\vx}^{{(k-1)}} - \vx^{(k-1)}\|_F^2 
\end{align}
where the last inequality holds when $\gamma < \frac{1}{4L}$. Substituting the above inequality into \eqref{23hnsd} and taking expectation over the filtration, we reach the result in \eqref{23bsd999}.

\subsection{Proofs of Lemma \ref{lm-consensus-I} and \ref{lm-consensus-2}.} Note the gossip averaging is conducted when $k = \tau(k), \tau(k)+1, \cdots, \tau(k)+H-1$, i.e., 
\begin{align}\label{2nds-test}
\vx^{(k+1)} = W(\vx^{(k)}  - \gamma \nabla F(\vx^{(k)}; \bxi^{(k+1)}) ).
\end{align}
Since $\bar{\vx}^{(k+1)} = \frac{1}{n}\mathds{1}\mathds{1}^T \vx^{(k+1)}$, it holds that 
\begin{align}
\bar{\vx}^{(k+1)}  =  \frac{1}{n}\mathds{1}\mathds{1}^T (\vx^{(k)}  - \gamma \nabla F(\vx^{(k)}; \bxi^{(k+1)}) ).
\end{align}
With the above two recursions, we have
\begin{align}\label{xchsdh}
\vx^{(k+1)} - \bar{\vx}^{(k+1)} = (W - \frac{1}{n}\mathds{1}\mathds{1}^T) (\vx^{(k)}  - \gamma \nabla F(\vx^{(k)}; \bxi^{(k+1)}) )
\end{align}
In the following we will derive two upper bounds for $\mathbb{E}\|\vx^{(k+1)} - \bar{\vx}^{(k+1)}\|^2$.

\vspace{1mm}
\noindent \textbf{Bound in Lemma \ref{lm-consensus-I}}. With \eqref{xchsdh}, we have 
\begin{align}
\vx^{(k+1)} - \bar{\vx}^{(k+1)} &= (W - \frac{1}{n}\mathds{1}\mathds{1}^T) (\vx^{(k)}  - \gamma \nabla F(\vx^{(k)}; \bxi^{(k+1)}) ) \nonumber \\
&= (W - \frac{1}{n}\mathds{1}\mathds{1}^T) (\vx^{(k)} - \bar{\vx}^{(k)}  - \gamma \nabla F(\vx^{(k)}; \bxi^{(k+1)}) ) \nonumber \\
&= (W - \frac{1}{n}\mathds{1}\mathds{1}^T)^{k+1-\tau(k)} (\vx^{(\tau(k))} - \bar{\vx}^{(\tau(k))}) - \gamma \sum_{\ell=\tau(k)}^{k}(W - \frac{1}{n}\mathds{1}\mathds{1}^T)^{k+1-\ell} \nabla F(\vx^{(\ell)}; \bxi^{(\ell+1)}) \nonumber \\
&= - \gamma \sum_{\ell=\tau(k)}^{k}(W - \frac{1}{n}\mathds{1}\mathds{1}^T)^{k+1-\ell} \nabla F(\vx^{(\ell)}; \bxi^{(\ell+1)})
\end{align}
where the last equality holds because $\vx^{(\tau(k))} = \bar{\vx}^{(\tau(k))}$ after the global averaging at iteration $\tau(k)$. With the above inequality, we have
\begin{align}
&\ \mathbb{E}[\|\vx^{(k+1)} - \bar{\vx}^{(k+1)}\|_F^2|\cF^{(\tau(k))}] \nonumber \\
=&\ \gamma^2 \mathbb{E}[\| \sum_{\ell=\tau(k)}^{k}(W - \frac{1}{n}\mathds{1}\mathds{1}^T)^{k+1-\ell} \nabla F(\vx^{(\ell)}; \bxi^{(\ell+1)}) \|_F^2|\cF^{(\tau(k))}] \nonumber \\
\le&\ 2\gamma^2 \mathbb{E}[\| \sum_{\ell=\tau(k)}^{k}(W - \frac{1}{n}\mathds{1}\mathds{1}^T)^{k+1-\ell} \nabla f(\vx^{(\ell)}) \|_F^2|\cF^{(\tau(k))}] \nonumber \\
&\ +\hspace{-0.8mm} 2 \gamma^2 \mathbb{E}\Big[ \|\sum_{\ell = \tau(k)}^k (W \hspace{-0.8mm}-\hspace{-0.8mm} \frac{1}{n}\mathds{1}\mathds{1}^T)^{k+1-\ell} [\nabla F(\vx^{(\ell)}; \bxi^{(\ell+1)}) \hspace{-0.8mm}-\hspace{-0.8mm} \nabla f(\vx^{\ell})] \|_F^2|\cF^{(\tau(k))}\big] \nonumber \\
\overset{\eqref{gd-1-appendix}}{=}&\ 2\gamma^2 \mathbb{E}[\| \sum_{\ell=\tau(k)}^{k}(W - \frac{1}{n}\mathds{1}\mathds{1}^T)^{k+1-\ell} \nabla f(\vx^{(\ell)}) \|_F^2|\cF^{(\tau(k))}] \nonumber \\
&\ +\hspace{-0.8mm} 2\gamma^2 \hspace{-2mm}\sum_{\ell = \tau(k)}^k \mathbb{E}\Big[\|(W \hspace{-0.8mm}-\hspace{-0.8mm} \frac{1}{n}\mathds{1}\mathds{1}^T)^{k+1-\ell} [\nabla F(\vx^{(\ell)}; \bxi^{(\ell+1)}) \hspace{-0.8mm}-\hspace{-0.8mm} \nabla f(\vx^{\ell})] \|_F^2|\cF^{(\tau(k))}\big] \nonumber \\
\overset{(a)}{\le}&\ 2\gamma^2 (k+1-\tau(k)) \sum_{\ell=\tau(k)}^{k} \beta^{2(k+1-\ell)} \mathbb{E}[\|\nabla f(\vx^{(\ell)})\|_F^2|\cF^{(\tau(k))}] + 2n \gamma^2 \sigma^2 \sum_{\ell = \tau(k)}^k \beta^{2(k+1-\ell)} \nonumber \\
\le&\ 2\gamma^2 H\sum_{\ell=\tau(k)}^{k} \beta^{2(k+1-\ell)} \mathbb{E}[\|\nabla f(\vx^{(\ell)})\|_F^2|\cF^{(\tau(k))}] + 2n \gamma^2 \beta^2 \sigma^2 C_\beta \label{4hyd}
\end{align}
where inequality (a) holds because of \eqref{gd-2-appendix}, \eqref{network-inequaliy} and \eqref{submulti}, and the last inequality holds because $\sum_{\ell = \tau(k)}^k \beta^{2(k+1-\ell)} \le \sum_{\ell=1}^H \beta^{2\ell} = \beta^2 (1 - \beta^{2H})/(1-\beta^2)\le \beta^2 (1 - \beta^{H})/(1-\beta) = 
\beta^2 C_\beta $ where we define $C_\beta = \sum_{\ell=0}^{H-1} \beta^{\ell} = (1 - \beta^{H})/(1-\beta)$. Note that 
\begin{align}
\|\nabla f(\vx^{(\ell)})\|_F^2 &= \|\nabla f(\vx^{(\ell)}) - \nabla f(\bar{\vx}^{(\ell)}) + \nabla f(\bar{\vx}^{(\ell)}) - \nabla f(\vx^\star) + \nabla f(\vx^\star) \|_F^2 \nonumber \\
&\le 3\|\nabla f(\vx^{(\ell)}) - \nabla f(\bar{\vx}^{(\ell)})\|_F^2 + 3 \|\nabla f(\bar{\vx}^{(\ell)}) - \nabla f(\vx^\star)\|_F^2 + 3\|\nabla f(\vx^\star)\|_F^2 \nonumber \\
&\le 3L^2\|\vx^{(\ell)} - \bar{\vx}^{(\ell)}\|_F^2 + 6nL(f(\bar{\x}^{(\ell)}) - f(x^\star)) + 3n b^2 \label{2nbsdb}
\end{align}
where the last inequality holds because of \eqref{smooth-1} and \eqref{sdu-1}. Notation $b^2$ is  defined as $b^2 = \frac{1}{n}\sum_{i=1}^n\|\nabla f_i(x^\star)\|^2$. Substituting \eqref{2nbsdb} into \eqref{4hyd}, it holds for $k = \tau(k), \tau(k)+1, \cdots, \tau(k)+H-1$ that

\begin{align}\label{p624bgsd9}
&\ \mathbb{E}[\|\vx^{(k+1)} - \bar{\vx}^{(k+1)}\|_F^2|\cF^{(\tau(k))}] \nonumber \\
\le&\ 6 H \gamma^2 L^2 \sum_{\ell=\tau(k)}^{k} \beta^{2(k+1-\ell)} \mathbb{E}[\|\vx^{(\ell)} - \bar{\vx}^{(\ell)}\|_F^2|\cF^{(\tau(k))}]\nonumber \\
&\ + 12 nH\gamma^2 L \sum_{\ell=\tau(k)}^{k}\beta^{2(k+1-\ell)} \mathbb{E}[f(\bar{\x}^{(\ell)}) - f(x^\star)|\cF^{(\tau(k))}] \hspace{-0.8mm}+\hspace{-0.8mm} 2 n \gamma^2 \beta^2 C_\beta (3Hb^2 + \sigma^2)
\end{align}
By taking expectations over the filtration $\cF^{(\tau(k))}$, we have
\begin{align}\label{0shd}
&\ \mathbb{E}\|\vx^{(k+1)} - \bar{\vx}^{(k+1)}\|_F^2 \nonumber \\
\le&\ 6 H \beta^2 \gamma^2 L^2 \sum_{\ell=\tau(k)}^{k} \beta^{2(k-\ell)} \mathbb{E}\|\vx^{(\ell)} - \bar{\vx}^{(\ell)}\|_F^2 + 12 nH \beta^2\gamma^2 L \sum_{\ell=\tau(k)}^{k}\beta^{2(k-\ell)} \mathbb{E}(f(\bar{\x}^{(\ell)}) - f(x^\star)) \nonumber \\
&\quad + 2 n \gamma^2 \beta^2 C_\beta (3Hb^2 + \sigma^2) \nonumber \\
\le&\ 6 H \beta^2 \gamma^2 L^2 \sum_{\ell=\tau(k)}^{k} \beta^{k-\ell}\, \mathbb{E}\|\vx^{(\ell)} - \bar{\vx}^{(\ell)}\|_F^2 + 12 nH \beta^2\gamma^2 L \sum_{\ell=\tau(k)}^{k}\beta^{k-\ell}\, \mathbb{E}(f(\bar{\x}^{(\ell)}) - f(x^\star)) \nonumber \\
&\quad + 2 n \gamma^2 \beta^2 C_\beta (3Hb^2 + \sigma^2)
\end{align}

\vspace{1mm}
\noindent \textbf{Bound in Lemma \ref{lm-consensus-2}}. With \eqref{xchsdh}, 
it holds for $k = \tau(k), \tau(k)+1, \cdots, \tau(k)+H-1$ that 
\begin{align}\label{23bnsdbsd}
&\ \mathbb{E}[\|\vx^{(k+1)} - \bar{\vx}^{(k+1)}\|_F^2|\cF^{(k)}] \nonumber \\
=&\ \mathbb{E}[\|(W - \frac{1}{n}\mathds{1}\mathds{1}^T)\Big( \vx^{(k)} - \bar{\vx}^{(k)} - \gamma \nabla F(\vx^{(k)};\bxi^{(k+1)})\Big)\|_F^2|\cF^{(k)}] \nonumber \\
\overset{\eqref{gd-1}}{=}&\ \|(W - \frac{1}{n}\mathds{1}\mathds{1}^T)\Big( \vx^{(k)} - \bar{\vx}^{(k)} - \gamma \nabla f(\vx^{(k)})\Big)\|_F^2 + \gamma^2\mathbb{E}[\|(W - \frac{1}{n}\mathds{1}\mathds{1}^T)\big(\nabla F(\vx^{(k)};\bxi^{(k+1)}) - \nabla f(\vx^{(k)})\big)\|_F^2|\cF^{(k)}] \nonumber \\
\le&\ \|(W - \frac{1}{n}\mathds{1}\mathds{1}^T)\Big( \vx^{(k)} - \bar{\vx}^{(k)} - \gamma \nabla f(\vx^{(k)})\Big)\|_F^2 + n \gamma^2 \beta^2 \sigma^2
\end{align}
where the last inequality holds because of \eqref{gd-2} and \eqref{network-inequaliy}. We now bound the first term:
\begin{align}\label{xnwe8sdc7}
&\ \|(W - \frac{1}{n}\mathds{1}\mathds{1}^T)\Big( \vx^{(k)} - \bar{\vx}^{(k)} - \gamma \nabla f(\vx^{(k)})\Big)\|_F^2 \nonumber \\
\overset{(a)}{\le} & \frac{1}{t}\|(W - \frac{1}{n}\mathds{1}\mathds{1}^T)\big( \vx^{(k)} - \bar{\vx}^{(k)}\big) \|_F^2 + \frac{\gamma^2}{1-t} \|(W - \frac{1}{n}\mathds{1}\mathds{1}^T)\nabla f(\vx^{(k)})\|_F^2 \nonumber \\
\overset{(b)}{=} &\ \beta \| \vx^{(k)} - \bar{\vx}^{(k)}\|_F^2 + \frac{\beta^2 \gamma^2}{1-\beta}\|\nabla f(\vx^{(k)})\|_F^2 \nonumber  \\
=&\ \beta \| \vx^{(k)} - \bar{\vx}^{(k)}\|_F^2 + \frac{\beta^2 \gamma^2}{1-\beta}\|\nabla f(\vx^{(k)}) - \nabla f(\bar{\vx}^{(k)}) + \nabla f(\bar{\vx}^{(k)}) - \nabla f(\vx^\star) + \nabla f(\vx^\star)\|_F^2 \nonumber \\
\overset{(c)}{\le}&\ \beta \| \vx^{(k)} - \bar{\vx}^{(k)}\|_F^2 + \frac{3 \beta^2 \gamma^2 L^2}{1-\beta}\|\vx^{(k)} - \bar{\vx}^{(k)}\|_F^2 + \frac{6 n \beta^2 \gamma^2 L}{1-\beta}\big( f(\bar{\x}^{(k)}) - f(x^\star) \big) + \frac{3n\beta^2 \gamma^2  b^2}{1-\beta}
\end{align}
where (a) holds because of the Jensen's inequality for any $t\in(0,1)$, (b) holds by setting $t=\beta$, and (c) holds because of \eqref{smooth-1} and \eqref{sdu-1}. Quantity $b^2 = \frac{1}{n}\sum_{i=1}^n \|\nabla f_i(x^\star)\|^2$ in the last inequality. Substituting \eqref{xnwe8sdc7} into \eqref{23bnsdbsd}, we have
\begin{align}\label{23bnsdbsd-sd7db}
&\ \mathbb{E}[\|\vx^{(k+1)} - \bar{\vx}^{(k+1)}\|_F^2|\cF^{(k)}] \nonumber \\
\le &\ \beta \| \vx^{(k)} - \bar{\vx}^{(k)}\|_F^2 + \frac{3 \beta^2 \gamma^2 L^2}{1-\beta}\|\vx^{(k)} - \bar{\vx}^{(k)}\|_F^2 + \frac{6 n \beta^2 \gamma^2 L}{1-\beta}\big( f(\bar{\x}^{(k)}) - f(x^\star) \big) + n\gamma^2\beta^2\sigma^2 + \frac{3n\beta^2 \gamma^2  b^2}{1-\beta} \nonumber \\
= &\ \beta^{k+1-\tau(k)}\| \vx^{(\tau(k))} - \bar{\vx}^{(\tau(k))}\|_F^2 + \frac{3 \beta^2 \gamma^2 L^2}{1-\beta} \sum_{\ell=\tau(k)}^{k} \beta^{k-\ell}\, \|\vx^{(\ell)} - \bar{\vx}^{(\ell)}\|_F^2 + \frac{6 n \beta^2\gamma^2 L}{1-\beta} \sum_{\ell=\tau(k)}^{k}\beta^{k-\ell}\, (f(\bar{\x}^{(\ell)}) - f(x^\star)) \nonumber \\
&\quad + n \gamma^2 \beta^2 C_\beta (\frac{3b^2}{1-\beta} + \sigma^2) \nonumber \\
= &\ \frac{3 \beta^2 \gamma^2 L^2}{1-\beta} \sum_{\ell=\tau(k)}^{k} \beta^{k-\ell}\, \|\vx^{(\ell)} - \bar{\vx}^{(\ell)}\|_F^2 + \frac{6 n \beta^2\gamma^2 L}{1-\beta} \sum_{\ell=\tau(k)}^{k}\beta^{k-\ell}\, (f(\bar{\x}^{(\ell)}) - f(x^\star)) + n \gamma^2 \beta^2 C_\beta (\frac{3b^2}{1-\beta} + \sigma^2)
\end{align}
By taking expectation over the filtration $\cF^{(k)}$, we have
\begin{align}\label{pshnsd}
&\ \mathbb{E}\|\vx^{(k+1)} - \bar{\vx}^{(k+1)}\|_F^2 \nonumber \\
\le&\ \frac{3 \beta^2 \gamma^2 L^2}{1-\beta} \sum_{\ell=\tau(k)}^{k} \beta^{k-\ell}\, \mathbb{E}\|\vx^{(\ell)} - \bar{\vx}^{(\ell)}\|_F^2 + \frac{6 n \beta^2\gamma^2 L}{1-\beta} \sum_{\ell=\tau(k)}^{k}\beta^{k-\ell}\, \mathbb{E}(f(\bar{\x}^{(\ell)}) - f(x^\star)) + n \gamma^2 \beta^2 C_\beta (\frac{3b^2}{1-\beta} + \sigma^2) \nonumber \\
< &\ \frac{6 \beta^2 \gamma^2 L^2}{1-\beta} \sum_{\ell=\tau(k)}^{k} \beta^{k-\ell}\, \mathbb{E}\|\vx^{(\ell)} - \bar{\vx}^{(\ell)}\|_F^2 + \frac{12 n \beta^2\gamma^2 L}{1-\beta} \sum_{\ell=\tau(k)}^{k}\beta^{k-\ell}\, \mathbb{E}(f(\bar{\x}^{(\ell)}) - f(x^\star)) + 2 n \gamma^2 \beta^2 C_\beta (\frac{3b^2}{1-\beta} + \sigma^2)
\end{align}

\vspace{1mm}
\subsection{Proof of Lemma \ref{lm-weighted-consensus}.} To simplify the notation, we define
\begin{align}\label{2ndnbn-test}
A^{(k)} &= \mathbb{E}\|\vx^{(k)} - \bar{\vx}^{(k)}\|_F^2, \quad B^{(k)} = \mathbb{E} f(\bar{\x}^{(k)}) - f(x^\star), \nonumber \\
 c_1 &= 6D_{\beta}  \beta^2 \gamma^2 L^2, \quad \hspace{1.05cm} c_2 = 12n D_{\beta}  \beta^2 \gamma^2 L, \quad \hspace{1.15cm}  c_3 = 2n \beta^2 \gamma^2 C_\beta (3 D_{\beta} b^2 + \sigma^2).
\end{align}
Using these notations, we rewrite \eqref{2enn} for any $k=0,1,2,\cdots$ that
\begin{align}
\begin{cases}\label{238-test}
A^{(k)} \le c_1 \sum_{\ell = \tau(k)}^{k-1} \beta^{k- 1 - \ell} A^{(\ell)} + c_2 \sum_{\ell = \tau(k)}^{k-1} \beta^{k-1-\ell} B^{(\ell)} + c_3 & \mbox{if $k > \tau(k)$} \\
A^{(k)} = 0 & \mbox{if $k = \tau(k)$}
\end{cases}
\end{align}
We next define 
\begin{align}
\Gamma_T := \{k| 0 \le k \le T \mbox{ and $\mathrm{mod}(k,H) = 0$}\}, \quad \Gamma_T^{c} := \{k| 0 \le k \le T \mbox{ and $\mathrm{mod}(k,H) \neq 0$}\}.
\end{align}
By taking the running average over both sides in \eqref{238-test} and recalling $A^{(\tau(k))} = 0$, it holds that 
\begin{align}\label{2347dsh8-test}
\sum_{k=0}^T A^{(k)}  \le&\ c_1 \sum_{k \in \Gamma_T^{c}} \sum_{\ell = \tau(k)}^{k-1}  \beta^{k-1-\ell} A^{(\ell)} + c_2\sum_{k \in \Gamma_T^{c}} \sum_{\ell = \tau(k)}^{k-1}  \beta^{k-1-\ell} B^{(\ell)} + c_3 (T+1)
\end{align}
We further define 
\begin{align}
\Theta_T := \{\ell | 0 \le \ell \le T-1 \mbox{ and $\mathrm{mod}(\ell+1, H) = 0$}\}, \quad \Theta^{c}_T := \{\ell | 0 \le \ell \le T-1 \mbox{ and $\mathrm{mod}(\ell+1,H)\neq0$}\}.
\end{align}
With these notations, we have
\begin{align}
\sum_{k=0}^T A^{(k)}  &\le c_1 \sum_{k \in \Gamma_T^{c}} \sum_{\ell = \tau(k)}^{k-1} \beta^{k-1-\ell} A^{(\ell)} +  c_2 \sum_{k \in \Gamma_T^{c}} \sum_{\ell = \tau(k)}^{k-1}  \beta^{k-1-\ell} B^{(\ell)}  +c_3  (T+1) \nonumber \\
&= c_1 \sum_{\ell \in \Theta_T^{c}} A^{(\ell)} \big(\sum_{k=\ell+1}^{\tau(\ell)+H-1} \beta^{k-1-\ell}\big) + c_2 \sum_{\ell \in \Theta_T^{c}} B^{(\ell)} \big(\sum_{k=\ell+1}^{\tau(\ell)+H-1} \beta^{k-1-\ell}\big)  +c_3  (T+1) \nonumber \\
&\overset{(a)}{\le} c_1  C_\beta \sum_{\ell \in \Theta_T^{c}} A^{(\ell)} + c_2 C_\beta \sum_{\ell \in \Theta_T^{c}} B^{(\ell)} +  c_3 (T+1) \nonumber \\
&\overset{(b)}{\le} c_1   C_\beta \sum_{k =0}^T A^{(k)} + c_2  C_\beta \sum_{k =0}^T B^{(k)}  + c_3 (T+1) \nonumber \\
&\overset{(c)}{\le} c_1  D_\beta \sum_{k =0}^T  A^{(k)} + c_2  D_\beta \sum_{k =0}^T  B^{(k)}  + c_3 (T+1)
\end{align}
where (a) holds because $\sum_{k=\ell+1}^{\tau(\ell)+H-1} {\beta}^{k-1-\ell} \le \sum_{k=0}^{H-1} {\beta}^{k} = C_\beta$, (b) holds because $A^{(k)} \ge 0$ and $B^{(k)}\ge 0$, and (c) holds because $C_\beta = (1-\beta^H)/(1-\beta) \le \min\{\frac{1}{1-\beta}, H\} = D_\beta$. If step-size $\gamma$ is sufficiently small such that $1 - c_1 D_\beta \ge \frac{1}{2}$, it holds that 
\begin{align}\label{3dbas}
\sum_{k=0}^T A^{(k)} \le 2 c_2  D_\beta \sum_{k =0}^T B^{(k)}  + 2 c_3 (T+1).
\end{align}
To guarantee $1 - c_1 D_\beta \ge \frac{1}{2}$, it is enough to let $\gamma \le {1}/(4L\beta D_\beta)$.

\subsection{Proof of Theorem \ref{thm-convex}} 
\label{proof-thm1}
Following the notation in \eqref{2ndnbn-test}, we further define $F^{(k)}:=\mathbb{E}\|\bar{\x}^{(k)} - x^\star\|_F^2$. With these notations, the inequality \eqref{23bsd999} becomes
\begin{align}
B^{(k)} \le \frac{F^{(k)}}{\gamma} - \frac{ F^{(k+1)}}{\gamma} + \frac{\gamma \sigma^2}{n} + \frac{3 L}{2n} A^{(k)}
\end{align}
Taking weighted running average over the above inequality to get
\begin{align}\label{xncxn-cn}
\frac{1}{T+1}\sum_{k=0}^T B^{(k)} &\le \frac{F^{(0)}}{(T+1)\gamma} + \frac{3 L}{2n(T+1)}\sum_{k=0}^T A^{(k)} + \frac{\gamma  \sigma^2}{n} \nonumber \\
&\overset{\eqref{3dbas}}{\le} \frac{F^{(0)}}{(T+1)\gamma}  + \frac{6L c_2 D_\beta }{n(T+1)} \sum_{k=0}^T B^{(k)} +  \frac{3 L c_3}{n} + \frac{\gamma \sigma^2}{n} \nonumber \\
&\le \frac{ F^{(0)}}{(T+1)\gamma}  + \frac{1}{2(T+1)} \sum_{k=0}^T B^{(k)} +  \frac{3 L c_3}{n} + \frac{\gamma \sigma^2}{n}
\end{align}
where the last inequality holds when $\gamma \le {1}/(12 L \beta D_\beta )$. Substituting $c_3$ into the above inequality, we have
\begin{align}\label{xncxn-cn-2-test}
\frac{1}{T+1}\sum_{k=0}^T B^{(k)} \le \frac{2 F^{(0)}}{(T+1)\gamma} + \frac{2\gamma \sigma^2}{n}
+ 12 L\beta^2 \gamma^2 C_\beta \sigma^2 + 36 L\beta^2 \gamma^2 C_\beta D_\beta b^2.
\end{align}
The way to choose step-size $\gamma$ is adapted from Lemma 15 in \cite{koloskova2020unified}. For simplicity, we let
\begin{align}\label{sndsnnds}
r_0 = 2 F^{(0)}, \quad r_1 = \frac{2\sigma^2}{n}, \quad r_2 = 12 L\beta^2 C_\beta \sigma^2 + 36 L\beta^2 C_\beta D_\beta b^2,
\end{align}
and inequality \eqref{xncxn-cn-2-test} becomes
\begin{align}\label{sndsnnds-2}
\frac{1}{T+1}\sum_{k=0}^T B^{(k)} \le \frac{r_0}{(T+1)\gamma} + r_1 \gamma
+ r_2 \gamma^2.
\end{align}

Now we let $\gamma = \min\big\{\frac{1}{12 \beta L D_\beta }, \left(\frac{r_0}{r_1(T+1)}\right)^{\frac{1}{2}}, \left(\frac{r_0}{r_2(T+1)}\right)^{\frac{1}{3}}\big\}$:
\begin{itemize}
\item If $\left(\frac{r_0}{r_2(T+1)}\right)^{\frac{1}{3}}$ is the smallest, we let $\gamma = \left(\frac{r_0}{r_2(T+1)}\right)^{\frac{1}{3}}$. With $\left(\frac{r_0}{r_2(T+1)}\right)^{\frac{1}{3}} \le \left(\frac{r_0}{r_1(T+1)}\right)^{\frac{1}{2}}$, \eqref{sndsnnds-2} becomes 
\begin{align}\label{xncxn-cn-3-test}
\frac{1}{T+1}\sum_{k=0}^T B^{(k)} \le 2 r_2^{\frac{1}{3}} \Big(\frac{r_0}{T+1}\Big)^{\frac{2}{3}} + r_1 \Big(\frac{r_0}{r_2(T+1)}\Big)^{\frac{1}{3}} \le 2r_2^{\frac{1}{3}} \Big(\frac{r_0}{T+1}\Big)^{\frac{2}{3}} +  \left(\frac{r_0 r_1}{T+1}\right)^{\frac{1}{2}}.
\end{align}

\item If $\left(\frac{r_0}{r_1(T+1)}\right)^{\frac{1}{2}}$ is the smallest, we let $\gamma = \left(\frac{r_0}{r_1(T+1)}\right)^{\frac{1}{2}}$.  With $\left(\frac{r_0}{r_1(T+1)}\right)^{\frac{1}{2}} \le \left(\frac{r_0}{r_2(T+1)}\right)^{\frac{1}{3}}$, \eqref{sndsnnds-2} becomes 
\begin{align}\label{xncxn-cn-4-test}
\frac{1}{T+1}\sum_{k=0}^T B^{(k)} \le 2 \Big(\frac{r_0 r_1}{T+1}\Big)^{\frac{1}{2}} + \frac{r_0r_2}{r_1(T+1)} \le 2 \Big(\frac{r_0 r_1}{T+1}\Big)^{\frac{1}{2}} + r_2^{\frac{1}{3}}\Big(\frac{r_0}{T+1}\Big)^{\frac{2}{3}}.
\end{align}

\item If $\frac{1}{12\beta L D_\beta } \le \left(\frac{r_0}{r_1(T+1)}\right)^{\frac{1}{2}}$ and 
$\frac{1}{12\beta L D_\beta } \le \left(\frac{r_0}{r_2(T+1)}\right)^{\frac{1}{3}}$, we let $\gamma = \frac{1}{12\beta L D_\beta }$ and \eqref{sndsnnds-2} becomes 
\begin{align}\label{xncxn-cn-sdf-test}
\frac{1}{T+1}\sum_{k=0}^T B^{(k)} \le \frac{12\beta L D_\beta r_0}{T+1} + \Big(\frac{r_0 r_1}{T+1}\Big)^{\frac{1}{2}} + r_2^{\frac{1}{3}}\Big(\frac{r_0}{T+1}\Big)^{\frac{2}{3}}.
\end{align}
\end{itemize}
Combining \eqref{xncxn-cn-3-test}, \eqref{xncxn-cn-4-test} and \eqref{xncxn-cn-sdf-test}, we have 
\begin{align}
\frac{1}{T+1}\sum_{k=0}^T B^{(k)} \le \frac{12 r_0 L D_\beta \beta}{T+1} + 2\Big(\frac{r_0 r_1}{T+1}\Big)^{\frac{1}{2}} + 2 r_2^{\frac{1}{3}}\Big(\frac{r_0}{T+1}\Big)^{\frac{2}{3}}.
\end{align}
Substituting constants $r_0$, $r_1$, and $r_2$, we have the final result:
\begin{align}\label{xbsd87}
\frac{1}{T+1}\sum_{k=0}^T B^{(k)} = O\Big( \frac{\sigma}{\sqrt{nT}} + 
\frac{(C_\beta )^{\frac{1}{3}}\beta^{\frac{2}{3}}\sigma^{\frac{2}{3}}}{T^{\frac{2}{3}}} + 
 \frac{(C_\beta )^{\frac{1}{3}} (D_\beta )^{\frac{1}{3}} \beta^{\frac{2}{3}}{b}^{\frac{2}{3}}}{T^{\frac{2}{3}}} + \frac{\beta D_\beta }{T}\Big).
\end{align}

\section{Convergence analysis for non-convex scenario }
\label{app-convg-nc}

\vspace{1mm}
\subsection{Proof Outline for Theorem \ref{thm:nc}.}

The proof outline for Theorem \ref{thm:nc} is similar to that for Theorem \ref{thm-convex}. The descent lemma \citep[Lemma~10]{koloskova2020unified} was established follows.
\begin{lemma}[{\sc Descent Lemma} \citep{koloskova2020unified}] \label{lm-descent-nc}
  Under Assumption \ref{ass:smoothness}--\ref{ass:weight-matrix} and step-size $\gamma < \frac{1}{4L}$, it holds for $k=1,2,\cdots$ that 
  \begin{align}\label{23bsd999-nc}
  \mathbb{E} f(\bar{\x}^{(k)})\le &\ \mathbb{E} f(\bar{\x}^{(k-1)}) - \frac{\gamma}{4}\mathbb{E}\|\nabla f(\bar{\x}^{(k-1)})\|^2 + \frac{\gamma^2 L \sigma^2}{2n} + \frac{3\gamma L^2}{4n} \mathbb{E}\| \vx^{(k)} - \bar{\vx}^{(k)}\|_F^2.
  \end{align}
\end{lemma}
  
The consensus distance is examined in the following two lemmas. Similar to Lemma \ref{lm-consensus-unified}, we use $D_\beta = \min\{H, 1/(1-\beta)\}$. 
\begin{lemma}[\sc Unified Consensus Lemma]\label{lm-consensus-nc} Under Assumptions \ref{ass:smoothness}--\ref{ass:weight-matrix} and \ref{ass:data-h}, it holds for $k = \tau(k), \tau(k)+1, \cdots, \tau(k)+H-1$ that 
\begin{align}\label{2enn-nc}
\mathbb{E}\|\vx^{(k+1)} - \bar{\vx}^{(k+1)}\|_F^2 \le & 6 D_\beta \beta^2 \gamma^2 L^2 \sum_{\ell=\tau(k)}^{k} \beta^{2(k+1-\ell)} \mathbb{E}\|\vx^{(\ell)} - \bar{\vx}^{(\ell)}\|_F^2 \nonumber \\
&\quad + 6 n D_\beta \beta^2 \gamma^2  \sum_{\ell=\tau(k)}^{k}\beta^{2(k+1-\ell)} \mathbb{E} \|\nabla f(\bar{\x}^{(\ell)})\|^2  + 2 n \beta^2 \gamma^2  C_\beta (3H \hat{b}^2 + \sigma^2)
\end{align}
where $D_\beta = \min\{H, \frac{1}{1-\beta}\}$.
\end{lemma} 

\begin{lemma}[\sc Running consensus lemma] \label{lm-weighted-consensus-nc}
When Assumptions \ref{ass:smoothness}--\ref{ass:weight-matrix} and \ref{ass:data-h} hold and step-size $\gamma < \frac{1}{4L\beta D_\beta }$, it holds for any $T>0$ that 
\begin{align}\label{23nbd-nc}
\frac{1}{T+1}\sum_{k=0}^T \mathbb{E}\|\vx^{(k)} - \bar{\vx}^{(k)}\|_F^2 \le\ \frac{2 c_2 D_\beta }{T+1} \sum_{k =0}^T \mathbb{E}\| \nabla f(\bar{\x}^{(k)})\|^2 + 2c_3
\end{align}
where $c_2$ and $c_3$ are constants defined as
\begin{align}
c_2 &= 6n D_\beta \beta^2 \gamma^2,  \\
c_3 &= 2n  \beta^2 \gamma^2 C_\beta (3 D_\beta \hat{b}^2 + \sigma^2).
\end{align}
\end{lemma}

With Lemmas \ref{lm-descent-nc} and \ref{lm-weighted-consensus-nc}, we can  establish the convergence rate in Theorem \ref{thm:nc}.

\subsection{Proof of Lemma \ref{lm-descent-nc}.} 
This lemma was first established in \citep[Lemma~10]{koloskova2020unified}. We made slight improvement to tight constants appeared in step-size ranges and upper bound \eqref{23bsd999-nc}. For readers' convenience, we repeat arguments here. Recall that 
\begin{align}
\bar{\x}^{{(k+1)}} = \bar{\x}^{{(k)}} -  \frac{\gamma}{n}\sum_{i=1}^n \nabla F_i(\x_i^{(k)};\bxi_i^{(k+1)}).
\end{align}
Since $f(x)$ is $L$-smooth, it holds that
\begin{align}\label{xcn}
\mathbb{E}[f(\bar{\x}^{{(k+1)}})|\cF^k] &\overset{\eqref{sdu-2}}{\le} f(\bar{\x}^{{(k)}}) - \mathbb{E} \big[\langle \nabla f(\bar{\x}^{{(k)}}),  \frac{\gamma}{n}\sum_{i=1}^n \nabla F_i(\x_i^{(k)};\bxi_i^{(k+1)})\rangle|\cF^k\big] + \frac{\gamma^2 L}{2}\mathbb{E}[\|\frac{1}{n}\sum_{i=1}^n \nabla F_i(\x_i^{(k)};\bxi_i^{(k+1)})\|^2|\cF^k] \nonumber \\
&\overset{\eqref{gd-1}}{=} f(\bar{\x}^{{(k)}}) - \langle \nabla f(\bar{\x}^{{(k)}}),  \frac{\gamma}{n}\sum_{i=1}^n \nabla f_i(\x_i^{(k)})\rangle + \frac{\gamma^2 L}{2}\mathbb{E}[\|\frac{1}{n}\sum_{i=1}^n \nabla F_i(\x_i^{(k)};\bxi_i^{(k+1)})\|^2|\cF^k] \nonumber \\
&\overset{(a)}{\le}f(\bar{\x}^{{(k)}}) - \langle \nabla f(\bar{\x}^{{(k)}}),  \frac{\gamma}{n}\sum_{i=1}^n \nabla f_i(\x_i^{(k)})\rangle + \frac{\gamma^2 L \sigma^2}{2n} +\frac{\gamma^2 L}{2} \|\frac{1}{n}\sum_{i=1}^n \nabla f_i(\x_i^{(k)})\|^2
\end{align}
where (a) holds because 
\begin{align}
&\ \mathbb{E}[\|\frac{1}{n}\sum_{i=1}^n \nabla F_i(\x_i^{(k)};\bxi_i^{(k+1)}) - \nabla f_i(\x_i^{(k)}) + \nabla f_i(\x_i^{(k)})\|^2|\cF^k] \nonumber \\
\overset{\eqref{gd-1}}{=}&\ \mathbb{E}[\|\frac{1}{n}\sum_{i=1}^n \nabla F_i(\x_i^{(k)};\bxi_i^{(k+1)}) - \nabla f_i(\x_i^{(k)})\|^2|\cF^k] + \|\frac{1}{n}\sum_{i=1}^n \nabla f_i(\x_i^{(k)})\|^2 \overset{\eqref{wendn}}{\le}\ \frac{\sigma^2}{n} + \|\frac{1}{n}\sum_{i=1}^n \nabla f_i(\x_i^{(k)})\|^2.
\end{align}
Note that
\begin{align}\label{237}
-\langle \nabla f(\bar{\x}^{{(k)}}),  \frac{\gamma}{n}\sum_{i=1}^n \nabla f_i(\x_i^{(k)})\rangle &= -\langle \nabla f(\bar{\x}^{{(k)}}),  \frac{\gamma}{n}\sum_{i=1}^n [\nabla f_i(\x_i^{(k)}) - \nabla f_i(\bar{\x}^{(k)}) + \nabla f_i(\bar{\x}^{(k)})]\rangle \nonumber \\
&\le -\gamma \|\nabla f(\bar{\x}^{{(k)}})\|^2 + \frac{\gamma}{2}\|\nabla f(\bar{\x}^{{(k)}})\|^2 + \frac{\gamma}{2n}\sum_{i=1}^n\|\nabla f_i(\x_i^{(k)}) - \nabla f_i(\bar{\x}^{(k)})\|^2 \nonumber \\
&\le -\frac{\gamma}{2}\|\nabla f(\bar{\x}^{{(k)}})\|^2 + \frac{\gamma L^2}{2n}\|\vx^{(k)} - \bar{\vx}^{(k)}\|_F^2
\end{align}
and 
\begin{align}\label{2nd}
\|\frac{1}{n}\sum_{i=1}^n \nabla f_i(\x_i^{(k)})\|^2 \le \frac{2L^2}{n}\|\vx^{(k)} - \bar{\vx}^{(k)}\|_F^2 + 2\|\nabla f(\bar{\x}^{(k)})\|^2
\end{align}
Substituting \eqref{237} and \eqref{2nd} into \eqref{xcn}, taking expectations over $\cF^{(k)}$ and using the fact that $\gamma < \frac{1}{4L}$, we reach the result in \eqref{23bsd999-nc}.

\vspace{2mm}
\subsection{Proof of Lemma \ref{lm-consensus-nc}.} 

Similar to the proof of Lemmas \ref{lm-consensus-I} and \ref{lm-consensus-2}, we will derive two bounds for $\mathbb{E}\|\vx^{(k+1)} - \bar{\vx}^{(k+1)}\|_F^2$:

\vspace{1mm}
\noindent \textbf{Bound 1}. Following \eqref{2nds-test}-\eqref{4hyd}, it holds for $k = \tau(k), \tau(k)+1, \cdots, \tau(k)+H-1$ that 
\begin{align}
\mathbb{E}[\|\vx^{(k+1)} - \bar{\vx}^{(k+1)}\|_F^2|\cF^{(\tau(k))}] \le 2 \gamma^2 H\sum_{\ell=\tau(k)}^{k} \beta^{2(k+1-\ell)} \mathbb{E}[\|\nabla f(\vx^{(\ell)})\|_F^2|\cF^{(\tau(k))}] + 2 n \gamma^2 \beta^2 \sigma^2 C_\beta \label{4hyd-nc}
\end{align}
Note that
\begin{align}\label{23476}
\|\nabla f(\vx^{(k)})\|_F^2 &= \sum_{i=1}^n\|\nabla f_i(\x_i^{(k)})\|^2 \nonumber \\
& = \sum_{i=1}^n\|\nabla f_i(\x_i^{(k)}) - \nabla f_i(\bar{\x}^{(k)}) + \nabla f_i(\bar{\x}^{(k)}) - \nabla f(\bar{\x}^k) + \nabla f(\bar{\x}^k)\|^2 \nonumber \\
&\le 3L^2 \|\vx^{(k)} - \bar{\vx}^{(k)}\|_F^2 + 3n\hat{b}^2 + 3n \|\nabla f(\bar{\x}^k)\|^2.
\end{align}
where the last inequality holds because of Assumption \ref{ass:data-h}. Substituting \eqref{23476} into \eqref{4hyd-nc} and taking expectation on $\cF^{(\tau(k))}$, we get
\begin{align}\label{xnsdh}
&\ \mathbb{E}\|\vx^{(k+1)} - \bar{\vx}^{(k+1)}\|_F^2 \nonumber \\
\le&\ 6 H \beta^2 \gamma^2 L^2 \sum_{\ell=\tau(k)}^{k} \beta^{2(k-\ell)} \mathbb{E}\|\vx^{(\ell)} - \bar{\vx}^{(\ell)}\|_F^2 \hspace{-0.8mm}+\hspace{-0.8mm} 6 nH \beta^2 \gamma^2 \sum_{\ell=\tau(k)}^{k}\beta^{2(k-\ell)} \|\nabla f(\bar{\x}^{(\ell)})\|^2  \hspace{-0.8mm}+\hspace{-0.8mm} 2n \gamma^2 \beta^2 C_\beta (3H \hat{b}^2 + \sigma^2) \nonumber \\
\le&\ 6 H \gamma^2 L^2 \beta^2  \sum_{\ell=\tau(k)}^{k} \beta^{k-\ell}\, \|\vx^{(\ell)} - \bar{\vx}^{(\ell)}\|_F^2 \hspace{-0.8mm}+\hspace{-0.8mm} 6 nH\gamma^2 \beta^2  \sum_{\ell=\tau(k)}^{k}\beta^{k-\ell} \|\nabla f(\bar{\x}^{(\ell)})\|^2 \hspace{-0.8mm}+\hspace{-0.8mm} 2n \gamma^2 \beta^2 C_\beta (3H \hat{b}^2 + \sigma^2)
\end{align}

\noindent \textbf{Bound 2}. Following \eqref{23bnsdbsd} and first two lines in \eqref{xnwe8sdc7}, it holds for any $k = \tau(k), \cdots, \tau(k)+H-1$ that 
\begin{align}\label{xn2we4}
\mathbb{E}[\|\vx^{(k+1)} - \bar{\vx}^{(k+1)}\|_F^2|\cF^{(k)}] \le \beta \| \vx^{(k)} - \bar{\vx}^{(k)}\|_F^2 + \frac{\beta^2 \gamma^2}{1-\beta}\|\nabla f(\vx^{(k)})\|_F^2 + n \gamma^2 \beta^2 \sigma^2.
\end{align}
Substituting \eqref{23476} into \eqref{xn2we4}, we get
\begin{align}
&\ \mathbb{E}[\|\vx^{(k+1)} - \bar{\vx}^{(k+1)}\|_F^2|\cF^{(k)}] \le \big(\beta + \frac{3\beta^2\gamma^2 L^2}{1-\beta}\big) \| \vx^{(k)} - \bar{\vx}^{(k)}\|_F^2 + \frac{3n\beta^2 \gamma^2\|\nabla f(\bar{\x}^k)\|^2}{1-\beta} + n \gamma^2 \beta^2 \sigma^2 + \frac{3n\beta^2 \gamma^2 \hat{b}^2}{1-\beta}.
\end{align}
We next follow \eqref{23bnsdbsd}--\eqref{pshnsd} and take expectation on $\cF^{(k)}$ to get  
\begin{align}\label{0shd-psd6-234}
&\ \mathbb{E}\|\vx^{(k+1)} - \bar{\vx}^{(k+1)}\|_F^2 \nonumber \\
\le&\ \frac{3 \beta^2 \gamma^2 L^2}{1-\beta} \sum_{\ell=\tau(k)}^{k} \beta^{k-\ell}\, \mathbb{E}\|\vx^{(\ell)} - \bar{\vx}^{(\ell)}\|_F^2 + \frac{3 n  \beta^2\gamma^2}{1-\beta} \sum_{\ell=\tau(k)}^{k}\beta^{k-\ell}\, \mathbb{E}\| \nabla f(\bar{\x}^{(\ell)}) \|^2 + n \gamma^2 \beta^2 C_\beta ( \frac{3 \hat{b}^2}{1-\beta} + \sigma^2) \nonumber \\
\le &\ \frac{6 \beta^2 \gamma^2 L^2}{1-\beta} \sum_{\ell=\tau(k)}^{k} \beta^{k-\ell}\, \mathbb{E}\|\vx^{(\ell)} - \bar{\vx}^{(\ell)}\|_F^2 + \frac{6 n  \beta^2\gamma^2}{1-\beta} \sum_{\ell=\tau(k)}^{k}\beta^{k-\ell}\, \mathbb{E}\| \nabla f(\bar{\x}^{(\ell)}) \|^2 + 2 n \gamma^2 \beta^2 C_\beta ( \frac{3 \hat{b}^2}{1-\beta} + \sigma^2)
\end{align}
With bounds \eqref{xnsdh} and \eqref{0shd-psd6-234}, we reach the result \eqref{2enn-nc}.

\vspace{2mm}
\subsection{Proof of Lemma \ref{lm-weighted-consensus-nc}.} We first simplify the notation as follows: 
\begin{align}\label{2ndnbn-nc}
A^{(k)} &= \mathbb{E}\|\vx^{(k)} - \bar{\vx}^{(k)}\|^2, \quad B^{(k)} = \mathbb{E} \|\nabla f(\bar{\x}^{(k)})\|^2, \nonumber \\
c_1 &= 6D_\beta \beta^2 \gamma^2 L^2, \quad \hspace{1.05cm} c_2 = 6 n D_\beta \beta^2 \gamma^2, \quad \hspace{0.7cm}  c_3 = 2n  \beta^2 \gamma^2 C_\beta (3 D_\beta \hat{b}^2 + \sigma^2).
\end{align}
With these notations, we can follow the proof of Lemma \ref{lm-weighted-consensus} to get the final result.

\vspace{2mm}
\subsection{Proof of Theorem \ref{thm:nc}.} Following the notation in \eqref{2ndnbn-nc}, we further define $F^{(k)}:=\mathbb{E}f(\bar{\x}^{(k)})$. With these notations, the inequality \eqref{23bsd999-nc} becomes
\begin{align}
B^{(k)} \le \frac{4 F^{(k)}}{\gamma} - \frac{4 F^{(k+1)}}{\gamma} + \frac{2\gamma L \sigma^2}{n} + \frac{3 L^2}{n} A^{(k)}
\end{align}
Taking the weighted running average over the above inequality and divide $T+1$ to get
\begin{align}\label{xncxn-cn-sdfh}
\frac{1}{T+1}\sum_{k=0}^T B^{(k)} &\le \frac{4F^{(0)}}{(T+1)\gamma} + \frac{3 L^2}{n(T+1)}\sum_{k=0}^T A^{(k)} + \frac{2\gamma L \sigma^2}{n} \nonumber \\
&\overset{\eqref{23nbd-nc}}{\le} \frac{4F^{(0)}}{(T+1)\gamma}  + \frac{6\beta^2 L^2 H c_2}{n(T+1)} \sum_{k=0}^T B^{(k)} +  \frac{6L^2c_3}{n} + \frac{2\gamma L \sigma^2}{n} \nonumber \\
&\le \frac{4F^{(0)}}{(T+1)\gamma}  + \frac{1}{2(T+1)} \sum_{k=0}^T B^{(k)} +  \frac{6L^2c_3}{n} + \frac{2\gamma L \sigma^2}{n}
\end{align}
where the last inequality holds when $\gamma \le \frac{1}{9 L H\beta}$. Substituting $c_3$ into the above inequality, we have
\begin{align}\label{xncxn-cn-2}
\frac{1}{T+1}\sum_{k=0}^T B^{(k)} \le \frac{8 F^{(0)}}{(T+1)\gamma} + \frac{4\gamma L \sigma^2}{n} +   24 L^2 \gamma^2 \beta^2 C_\beta \sigma^2 +  72 L^2 \gamma^2 \beta^2 C_\beta D_\beta \hat{b}^2.
\end{align}
By following the arguments \eqref{sndsnnds} -- \eqref{xbsd87}, we reach the result in \eqref{2bnsdb}.

\section{Transient stage and transient time}
\label{app-tran-time}
\subsection{Transient stage derivation.}

\vspace{0.5mm}
\noindent \textbf{(i) Gossip SGD}. We first consider the iid scenario where $b^2 = 0$. To make the first term dominate the other terms (see the first line in Table \ref{table-comparison-with-Gossip-SGD}), $T$ has to be sufficiently large such that (ignoring the affects of $\sigma$)
\begin{align}\label{xcnxcncxn-246}
\max\big\{\frac{\beta^{\frac{2}{3}}}{T^{\frac{2}{3}}(1-\beta)^\frac{1}{3}}, \frac{\beta}{(1-\beta)T}\big\} \le   \frac{1}{\sqrt{nT}} \quad \Longrightarrow \quad T \ge \max\big\{ \frac{n^3 \beta^4}{(1-\beta)^2}, \frac{n \beta^2}{(1-\beta)^2} \big\}. 
\end{align}
We next consider the non-iid scenario where $b^2 \neq 0$. To make the first term dominate the other terms, $T$ has to be sufficiently large such that (ignoring the affects of $\sigma$ and $b$)
\begin{align}\label{xcnxcncxn-2456}
\max\big\{\frac{\beta^{\frac{2}{3}}}{T^{\frac{2}{3}}(1-\beta)^\frac{1}{3}}, \frac{\beta^{\frac{2}{3}}}{T^{\frac{2}{3}}(1-\beta)^\frac{2}{3}}, \frac{\beta}{(1-\beta)T}\big\} \le   \frac{1}{\sqrt{nT}} \quad \Longrightarrow \quad T \ge \max\big\{ \frac{n^3 \beta^4}{(1-\beta)^2}, \frac{n^3 \beta^4}{(1-\beta)^4}, \frac{n \beta^2}{(1-\beta)^2} \big\}. 
\end{align}
When $n \beta > 1$ which usually holds for most commonly-used network topologies, inequalities \eqref{xcnxcncxn-246} and \eqref{xcnxcncxn-2456} will result in the transient stage $T = \Omega(\frac{n^3 \beta^4}{(1-\beta)^2})$ and $T = \Omega(\frac{n^3 \beta^4}{(1-\beta)^4})$ for iid and non-iid scenarios, respectively.

\vspace{0.5mm}
\noindent \textbf{(ii) Gossip-PGA}. We first consider the iid scenario where $b^2 = 0$. To make the first term dominate the other terms (see the first line in Table \ref{table-comparison-with-Gossip-SGD}), $T$ has to be sufficiently large such that (ignoring the affects of $\sigma$)
\begin{align}\label{xcnxcncxn-246-pga}
\max\big\{\frac{C_\beta^{\frac{1}{3}}\beta^{\frac{2}{3}}}{T^{\frac{2}{3}}}, \frac{\beta D_\beta}{T}\big\} \le   \frac{1}{\sqrt{nT}} \quad \Longrightarrow \quad T \ge \max\big\{ n^3 \beta^4 C_\beta^2, n \beta^2 D_\beta^2 \big\} = \Omega(n^3 \beta^4 C_\beta^2). 
\end{align}
We next consider the non-iid scenario where $b^2 \neq 0$. To make the first term dominate the other terms, $T$ has to be sufficiently large such that (ignoring the affects of $\sigma$ and $b$)
\begin{align}\label{xcnxcncxn-2456-pga}
\max\big\{\frac{C_\beta^{\frac{1}{3}}\beta^{\frac{2}{3}}}{T^{\frac{2}{3}}}, \frac{C_\beta^{\frac{1}{3}} D_\beta^{\frac{1}{3}} \beta^{\frac{2}{3}}}{T^{\frac{2}{3}}}, \frac{\beta D_\beta}{T}\big\} \le   \frac{1}{\sqrt{nT}} \quad \Longrightarrow \quad T \ge \max\big\{ n^3 \beta^4 C_\beta^2, n^3 \beta^4 C_\beta^2 D_\beta^2, n \beta^2 D_\beta^2 \big\} =  \Omega(n^3 \beta^4 C_\beta^2 D_\beta^2)
\end{align}
when $n\beta > 1$.

\vspace{0.5mm}
\noindent \textbf{(iii) Local SGD}. We first consider the iid scenario where $b^2 = 0$. To make the first term dominate the other terms (see the first line in Table \ref{table-comparison-with-Gossip-SGD}), $T$ has to be sufficiently large such that (ignoring the affects of $\sigma$)
\begin{align}\label{xcnxcncxn-246-local}
\max\big\{\frac{H^{\frac{1}{3}}}{T^{\frac{2}{3}}}, \frac{H}{T}\big\} \le   \frac{1}{\sqrt{nT}} \quad \Longrightarrow \quad T \ge \max\big\{ n^3 H^2, n H^2 \big\} = \Omega(n^3 H^2). 
\end{align}
We next consider the non-iid scenario where $b^2 \neq 0$. To make the first term dominate the other terms, $T$ has to be sufficiently large such that (ignoring the affects of $\sigma$ and $b$)
\begin{align}\label{xcnxcncxn-2456-local}
\max\big\{\frac{H^{\frac{1}{3}}}{T^{\frac{2}{3}}}, \frac{H^{\frac{2}{3}}}{T^{\frac{2}{3}}}, \frac{H}{T}\big\} \le   \frac{1}{\sqrt{nT}} \quad \Longrightarrow \quad T \ge \max\big\{ n^3 H^2, n^3 H^4, n H^2 \big\} =  \Omega(n^3 H^4)
\end{align}

\vspace{1mm}
\subsection{Transient time comparison}
\vspace{0.5mm}

The transient time comparisons between Gossip SGD and Gossip-PGA for the iid or non-iid scenario over the grid or ring topology are listed in Tables  \ref{table-specfic-example-grid-iid}, \ref{table-specfic-example-ring-noniid} and \ref{table-specfic-example-ring-iid}.

\begin{table}[b!]
\vskip 0.15in
\begin{center}
\begin{small}
\begin{sc}
\begin{tabular}{lcc}
\toprule
& Gossip SGD & Gossip-PGA  \\
\midrule
Transient iter. & $O(n^5)$                     & $O(n^4)$                           \\
Single comm.    & $O(\theta d + \alpha)$       & $O(\theta d + \sqrt{n}\alpha)$     \\
Transient time      & $O(n^5\theta d + n^5\alpha)$ & $O(n^4 \theta d + n^{4.5} \alpha)$ \\ \bottomrule
\end{tabular}
\end{sc}
\end{small}
\end{center}
\vskip -0.1in
\caption{Transient time comparison between Gossip SGD and Gossip-PGA for iid scenario over the specific grid ($1-\beta = O(1/n)$) topologiy. We choose $H=\sqrt{n}$ as the period in Gossip-PGA.}
\label{table-specfic-example-grid-iid}
\end{table}

\begin{table}[h!]
\vskip 0.15in
\begin{center}
\begin{small}
\begin{sc}
\begin{tabular}{lcc}
\toprule
& Gossip SGD & Gossip-PGA  \\
\midrule
Transient iter. & $O(n^{11})$                     & $O(n^5)$                           \\
Single comm.    & $O(\theta d + \alpha)$       & $O(\theta d + \sqrt{n}\alpha)$     \\
Transient time      & $O(n^{11}\theta d + n^{11}\alpha)$ & $O(n^5 \theta d + n^{5.5} \alpha)$ \\ \bottomrule
\end{tabular}
\end{sc}
\end{small}
\end{center}
\caption{Transient time comparison between Gossip SGD and Gossip-PGA for non-iid scenario over the specific ring ($1-\beta = O(1/n^2)$) topologiy. We choose $H=\sqrt{n}$ as the period in Gossip-PGA.}
\label{table-specfic-example-ring-noniid}
\vskip -0.1in
\end{table}

\begin{table}[h!]
\vskip 0.15in
\begin{center}
\begin{small}
\begin{sc}
\begin{tabular}{lcc}
\toprule
& Gossip SGD & Gossip-PGA  \\
\midrule
Transient iter. & $O(n^{7})$                     & $O(n^4)$                           \\
Single comm.    & $O(\theta d + \alpha)$       & $O(\theta d + \sqrt{n}\alpha)$     \\
Transient time      & $O(n^{7}\theta d + n^{7}\alpha)$ & $O(n^4 \theta d + n^{4.5} \alpha)$ \\ \bottomrule
\end{tabular}
\end{sc}
\end{small}
\end{center}
\caption{Transient time comparison between Gossip SGD and Gossip-PGA for iid scenario over the specific ring ($1-\beta = O(1/n^2)$) topologiy. We choose $H=\sqrt{n}$ as the period in Gossip-PGA.}
\label{table-specfic-example-ring-iid}
\vskip -0.1in
\end{table}

\section{Proof of Corollary 1}
\label{app-corollary-1}
The proof of Corollary 1 closely follows Theorem \ref{thm-convex}. First, the descent lemma \ref{lm-consensus-nc} still holds for time-varying period. Second, with the facts that $k+1 - \tau(k) \le H_{\rm max}$, $\sum_{\ell=\tau(k)}^k \beta^{k+1-\ell} \le \sum_{k=0}^{H_{\rm max}}\beta^k := C_\beta$, and $\sum_{k=\ell+1}^{\tau(\ell) + H^{(\ell)}-1}\beta^{k-1-\ell} \le \sum_{k=0}^{H_{\rm max}}\beta^k = C_\beta$, we follow Appendix C.3 and C.4 to reach the consensus distance inequality:
\begin{align}\label{23nbd-nc-adaptive}
\frac{1}{T+1}\sum_{k=0}^T \mathbb{E}\|\vx^{(k)} - \bar{\vx}^{(k)}\|^2 \le\ \frac{2 c_2 D_\beta }{T+1} \sum_{k =0}^T \mathbb{E}\| \nabla f(\bar{\x}^{(k)})\|^2 + 2c_3
\end{align}
where $c_2$ and $c_3$ are constants defined as
\begin{align}
c_2 &= 6n D_\beta \beta^2 \gamma^2,  \\
c_3 &= 2 n  \beta^2 \gamma^2 C_\beta (3 D_\beta \hat{b}^2 + \sigma^2)
\end{align}
and $D_\beta = \min\{1/(1-\beta), H_{\rm max}\}$, $C_\beta = \sum_{k=0}^{H_{\rm max}}\beta^k$. With Lemma \ref{lm-descent-nc} and inequality \eqref{23nbd-nc-adaptive}, we can follow Appendix B.5 to reach the result in Corollary 1.

\section{Additional Experiments}
\label{app-add-experiments}
\vspace{1mm}
\subsection{Implementation Details.}

\vspace{0.5mm}

We implement all the aforementioned algorithms with PyTorch \cite{paszke2019pytorch} 1.5.1 using NCCL 2.5.7 (CUDA 10.1) as the communication backend. Each server contains 8 V100 GPUs in our cluster and is treated as one node. The inter-node network fabrics are chosen from 25 Gbps TCP (which is a common distributed training platform setting) and 4$\times$100 Gbps RoCE (which is a high-performance distributed training platform setting). 

All deep learning experiments are trained in the mixed precision using Pytorch extension package NVIDIA apex (https://github.com/NVIDIA/apex). For Gossip SGD related training, we use the time-varying one-peer exponential graph following \cite{assran2019stochastic}. Workers send and receive a copy of the model's parameters to and from its peer, thus keeping the load balancing among workers. All data are stored in the cloud storage service and downloaded to workers using HTTP during training.
\vspace{0.5mm}

\noindent \textbf{Image Classfication} The Nesterov momentum SGD optimizer is used with a linear scaling learning rate strategy. 32 nodes (each node is with 8 V100 GPUs) are used in all the experiments and the batch-size is set as 256 per node (8,192 in total). The learning rate is warmed up in the first 5 epochs and is decayed by a factor of 10 at 30, 60 and 90 epochs.  We train 120 epochs by default (unless specified otherwise) in every experiment and record the epoch and runtime when a 76\% top-1 accuracy in the validation set has reached. 25 Gbps TCP network is used for inter-node communication in ResNet-50 training. In 4$\times$100 Gbps RoCE network, the communication overhead is negligible given the high computation-to-communication ratio nature of ResNet models and Parallel SGD with computation and communication pipeline is recommended. We use a period 6 for both Local SGD and Gossip-PGA. In Gossip-AGA, the averaging period is set to 4 in the warm-up stage and changed adaptively afterwards, roughly 9\% iterations conduct global averaging. 
\vspace{0.5mm}

\noindent \textbf{Language Modeling} All experiments are based on NVIDIA BERT implementation with mixed precision support and LAMB optimizer \cite{you2019large}. 8 nodes are used in all the experiments with a batch-size 64 per GPU (4096 in total). We do not use gradient accumulation as it is not vertical with Local SGD. We only do phase 1 training and indicate the decreasing of training loss as convergence speed empirically. The learning rate is scaled to $3.75e^{-4}$ initially and decayed in a polynomial policy with warm-up. The phase 1 training consists of 112,608 steps in all experiments. We use a period 6 for both Local SGD and Gossip-PGA. In Gossip-AGA, the averaging period is set to 4 in the warm-up phase and changed adaptively afterwards, roughly 9.6\% iterations conduct global averaging.

\begin{figure*}[t!]
\vskip 0.2in
\centering
\includegraphics[width=\textwidth]{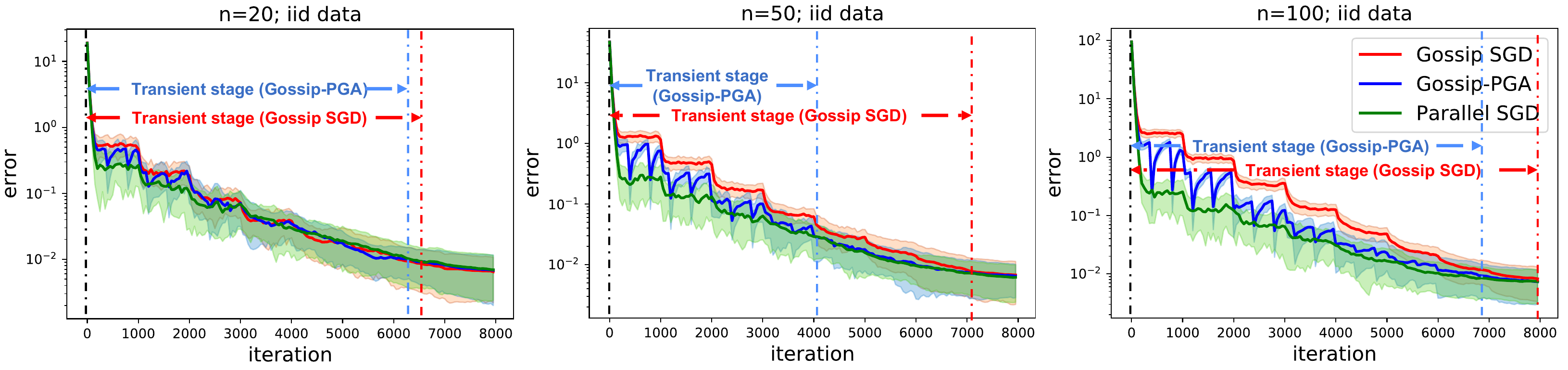} 
\caption{\small Convergence comparison between Gossip-PGA, Gossip SGD and parallel SGD on the logistic regression problem in iid data distributed setting over the ring topology.}
\label{Fig:convex-comparison-gossip-appendix-iid}
\vskip -0.2in
\end{figure*}

\begin{figure*}[t!]
\vskip 0.2in
\centering
\includegraphics[scale=0.4]{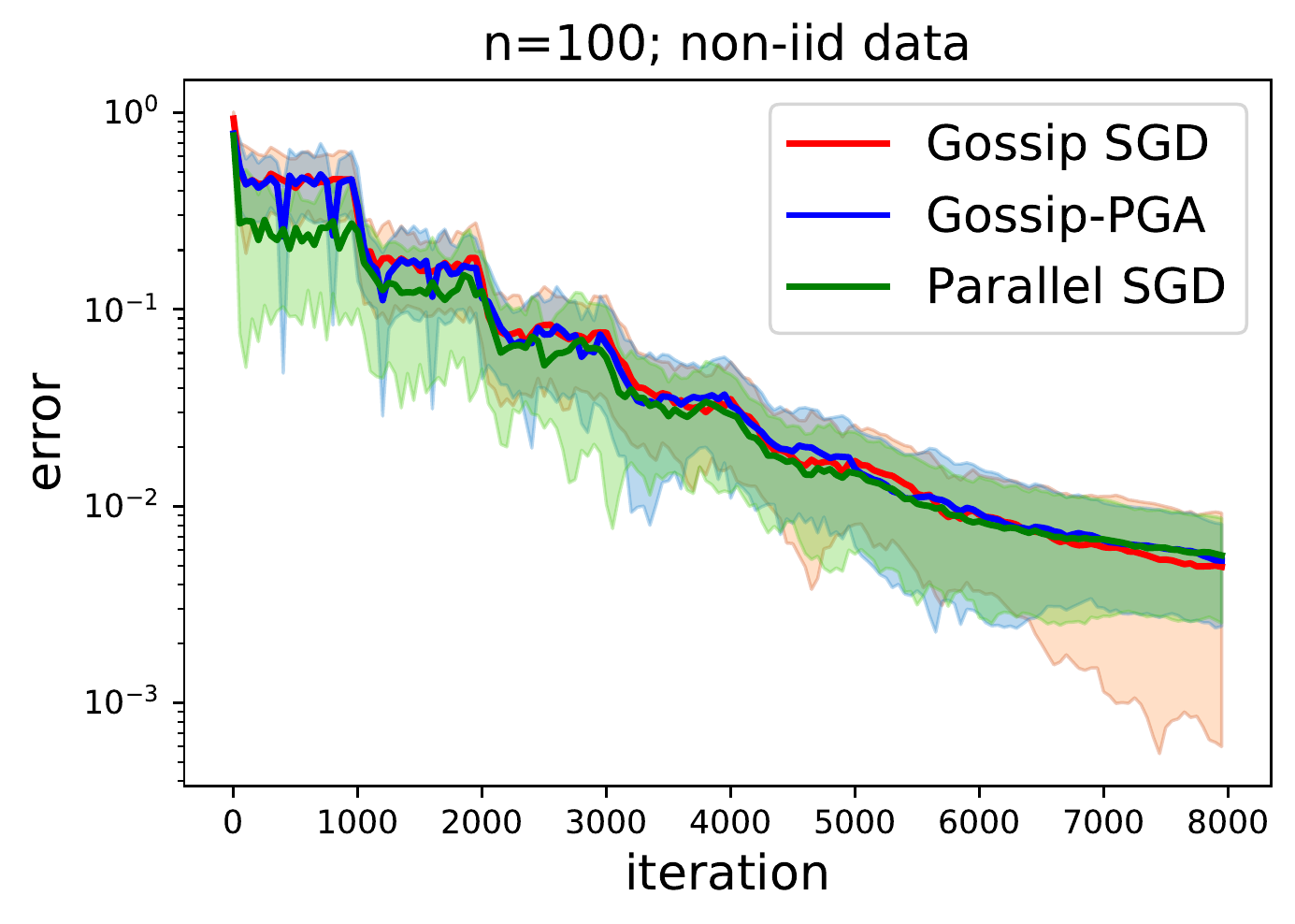} 
\includegraphics[scale=0.4]{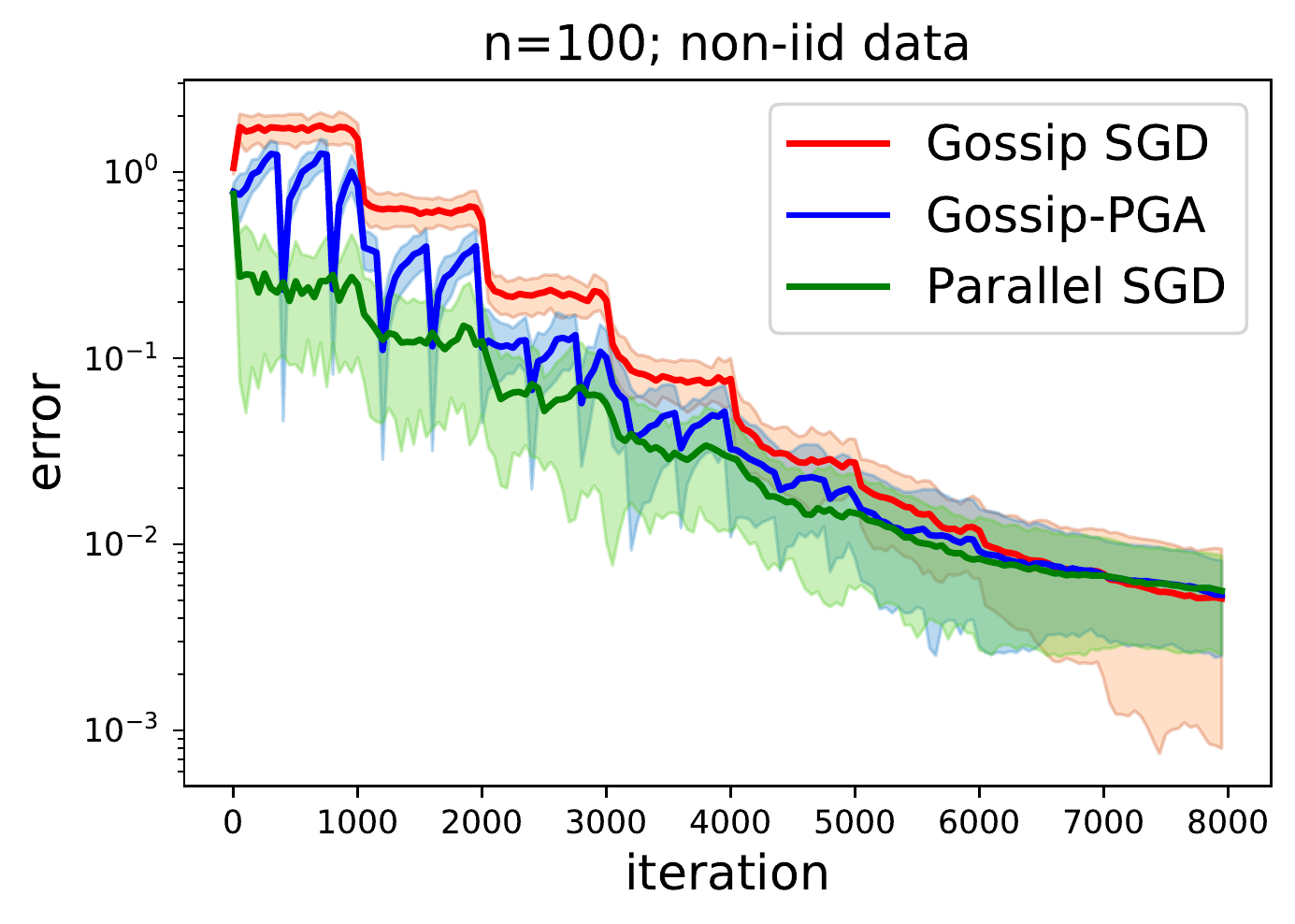} 
\includegraphics[scale=0.4]{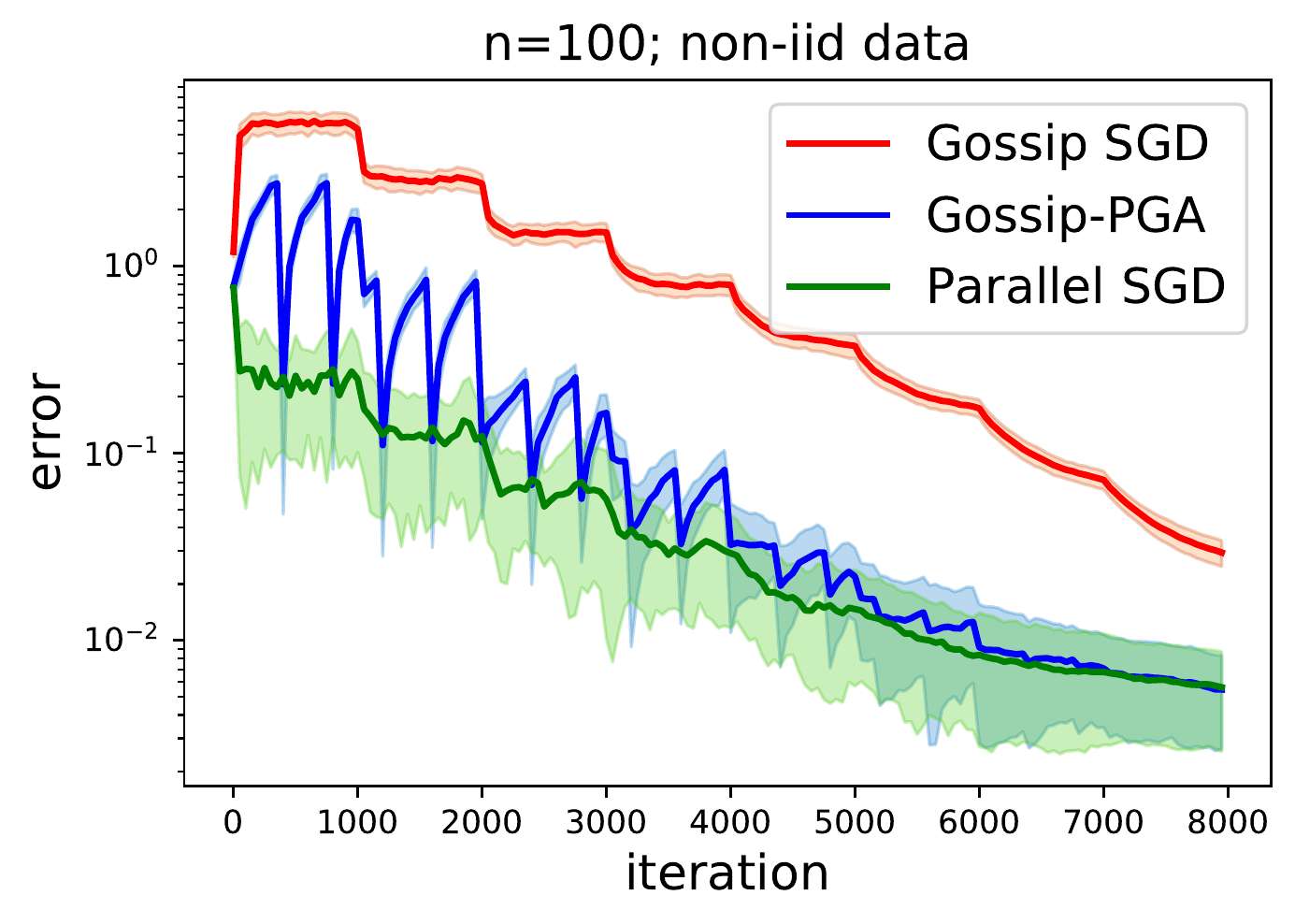} 
\caption{\small Convergence comparison between Gossip-PGA, Gossip SGD and parallel SGD on the logistic regression problem in non-iid data distributed setting over the exponential graph (left), grid (middle) and ring (right) topology.}
\label{Fig:convex-comparison-gossip-appendix-noniid-different-topo}
\vskip -0.2in
\end{figure*}

\begin{figure*}[t!]
\vskip 0.2in
\centering
\includegraphics[scale=0.4]{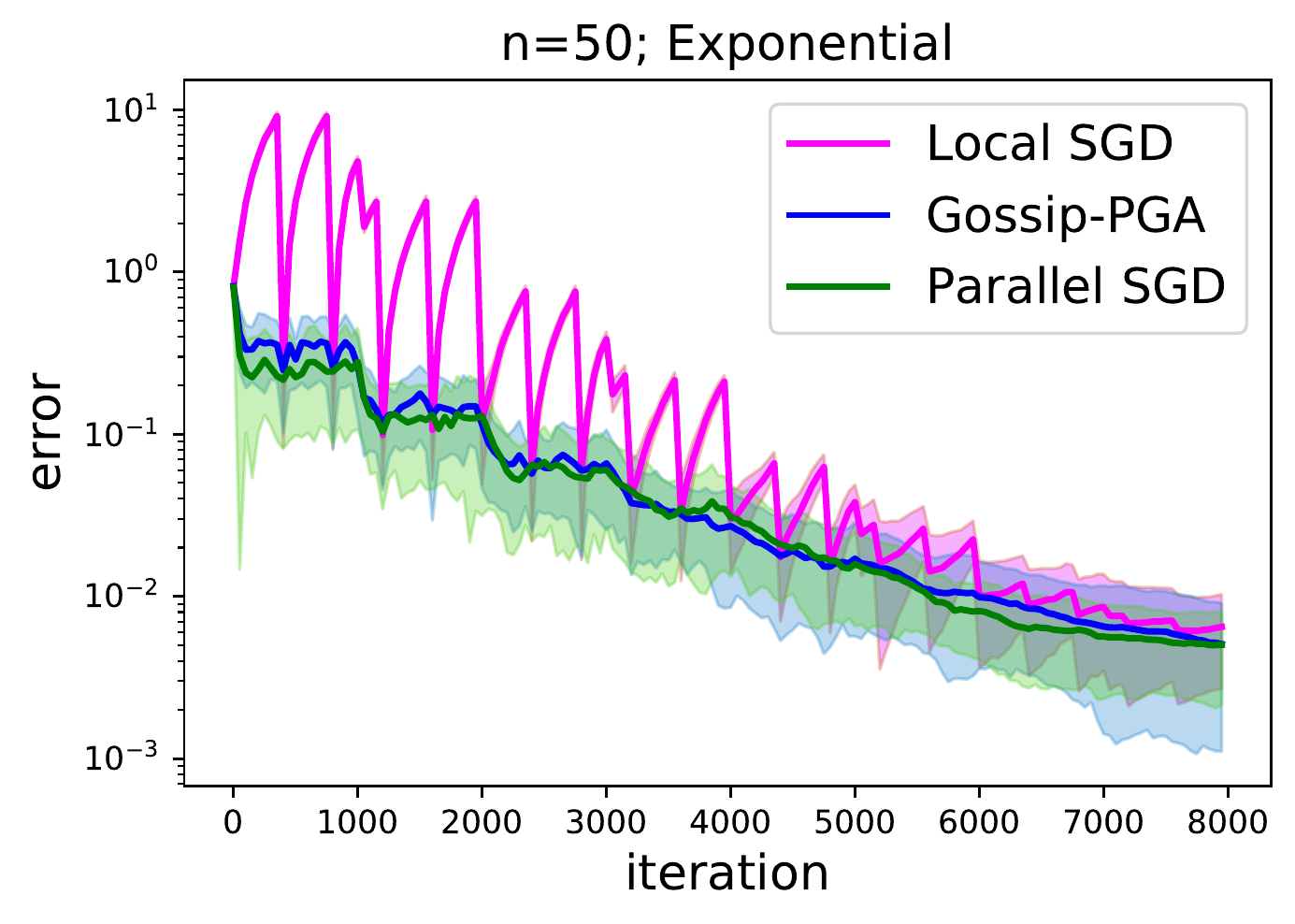} 
\includegraphics[scale=0.4]{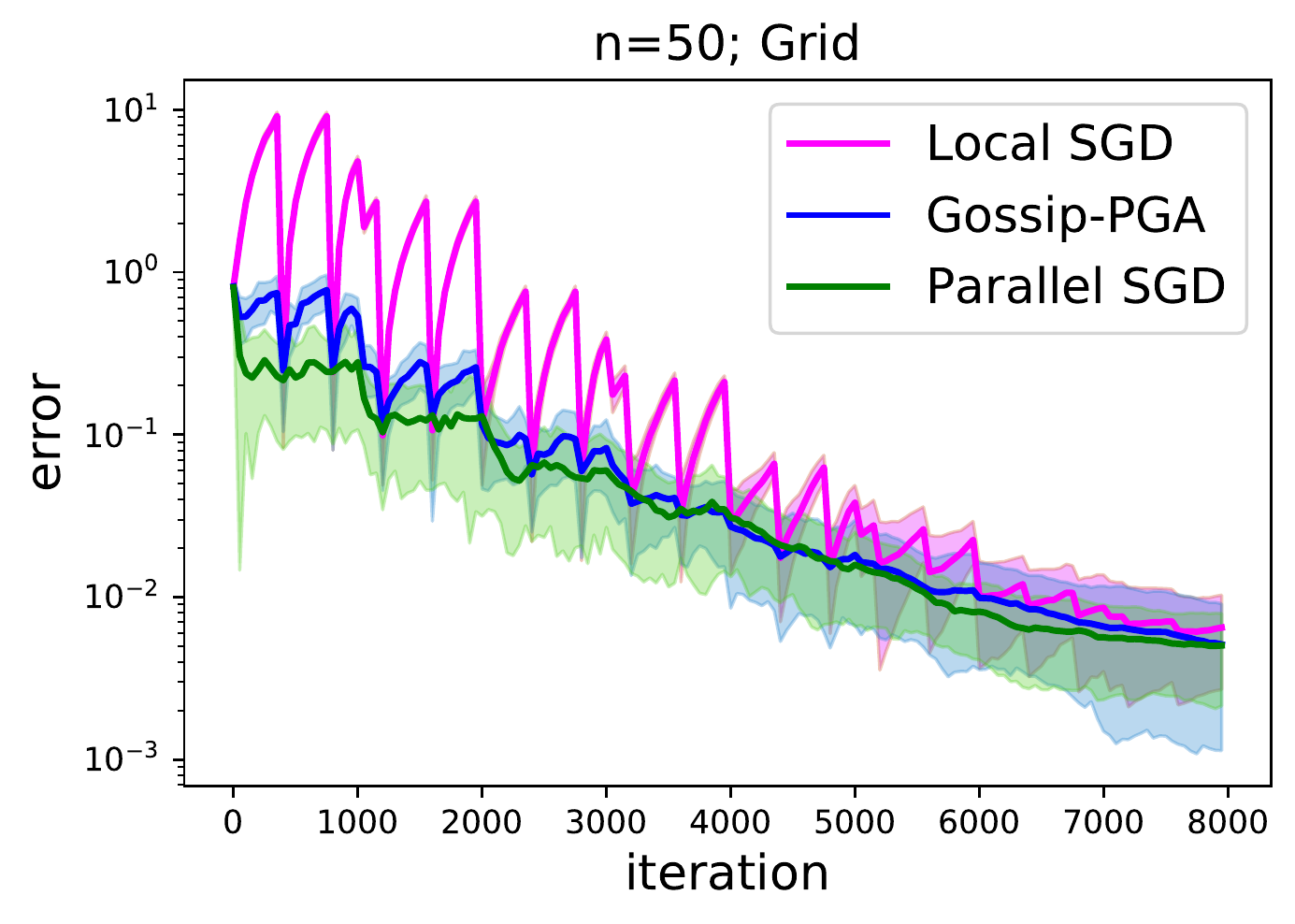} 
\includegraphics[scale=0.4]{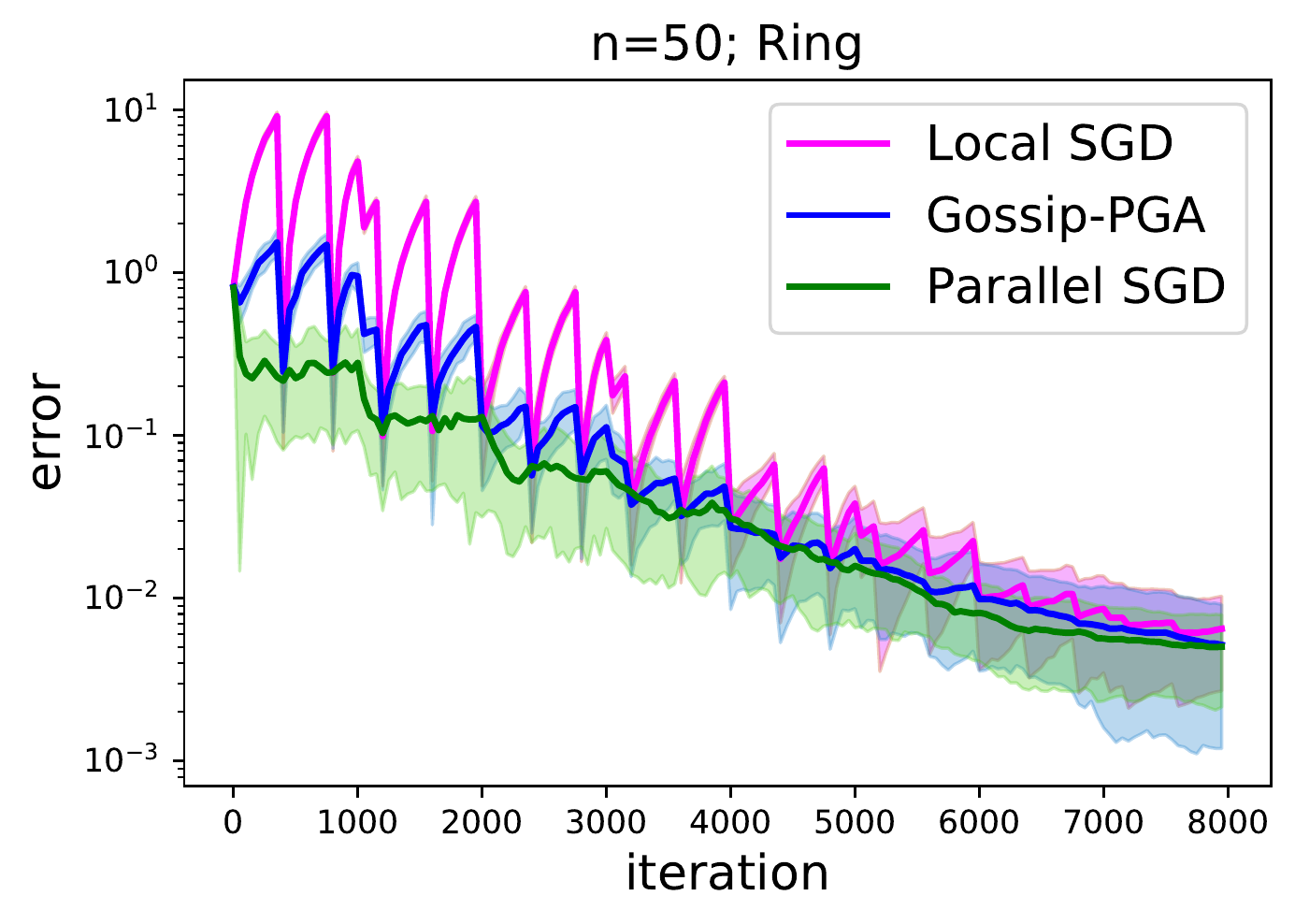} 
\caption{\small Convergence comparison between Gossip-PGA, Local SGD and parallel SGD on the logistic regression problem in non-iid data distributed setting over the exponential graph (left), grid (middle) and ring (right) topology.}
\label{Fig:convex-comparison-local-appendix-noniid-different-topo}
\vskip -0.2in
\end{figure*}

\begin{figure*}[t!]
\vskip 0.2in
\centering
\includegraphics[scale=0.4]{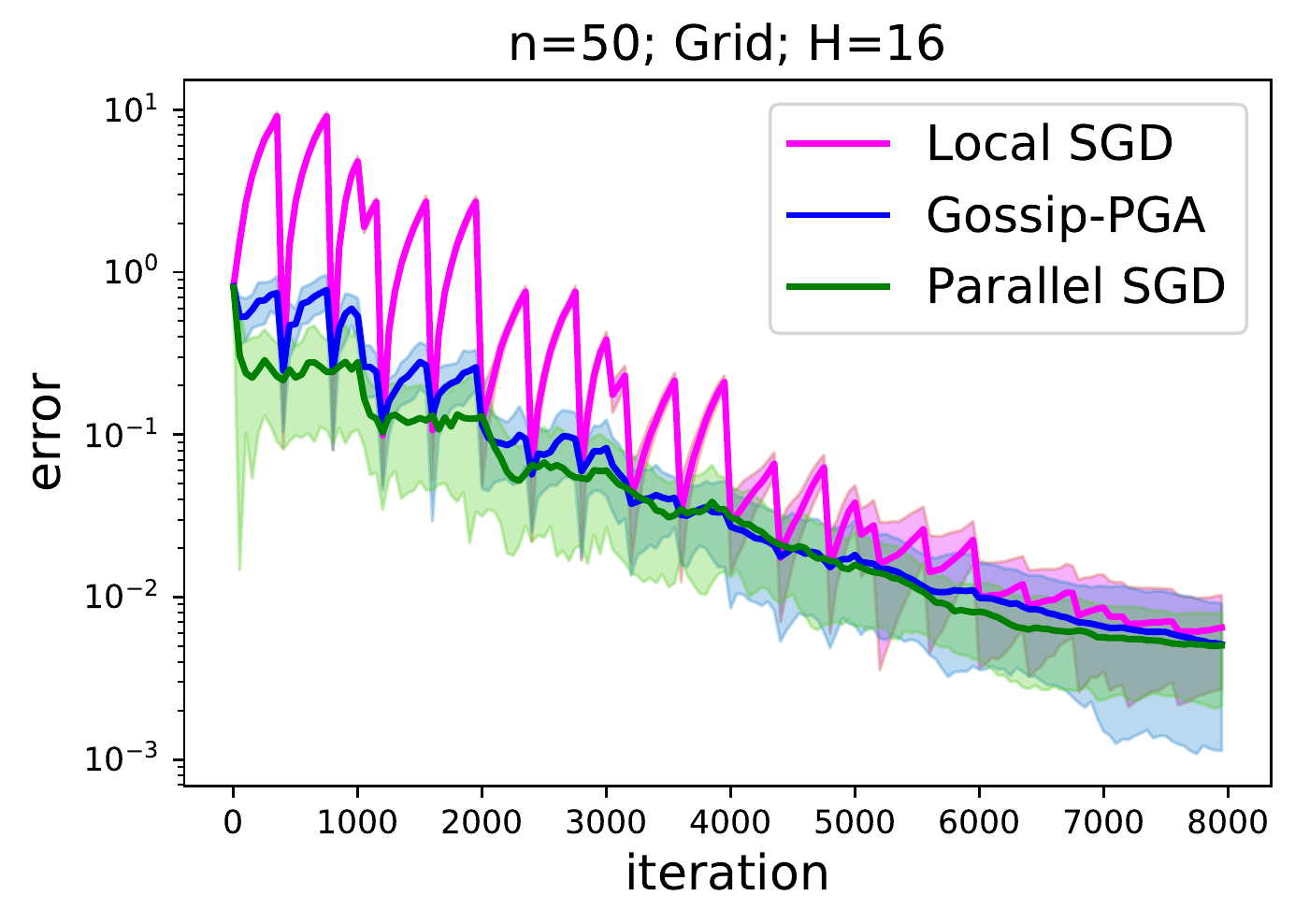} 
\includegraphics[scale=0.4]{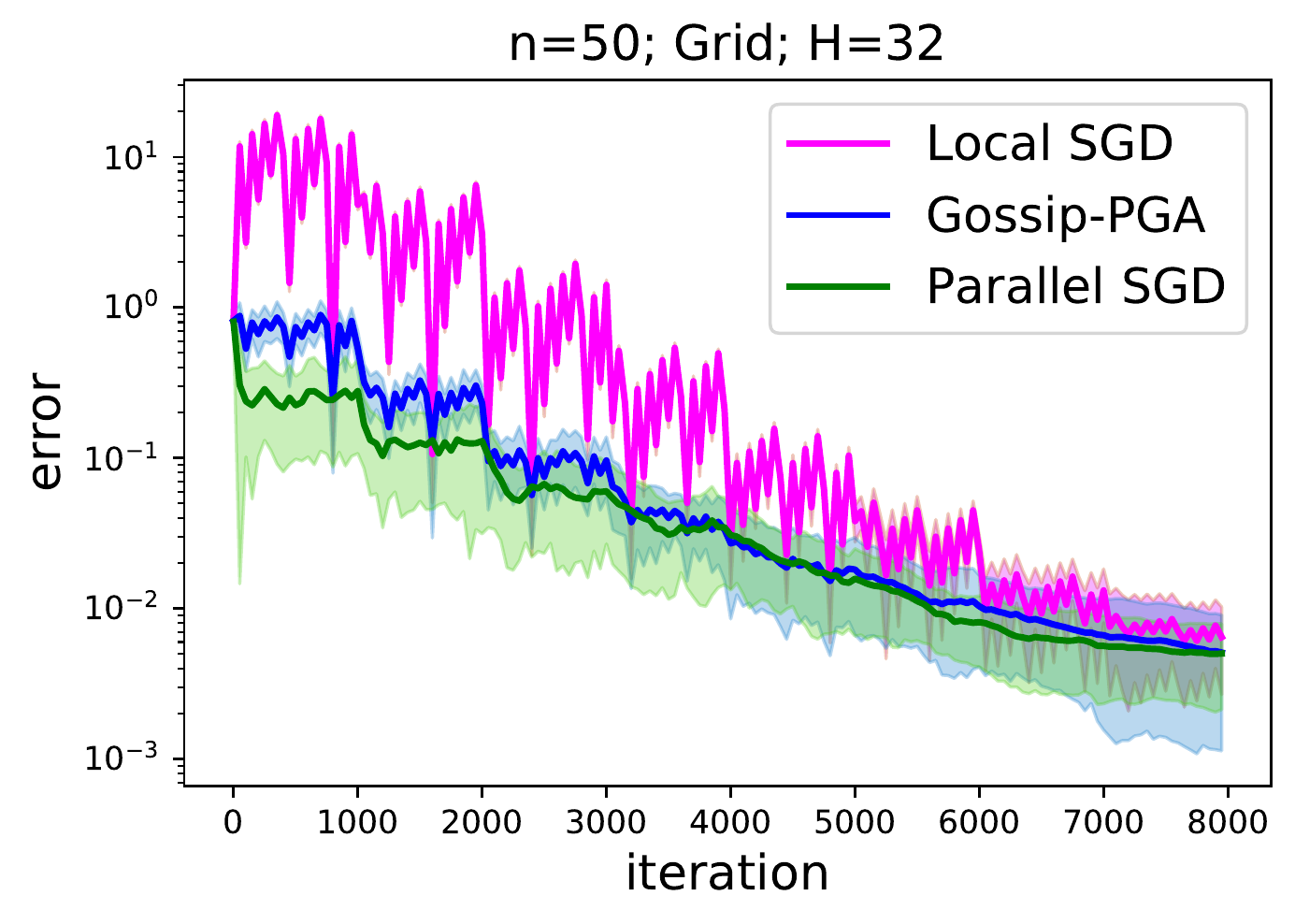} 
\includegraphics[scale=0.4]{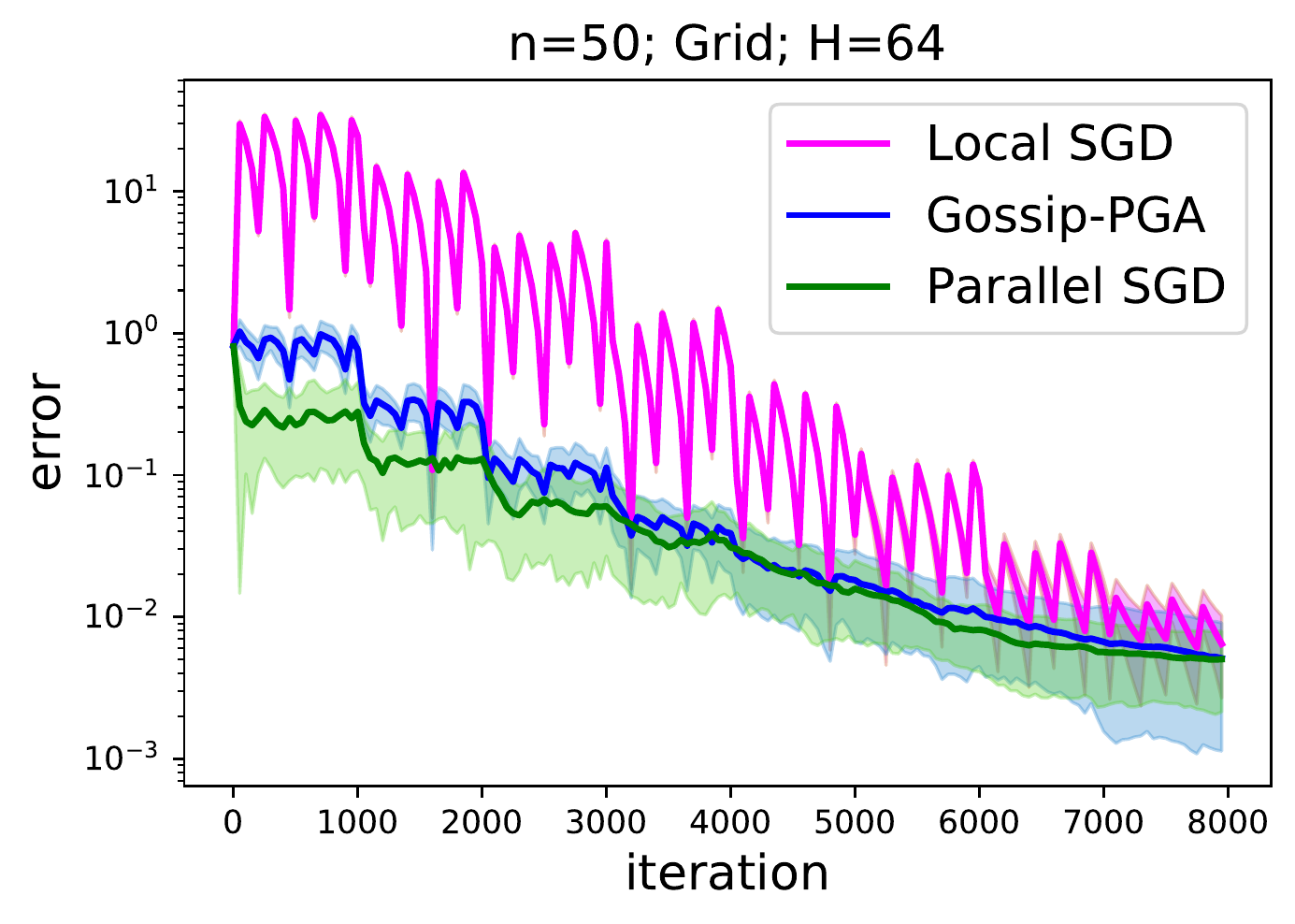} 
\caption{\small Convergence comparison between Gossip-PGA, Local SGD and parallel SGD on the logistic regression problem in non-iid data distributed setting over the grid topology with period $H=16$ (left), $H=32$ (middle), $H=64$ (right).}
\label{Fig:convex-comparison-local-appendix-noniid-different-period}
\vskip -0.2in
\end{figure*}

\vspace{1mm}
\subsection{More experiments on convex logistic regression.}

\vspace{0.5mm}

In this subsection we will test the performance of Gossip-PGA with iid data distribution and on different topologies. We will also compare it with Local SGD.

\noindent \textbf{Experiments on iid dataset}. Figure \ref{Fig:convex-comparison-gossip-appendix-iid} illustrates how Gossip SGD and Gossip-PGA converges under the iid data distributed setting over the ring topology. Similar to the non-iid scenario shown in Figure \ref{Fig:convex-comparison-gossip}, it is observed that Gossip-PGA always converges faster (or has shorter transient stages) than Gossip SGD. When network size gets larger and hence $\beta \to 1$, the superiority of Gossip-PGA gets more evident. Moreover, it is also noticed that the transient stage gap between Gossip SGD and Gossip-PGA is smaller than the non-iid scenario in all three plots in Figure \ref{Fig:convex-comparison-gossip-appendix-iid}. All these observations are consistent with the transient stage comparisons in Table \ref{table-transient-stage}.

\vspace{0.5mm}
\noindent \textbf{Experiments on different topologies}. Figure \ref{Fig:convex-comparison-gossip-appendix-noniid-different-topo} illustrates how Gossip SGD and Gossip-PGA converges over the exponential graph, grid and ring topology. For all plots, it is observed that Gossip-PGA is no worse than Gossip SGD. Moreover, as the network gets sparser and hence $\beta \to 1$ from the left plot to right, it is observed that the superiority of Gossip-PGA gets more evident, which is consistent with the transient stage comparisons between Gossip SGD and Gossip-PGA in Table \ref{table-transient-stage}.

\vspace{0.5mm}
\noindent \textbf{Comparison with Local SGD}. Figure \ref{Fig:convex-comparison-local-appendix-noniid-different-topo} illustrates how Local SGD and Gossip-PGA converges over the exponential graph, grid and ring topology. The period is set as $H = 16$. In all three plots, Gossip-PGA always converges faster than Local SGD because of the additional gossip communications. Moreover, since the exponential graph has the smallest $\beta$, it is observed Gossip-PGA has almost the same convergence performance as parallel SGD. Figure \ref{Fig:convex-comparison-local-appendix-noniid-different-period} illustrates how Local SGD and Gossip-PGA converges over the grid topology with different periods. It is observed that Gossip-PGA can be significantly faster when $H$ is large. All these observations are consistent with the transient stage comparisons in Table \ref{table-transient-stage-local}.

\vspace{1mm}
\subsection{More experiments on image classification.}
\label{app-more-exp-on-imagenet}

\begin{figure*}[t]
\vskip 0.2in
\begin{center}
\centerline{\includegraphics[width=0.45\textwidth]{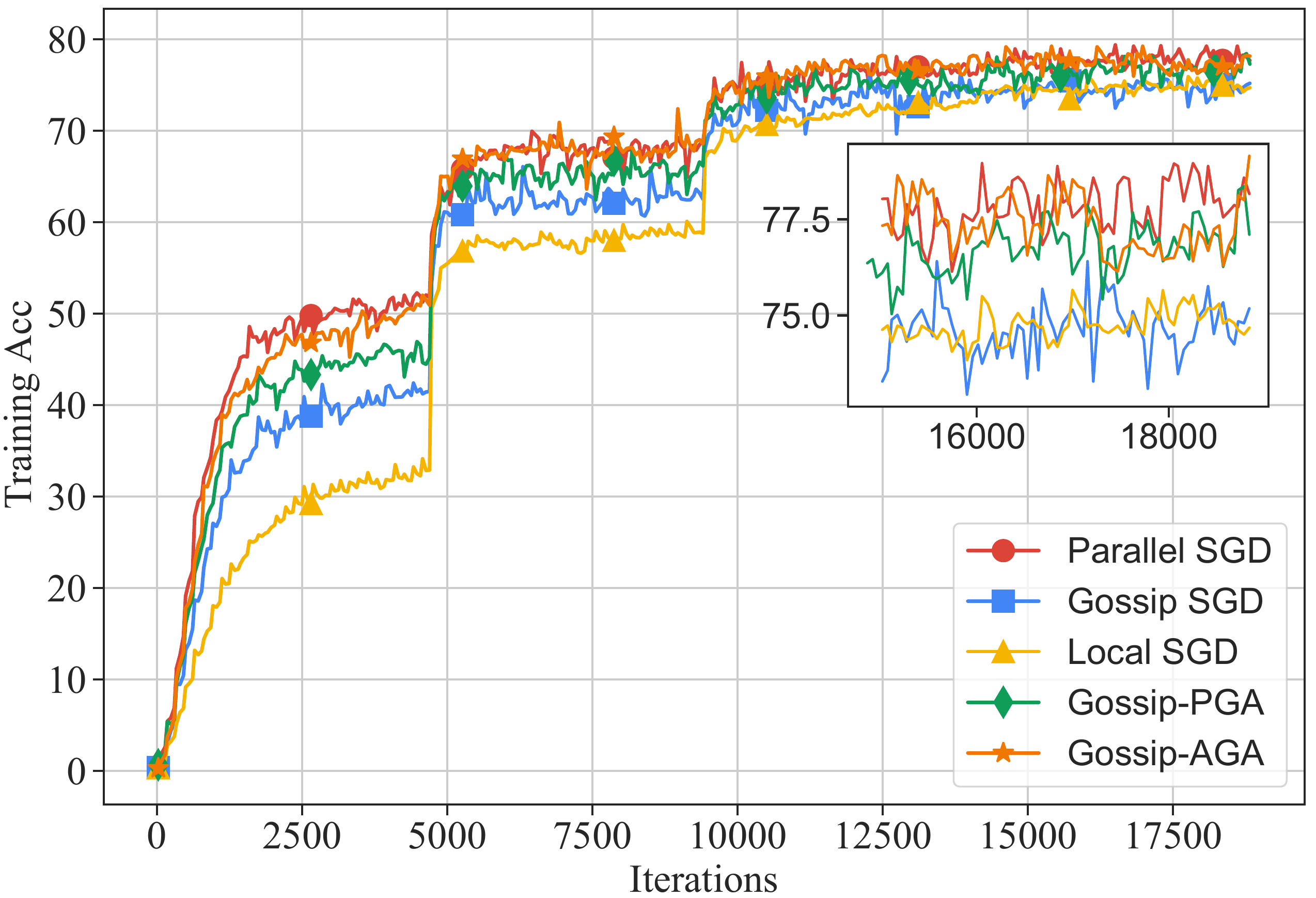} 
\includegraphics[width=0.45\textwidth]{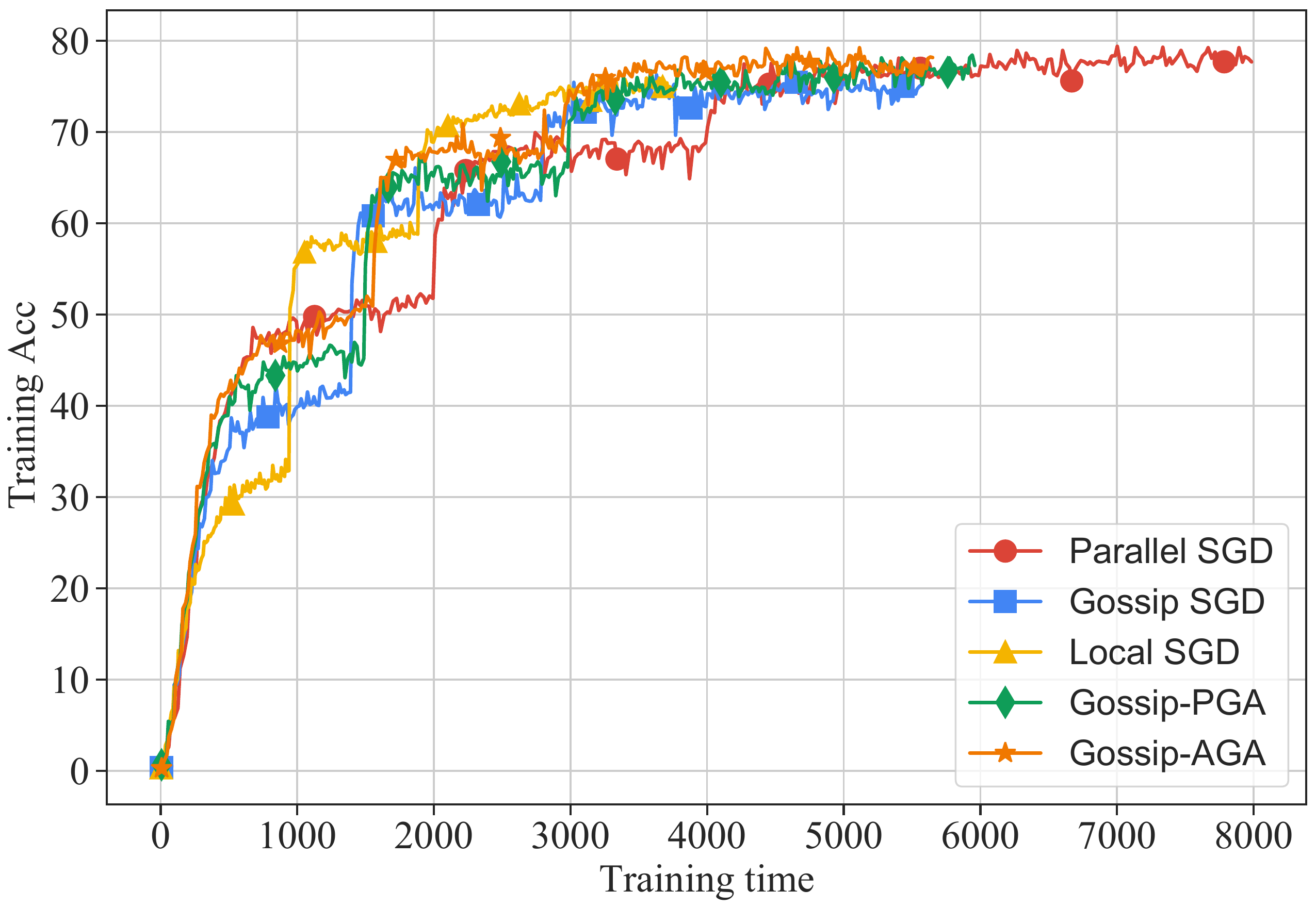}}
\caption{Convergence results on the ImageNet classification task.(a) Iteration-wise convergence in terms of training accuracy. (b) runtime-wise convergence speed in terms of training accuracy.}
\end{center}
\label{Fig:imagenet-acc}
\vskip -0.2in
\end{figure*}

\noindent \textbf{Training accuracy}. Figure \ref{Fig:imagenet-acc} shows the iteration-wise and time-wise training accuracy curves of aforementioned algorithms separately. In the left figure, it is observed Gossip-PGA/AGA converges faster (in iteration) and more accurate than local and Gossip SGD, which is consistent with our theory. In the right figure, it is observed that Gossip-PGA/AGA is the fastest method (in time) that can reach the same training accuracy as parallel SGD.

\begin{table}[h!]
\vskip 0.15in 
\begin{center}
\begin{small}
\begin{sc}
\begin{tabular}{cccccccc}
    \toprule
    & Parallel SGD & Gossip SGD & \multicolumn{5}{c}{Gossip-PGA} \\ 
    \midrule
    Period $H$ & - & - & 3 & 6 & 12 & 24 & 48 \\
    Val Acc.(\%) & 76.22 & 75.34 & 76.19 & 76.28  & 76.04 & 75.68 & 75.66    \\
    \bottomrule
\end{tabular}
\end{sc}
\end{small}
\end{center}
\vskip -0.1in
\caption{Comparison of Top-1 validation accuracy with different averaging period setting in Gossip-PGA.}
\label{Table:imagetnet_period}
\end{table}

\begin{algorithm}
\DontPrintSemicolon
\KwRequire{Initialize $x_{0,i}=x_0$, learning rate $\gamma>0$, topology matrix $W$ for all nodes $i \in \left\{1,2,...,n \right\}$, global averaging period $H=H_{init}$, $C \gets 0$,  $F_{init} \gets 0$, warmup iterations $K_w$}

\For{$k=0,1,2,...,T-1$, every node $i$}{
    $C \gets C + 1$ \;
    
    Sample mini-batch data $\bxi^{(k+1)}_{i}$ from local dataset\;
    Compute stochastic gradient ${\nabla}F_i(\x^{(k)}_{i};\bxi^{(k+1)}_{i})$ and loss $F_i(\x^{(k)}_{i};\bxi^{(k+1)}_{i})$ \;
    
    Conduct local udpate $\x^{(k+\frac{1}{2})}_{i} = \x^{(k)}_{i} - \gamma {\nabla}F_i(\x^{(k)}_{i};\bxi^{(k+1)}_{i}) $ \;
    
    \If{$C == H$}{
        $C \gets 0$ \;
        
        $x^{(k+1)}_{i} \gets \frac{1}{n}\sum_{j=1}^{n}x_{k+\frac{1}{2},j}$\;
        
        $F(x_{k};\xi_{k})  = \frac{1}{n} \sum_{i=1}^{n}F_i(x_{k,i};\xi_{k,i})$\;
        
        \If{$k<K_w$}{
        $F_{init} \gets \frac{1}{2}(F_{init}+F(x_{k};\xi_{k}))$ \;
        }
        \Else{
        $H \gets {\left\lceil\frac{F_{init}}{F(x_{k};\xi_{k})}H_{init}\right\rceil}$\;
        }
    }
    \Else{
    $\x^{(k+1)}_{i} = \sum_{j\in \cN_i}w_{ij} \x^{(k+\frac{1}{2})}_{j}$\;
    }
}
\caption{Gossip-AGA}
\label{Algorithm: Gossip-AGA}
\end{algorithm}

\noindent \textbf{The effect of averaging period.} Table \ref{Table:imagetnet_period} compares the top-1  accuracy in the validation set with a different averaging period setting in Gossip-PGA SGD. Compared to Gossip SGD, a relatively large global averaging period (48), roughly 2.1\% iterations with global averaging can still result in 0.32\% gain in validation accuracy. With a moderate global averaging period (6/12), the validation accuracy is comparable with the parallel SGD baseline. The communication overhead of global averaging can be amortized since it happens every $H$ iterations.

\noindent \textbf{Experiments on SGD optimizer (without momentum).} In previous Imagenet training, Nesterov momentum SGD optimizer is used. Following common practice \cite{lian2017can,assran2019stochastic,tang2018d}, we establish the convergence rate of the non-accelerated method while running experiments with momentum. For the sake of clarity, we further add a new series of experiments on Gossip-SGD without momentum, see Table \ref{Table:imagenet-sgd}. Gossip-PGA still outperform Gossip-SGD utilizing the SGD optimizer.

\begin{table}[h!]
\vskip 0.15in
\begin{center}
\begin{small}
\begin{sc}
\begin{tabular}{cc}
\toprule
Method & Acc. \%  \\
\midrule
Parallel SGD & 69.5                     \\
Gossip SGD & 68.47    \\
Gossip-PGA & 69.21 \\ \bottomrule
\end{tabular}
\end{sc}
\end{small}
\end{center}
\vskip -0.1in
\caption{Comparison of validation accuracy of Imagenet training on different algorithms with SGD optimizer.}
\label{Table:imagenet-sgd}
\end{table}

\section{Implementation of Gossip AGA }
\label{app-implem-AGA}
\noindent \textbf{Practical consideration}. 
The Gossip-AGA algorithm is listed in Algorithm \ref{Algorithm: Gossip-AGA}. We use a counter $C$ to record the number of gossip iterations since last global averaging. The global averaging period $H$ is initialized to a small value $H_{init}$ (e.g. 2$\sim$4). Once $C$ equals to current $H$, global averaging happens. In practice, we sample loss scores for the first fewer iterations and get a $F_{init}$ estimation in a running-average fashion. We remove the exponential term in the loss score ratio for flexible period adjustment.

\section{Comparison of communication overhead between gossip and All-Reduce}
\label{app-all-reduce}

\begin{table}[h]
\vskip 0.15in
\begin{center}
\begin{small}
\begin{sc}
\begin{tabular}{cccc}
\toprule
    model & \multicolumn{3}{c}{iteration time (ms)}  \\
     & no communication          & All-Reduce        &  Gossip \hspace{1.5mm}    \\ 
\midrule
ResNet-50 & 146 & 424 (278) & 296 (150)   \vspace{1.5mm}   \\
BERT    & 445 & 1913.8 (1468.8) & 1011.5 (566.5)
\\ \bottomrule
\end{tabular}
\end{sc}
\end{small}
\end{center}
\vskip -0.1in
\caption{Comparison of communication overhead between gossip and All-Reduce in terms of runtime.}
\label{table-communication-time-comparison}
\end{table}

Table \ref{table-communication-time-comparison} compares the overhead of different communication styles in two deep training tasks. The implementation details follow Appendix \ref{app-add-experiments}. For each profiling, we run a 500 iterations and take their average as the iteration time. As typically All-Reduce implementation containing overlapping between computation and communication, we run a series of separate experiments which do not perform communication (Column 2) to get communication overhead fairly (the figures in the brackets). For ResNet-50 training, gossip introduces 150ms communication overhead while All-Reduce needs 278ms. For BERT training, gossip introduces 566.5ms communication overhead while All-Reduce needs 1468.8ms with the tremendous model size of BERT-Large.

\end{document}